\PassOptionsToPackage{table}{xcolor}
\RequirePackage{amsmath}
\RequirePackage{amsfonts}

\documentclass{fairmeta}
\usepackage{newpxtext,newpxmath}
\renewcommand{\title}[1]{\newcommand{\titlelist}{{\fontsize{22}{26}\selectfont\sffamily\bfseries\color{clmblue} #1}}}

\usepackage{amsmath,amsfonts,bm}

\def\eqref#1{equation~\ref{#1}}

\def\1{\bm{1}}

\DeclareMathAlphabet{\mathsfit}{\encodingdefault}{\sfdefault}{m}{sl}
\SetMathAlphabet{\mathsfit}{bold}{\encodingdefault}{\sfdefault}{bx}{n}

\usepackage{xspace}
\usepackage{verbatim}
\usepackage{booktabs} 
\usepackage{arydshln} 
\usepackage{wrapfig} 
\usepackage{mathtools}
\usepackage{algorithm}
\usepackage{algorithmic}
\usepackage{pifont}       \usepackage{graphicx}     \usepackage{colortbl}     \usepackage{float}
\definecolor{tablered}{HTML}{FBEDEE}
\usepackage{listings}

\lstdefinestyle{drctx}{
  language=Python,
  basicstyle=\ttfamily\small,
  keywordstyle=\ttfamily,
  commentstyle=\color{gray},
  showstringspaces=false,
  keepspaces=true,
  columns=fixed,
  basewidth=0.53em,
  escapeinside={(*@}{@*)},
  xleftmargin=0pt,
  aboveskip=2pt,
  belowskip=0pt,
  frame=none,
}
\definecolor{ctxhl}{HTML}{C92A2A}
\definecolor{evoblue}{HTML}{1F77B4}
\definecolor{evopurple}{HTML}{7B3FA0}

\definecolor{readhl}{HTML}{2B8A3E}

\definecolor{PKUred}{HTML}{8B0012}
\colorlet{clmblue}{PKUred}

\newcommand{\clmhl}[1]{\textcolor{clmblue}{\textbf{#1}}}

\definecolor{ctxorigfill}{HTML}{BCD8EE}\definecolor{ctxorigedge}{HTML}{0072B2}\definecolor{ctxorigtext}{HTML}{0B3D6B}
\definecolor{ctxeditfill}{HTML}{F4C8D6}\definecolor{ctxeditedge}{HTML}{C23A63}\definecolor{ctxedittext}{HTML}{7A2040}

\colorlet{metablue}{clmblue}

\hypersetup{
    colorlinks=true,
    linkcolor=PKUred,
    citecolor=PKUred,
    urlcolor=PKUred
}

\addtocontents{toc}{\protect\setcounter{tocdepth}{-1}}
\usepackage{csquotes}
\newtcolorbox{quotebox}{
  enhanced,
  breakable,
  colback=gray!5,
  colframe=gray!35,
  boxrule=0.5pt,
  arc=2pt,
  left=8pt,
  right=8pt,
  top=6pt,
  bottom=6pt
}

\usepackage{wrapfig}
\usepackage{threeparttable} \usepackage{array}

\definecolor{oc-orange-9}{HTML}{D9480F}
\definecolor{oc-cyan-8}{HTML}{0C8599}
\definecolor{rs-color}{HTML}{e87ea1}
\definecolor{forestgreen}{HTML}{009B55}

\newtcolorbox{msgbox}[2][gray]{
    enhanced, breakable,
    colback=#1!6, colframe=#1!50!black!40,
    boxrule=0.5pt, arc=2pt, left=6pt, right=6pt, top=3pt, bottom=3pt,
    title={\scriptsize\sffamily\bfseries #2}, coltitle=black,
    colbacktitle=#1!15, toptitle=1.5pt, bottomtitle=1.5pt,
    before skip=4pt, after skip=4pt
}

\colorlet{swehl}{metabg}
\colorlet{storagepink}{metabg}

\usepackage{amsmath}

\newcommand{\modelname}{AUGUstus}
\newcommand{\modename}{AUGUstus-mode}
\newcommand{\grponame}{AUGUstus-GRPO}

\newcommand{\ours}{\clmhl{\modelname}\xspace}
\newcommand{\oursmode}{\clmhl{\modename}\xspace}
\newcommand{\oursgrpo}{\clmhl{\grponame}\xspace}

\newcommand{\papertitle}{AUGUstus: Autonomous Driving VLA Model with Unified Generation and Understanding}

\definecolor{tengbrown}{HTML}{7A4E2D}

\usepackage{tikz}

\title{\papertitle}
\hypersetup{pdftitle={AUGUstus: Autonomous Driving VLA Model with Unified Generation and Understanding},pdfauthor={Yuqi Ye et al.},pdfsubject={Unified vision-language-action models for autonomous driving}}

\author[1,*]{Yuqi~Ye}
\author[2,\ddagger]{Caigao~JIANG}
\author[3]{Zhuonan~Wang}
\author[2]{Congyuan~Pei}
\author[3]{Chao~Qu}
\author[1]{Changhao~Peng}
\author[1]{Junhong~Lin}
\author[1]{Qize~Yu}
\author[1]{Zetian~Song}
\author[2]{Guangzhao~Fu}
\author[2]{Chongruo~Wu}
\author[2]{Shiyu~Huang}
\author[2]{Shuchang~Zhou}
\author[2]{Honggou~Yang}
\author[1,\dagger]{Wei~Gao}

\affiliation[1]{Peking University}
\affiliation[2]{XPeng Motors}
\affiliation[3]{Fudan University}

\contribution[*]{First author}
\contribution[\ddagger]{Project lead}
\contribution[\dagger]{Corresponding author}

\metadata[Project Page]{\url{https://augustus-ad.github.io}}
\metadata[Code]{\url{https://github.com/haha-yuki-haha/AUGUstus}}
\metadata[Model]{\url{https://huggingface.co/yuki-hahaha/AUGUstus}}

\abstract{
Vision-language-action (VLA) models have shown strong potential for end-to-end autonomous driving by leveraging world knowledge and understanding to achieve great performance in long-tail scenarios. However, current VLAs suffer from two major limitations: 1) Degeneration into vision-action (VA) models. To achieve low latency, regression-based and flow-based methods directly generate actions, yet they do not sufficiently leverage the VLM's understanding, leading to unsafe and unexplainable actions. 2) Autoregressive high latency. Autoregressive methods sequentially generate chain-of-thought (CoT) and action tokens, fully leveraging the VLM's understanding, but high latency makes deployment difficult. To enable low-latency deployment while properly using CoT to assist action generation, we propose \ours{}\footnotemark[1], an $\underline{\textbf{A}}$utonomous-driving VLA model with $\underline{\textbf{U}}$nified $\underline{\textbf{G}}$eneration and $\underline{\textbf{U}}$nderstanding. Specifically, we first unify diffusion-based action generation and autoregressive CoT understanding in a single model, where understanding is optimized only to serve generation. More concretely, joint CoT supervision preserves the VLM's spatial understanding in the shared representation and keeps it useful for safer and more accurate trajectories. Second, we propose \oursmode, which orders the action generation before the CoT so that the CoT supervises the reliable action during training while inference needs only the action output, enabling low latency. Finally, we introduce \oursgrpo, a unified GRPO algorithm that jointly optimizes the action and CoT, further improving action performance. Extensive experiments demonstrate that \ours achieves 94.6 PDMS on NAVSIMv1 and 90.7 EPDMS on NAVSIMv2, establishing state-of-the-art (SOTA) planning performance without scorer module. On nuScenes, our method achieves an average L2 error of 0.27 m and an average collision rate of 0.05$\%$, both the best among the compared methods. To validate the practicality of \ours, we train it on our in-house dataset using 512 NVIDIA H800 GPUs for approximately two days, then evaluate it on an internal 1,000 km dataset and preliminarily deploy it on a real vehicle, achieving 12 Hz operation. 
}

\usepackage{placeins}
\begin{document}

\maketitle
\footnotetext[1]{AUGUstus refers to Augustus (Octavian), the first Roman emperor, whose consolidation of Roman authority symbolizes unification. Here, it denotes a unified generation and understanding VLA model.}
\setcounter{footnote}{1}

\section{Introduction} \label{intro}

Autonomous driving has gradually moved from modular stacks for perception, prediction, and planning toward end-to-end models to reduce error accumulation, including ST-P3~\citep{hu2022st}, UniAD~\citep{uniad}, and VAD~\citep{vad}. Yet their limited model capacity and training data restrict the world knowledge available to the policy, making long-tail scenes and semantically complex interactions difficult to handle reliably.

\begin{figure*}[t]
\centering
\includegraphics[width=\textwidth]{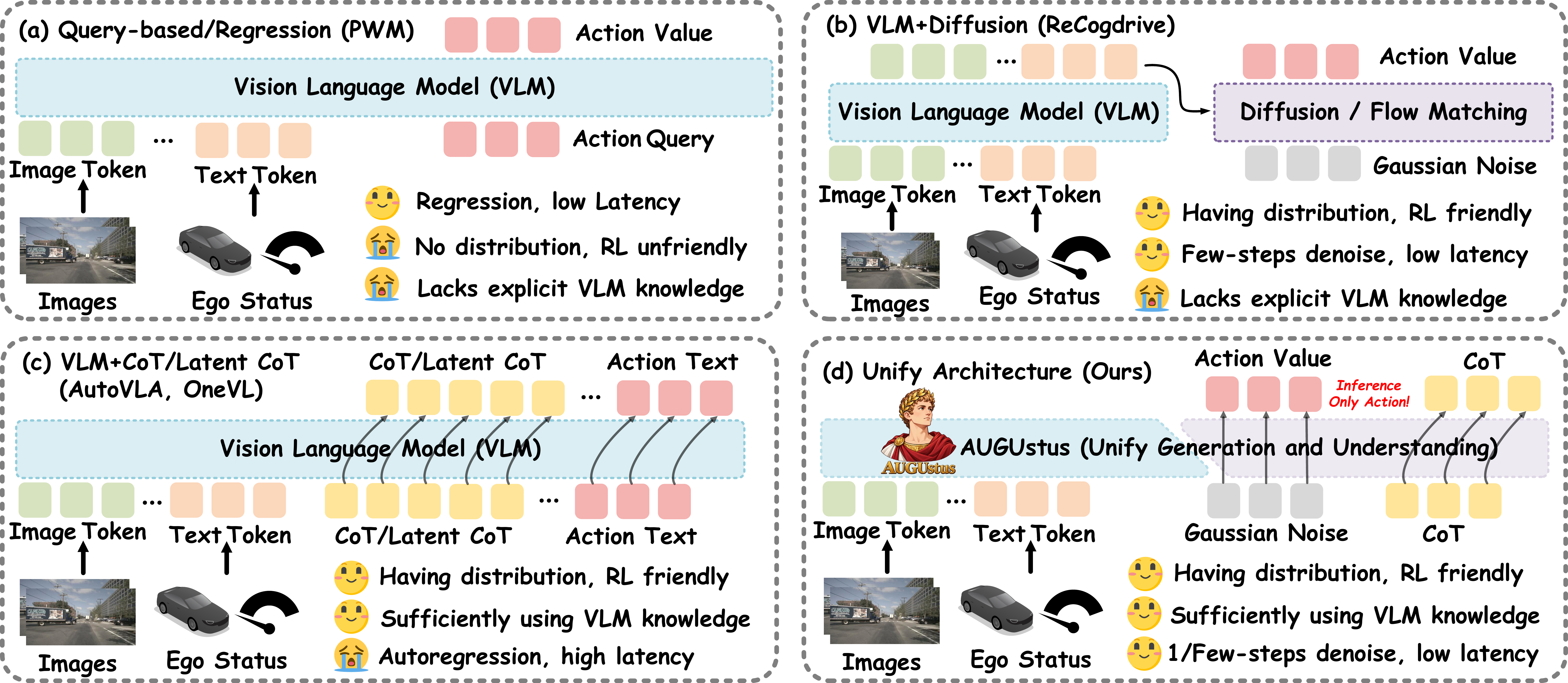}
\caption{\textbf{The difference between AUGUstus and other autonomous-driving VLA paradigms.} Our method unifies generation and understanding under AUGUstus-mode, so that joint CoT supervision preserves the VLM's understanding in the shared representation, yielding safer, more accurate actions with an RL-friendly action distribution and low-latency deployment.}
\label{fig:paradigm}
\end{figure*}

To address long-tail scenarios, many recent studies have explored large vision-language models (VLMs) as a promising foundation model for autonomous driving, leveraging their large-scale pre-training to supply broad visual concepts, language priors, and reasoning capabilities~\citep{pwm,recogdrive,autovla,autodrive_p3}. However, current mainstream autonomous driving VLA architectures still exhibit three limitations: \textbf{1) Regression head:} Despite low latency, these methods~\citep{pwm,imagidrive} produce no action distribution, making RL post-training difficult; moreover, they do not fully leverage VLM knowledge, resulting in low-quality generated actions (Fig.~\ref{fig:paradigm}a). \textbf{2) Diffusion or flow-matching expert:} To address the lack of action distribution for RL algorithm, some VLA methods~\citep{recogdrive} adopt diffusion or flow-matching experts for action generation, but they still do not fully exploit the VLM's spatial understanding, thereby producing unsafe and unexplainable actions (Fig.~\ref{fig:paradigm}b). \textbf{3) Autoregressive decoding:} To make full use of VLM understanding for action generation, some VLAs~\citep{autovla,autodrive_p3} serially generate CoT and action, which incurs high latency and makes real-time deployment difficult (Fig.~\ref{fig:paradigm}c). \textit{Overall, current VLAs struggle to reconcile the VLM's understanding with low-latency control. This raises a key question: can we design an architecture that keeps this action-relevant understanding while remaining low-latency and easy to deploy?}

We argue that these limitations lie in the separation between VLM-based understanding and action generation, rather than treating them as a unified process. Autonomous driving requires understanding and action generation to work in synergy, where accurate spatial understanding enables reliable action generation. However, conventional approaches decouple them: direct regression and diffusion/flow experts degenerate into Vision-Action (VA) models that produce unsafe and unexplainable actions, whereas autoregressive CoT leverages understanding but serializes reasoning before action, causing high latency. Accordingly, we reconsider the role of CoT understanding in autonomous driving. Rather than treating CoT as an independent reasoning stage or a serial prefix to actions, CoT should be jointly optimized with action generation within a unified architecture, where action generation is the core and CoT serves to improve it. Such a unified formulation keeps VLM spatial understanding coupled to and useful for the action policy, while inference only outputs actions, preserving an RL-friendly action distribution and low latency.

To address above fundamental limitations, we propose \clmhl{\modelname}, an $\underline{\textbf{A}}$utonomous-driving VLA model with $\underline{\textbf{U}}$nified $\underline{\textbf{G}}$eneration and $\underline{\textbf{U}}$nderstanding (Fig.~\ref{fig:paradigm}d). Our objective is action-centric: understanding is not trained for its own sake but to generate a safer and more accurate trajectory. Specifically, (1) we unify diffusion-based action generation and autoregressive CoT understanding in a single model, where joint CoT supervision preserves the VLM's spatial understanding in the shared representation for better action generation; (2) we propose \oursmode, an action-first and CoT-second ordering. This forces the model to learn a CoT that benefits action performance, while inference only needs the action output, thereby maintaining low latency. We explore various CoT designs and find that a CoT describing the specific positions and future orientations of other agents in the scene is most effective for action; and (3) we also introduce \oursgrpo, a unified GRPO algorithm that jointly optimizes the continuous action and CoT streams so that reasoning serves planning, further improving planning capabilities. 

We extensively evaluate \ours on closed-loop NAVSIM and open-loop nuScenes. \ours achieves 94.6 PDMS on NAVSIMv1 and 90.7 EPDMS on NAVSIMv2, without a scorer module. On nuScenes, our method achieves an average L2 error of 0.27\,m and an average collision rate of 0.05\%, both the best among the compared methods. To validate the practicality of \ours, we train it on our in-house dataset using 512 NVIDIA H800 GPUs for approximately two days, then evaluate it on an internal 1,000 km dataset and preliminarily deploy it on a real vehicle, achieving 12\,Hz operation. These results show that the shared model preserves the VLM's spatial understanding, improving action generation while meeting the latency requirements of real-time planning. The main contributions of this paper are summarized as follows:

\begin{enumerate}
\item We propose \ours, a unified VLA that integrates diffusion-based action generation and autoregressive CoT understanding in one transformer. The shared model preserves the VLM's spatial understanding for action generation, avoiding VLA-to-VA degeneration. \oursmode orders action before CoT to ensure low-latency inference.

\item We introduce \oursgrpo, a unified GRPO algorithm that jointly optimizes action and CoT. It combines JaT-GRPO, which uses closed-form Gaussian importance sampling ratios and KL terms, with CoT-GRPO, which uses token-level GRPO. The reasoning stream is rewarded for serving planning, thereby further improving action performance.

\item We demonstrate that \ours achieves state-of-the-art performance on multiple autonomous driving benchmarks, underscoring the effectiveness and generality of our approach. It also demonstrates practical scalability through large-scale training, validation on an internal driving dataset, and real-vehicle deployment.
\end{enumerate}

\section{Related Work} \label{sec:related}

\textbf{VLA models for autonomous driving.} Recent driving VLAs follow three designs. \emph{Query-based policies}~\citep{pwm,uniworldvla,lmgendrive} are low-latency but without action distribution for RL and weakly leveraging VLM semantics, which can yield unsafe and unexplainable actions. \emph{External flow-matching experts}~\citep{recogdrive} model continuous action distributions but keep the generator outside the VLM, weakening VLM semantics. \emph{Autoregressive policies}~\citep{autovla,autodrive_p3,vggdrive} expose semantic reasoning but serialize CoT and action tokens, incurring high latency. These designs struggle to trade off VLM understanding and low latency. \textbf{AUGUstus} unifies continuous action tokens and CoT understanding in one transformer under \textbf{AUGUstus-mode}, where understanding is trained to serve action generation rather than to maximize its own output quality, and deployment outputs actions directly.

\textbf{Reinforcement learning for driving VLAs.} Driving-oriented VLMs increasingly adopt GRPO-style post-training, such as AutoVLA, AutoDrive-R$^2$, AutoDrive-P$^3$, and RecogDrive~\citep{autovla,yuan2026autodrive,autodrive_p3,recogdrive}. RecogDrive directly supervises actions but lacks explicit CoT supervision, making it prone to reward hacking. Other autoregressive methods supervise token-level outputs, which slows both training and inference. In contrast, \textbf{AUGUstus-GRPO} jointly optimizes CoT and action under a shared group-relative advantage, with CoT further improving action performance.

\begin{figure*}[t]
\centering
\includegraphics[width=\textwidth]{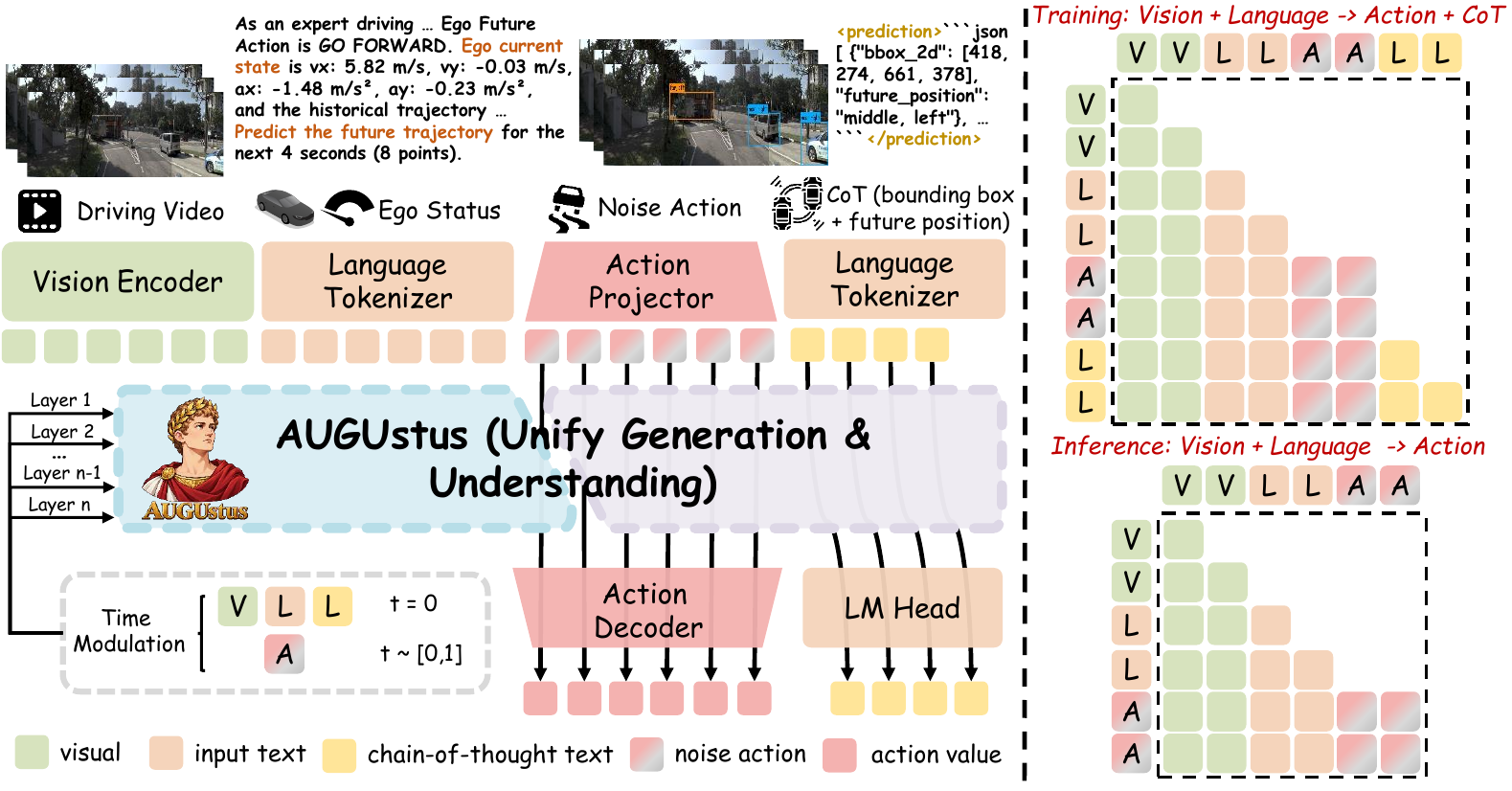}

\caption{\textbf{Overview of AUGUstus.} AUGUstus takes video and ego status as input and uses one shared transformer to first generate clean action and then understand a CoT containing key-object bounding boxes and future position. Action tokens use bidirectional attention and cannot attend to the CoT tokens. At inference, AUGUstus directly outputs the action to ensure low latency.}

\label{fig:pipeline}
\end{figure*}

\section{Methodology}
\label{sec:method}

This section introduces \ours framework, which unifies generation and understanding within a single transformer. First, we unify continuous, flow-based action generation with autoregressive, token-based CoT understanding in one transformer. Second, we propose \oursmode, an action-first, CoT-second training scheme in which the CoT stream is optimized to supervise and improve action generation. At inference, only the action stream is evaluated, ensuring low-latency deployment. Finally, we introduce \oursgrpo, a unified reinforcement learning algorithm that further improves the VLA's planning capability. 

\subsection{Unified framework}
\label{sec:arch}

As shown in Fig.~\ref{fig:pipeline}, for vision-based end-to-end driving, each sample contains camera video \(S\), ego state \(E\), route command \(C\), and history trajectory \(H\), which we write compactly as the driving context \(c=(S,E,C,H)\). During training, as summarized in Algorithm~\ref{alg:augustus}, the model additionally receives a noised trajectory \(x_t\) and the teacher-forced prefix \(y_{<k}\) CoT and predicts both a clean trajectory estimate and the next CoT token:
\begin{equation}
\bigl(c,\;x_t,\;y_{<k}\bigr)\;\longmapsto\;
\Bigl(\hat x_0,\;p_\theta\bigl(y_k\mid c,x_t,y_{<k}\bigr)\Bigr).
\label{eq:layout}
\end{equation}
Here \(x_0\in\mathbb{R}^{T\times3}\) is the continuous target trajectory, with each waypoint represented by planar position and heading. \(y\) is an autoregressive token sequence describing key objects with bounding box and future position. At inference, as summarized in Algorithm~\ref{alg:augustus_inference}, only context \(c\) is provided. The initial noise state \(x_1\) is denoised in one or few steps denoising to generate \(\hat x_0\), and CoT decoding is optional to ensure low latency. We denote this action-first, CoT-second ordering as \oursmode. As shown in Fig.~\ref{fig:pipeline}, its attention design is: context tokens are causal; action tokens form a bidirectional block attending to the full context and to each other; and CoT tokens are causal and additionally attend to the complete action block, while action tokens cannot attend to CoT tokens. 
Complete procedures are provided in Appendix~\ref{app:algorithms}.

\captionsetup{hypcap=false}
\begin{figure}[t]
\centering
\begin{minipage}[t]{0.485\linewidth}
\begin{algorithm}[H]
\small\raggedright
\captionsetup{font=small,justification=raggedright,singlelinecheck=false}
\caption{AUGUstus training}
\label{alg:augustus}
\begin{algorithmic}[1]
\REQUIRE Context $c$, trajectory $x_0$, CoT $y$.
\ENSURE $\hat x_0$, CoT probabilities, updated $\theta$.
\STATE Noise the target: $x_t=(1-t)x_0+t\varepsilon$.
\STATE Joint forward: context, noisy action, teacher-forced CoT.
\STATE Predict $\hat x_0$ and the next CoT tokens.
\STATE Update with flow-matching and CoT losses.
\end{algorithmic}
\end{algorithm}
\end{minipage}\hfill
\begin{minipage}[t]{0.485\linewidth}
\begin{algorithm}[H]
\small\raggedright
\captionsetup{font=small,justification=raggedright,singlelinecheck=false}
\caption{AUGUstus inference (1-step denoise)}
\label{alg:augustus_inference}
\begin{algorithmic}[1]
\REQUIRE Context $c$, noise $x_1$, step budget $N$.
\ENSURE Trajectory $\hat x_0$; optional post-hoc CoT.
\STATE Cache the causal context prefix.
\STATE Denoise the action block in $N$ evaluations.
\STATE For $N=1$, return $\hat x_0$ directly.
\STATE Skip CoT at inference; decode it afterward only if requested (e.g., during RL rollout sampling).
\end{algorithmic}
\end{algorithm}
\end{minipage}
\end{figure}

\subsection{Supervised Fine-Tuning for AUGUstus}
\label{sec:x0fm}

As summarized in Algorithm~\ref{alg:augustus}, supervised fine-tuning for \ours jointly trains conditional flow matching for actions and teacher-forced language modeling for CoT. Let \(x_0\) be the clean trajectory, \(\varepsilon\sim\mathcal N(0,I)\) the initial noise, and \(t\in[0,1]\) the flow time:
\begin{equation}
x_t=(1-t)x_0+t\varepsilon,\qquad
\hat x_0 = f_\theta(x_t,t,c).
\label{eq:cfm}
\end{equation}
Here \(f_\theta\) is \ours model, which directly predicts the clean trajectory \(\hat x_0\) rather than a velocity, similar to~\citep{jit} which predicts the clean image in the image domain. This \(x_0\) parameterization makes the deployed endpoint explicit, since one Euler step at \(t=1\) returns \(\hat x_0\) exactly, supporting both few-step sampling and a genuine one-step policy.

The supervised objective combines flow matching for continuous action reconstruction with teacher-forced CoT entropy minimization under the same context \(c\) and noise action \(x_t\):
\begin{equation}
\mathcal L_{\mathrm{SFT}}
=\lambda_{\mathrm{FM}}\,\underbrace{\mathbb E_{t,\varepsilon}\!\left[\left\|\hat x_0-x_0\right\|_1\right]}_{\text{flow matching loss}}
-\lambda_{\mathrm{CoT}}\,\underbrace{\mathbb E\!\left[\sum\nolimits_{k=1}^{L}\log p_\theta(y_k\mid y_{<k},c,x_t)\right]}_{\text{CoT likelihood}}.
\label{eq:sftloss}
\end{equation}

During SFT, we note that CoT for prediction, i.e., predicting the bounding boxes and future positions of key objects, is more beneficial for safe and efficient actions than CoT for perception or planning reasoning; additional CoT experiments are provided in the supplementary material.

\subsection{Reinforcement Learning for AUGUstus}
\label{sec:rl}
\label{sec:kernel}

\begin{figure}[t]
\centering
\includegraphics[width=\textwidth]{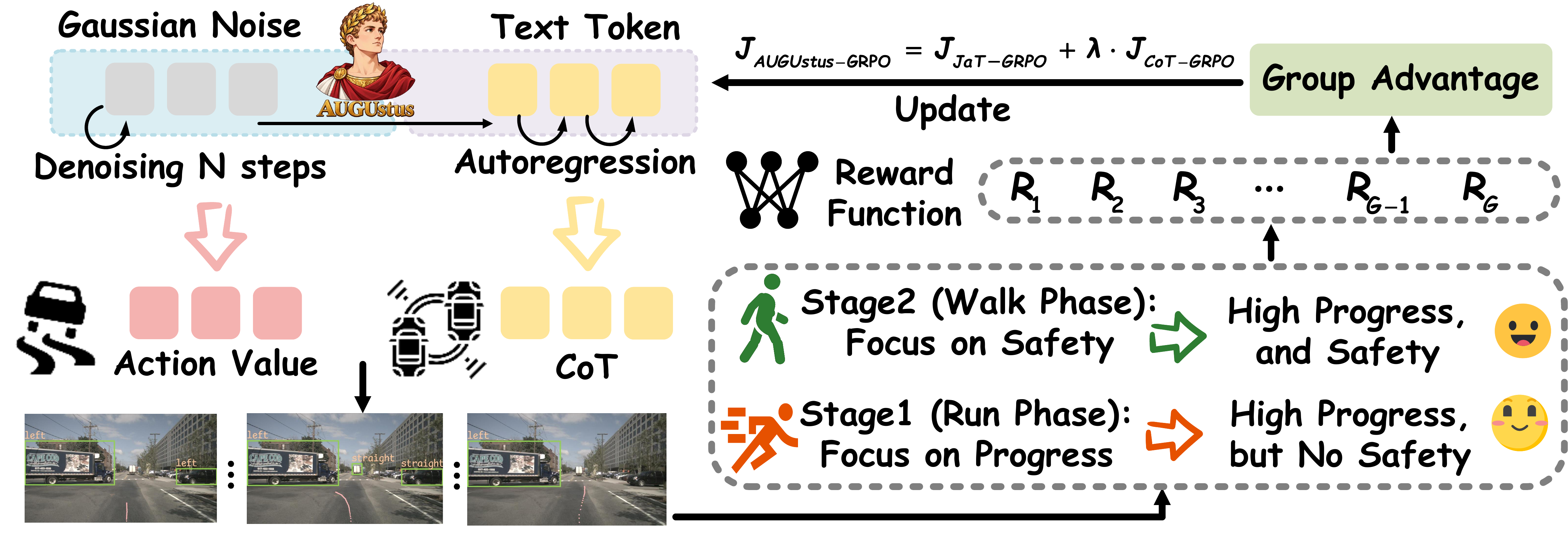}
\caption{\textbf{The pipeline of AUGUstus-GRPO.} Paired $x_0$ actions and autoregressive CoTs share a group-relative advantage, with closed-form JaT-GRPO and token-level CoT-GRPO. Stage I focuses on progress via PDMS reward, while Stage II focuses on safety via terminal and safety rewards.}
\label{fig:unify_rl}
\end{figure}

To further improve the VLA's planning capability, we introduce \oursgrpo, a unified RL method for continuous action and autoregressive CoT. As illustrated in Fig.~\ref{fig:unify_rl}, each rollout produces a paired action and CoT, and the two streams share a normalized group-relative advantage. JaT-GRPO applies GRPO to the continuous action branch under the \(x_0\) parameterization, while CoT-GRPO optimizes the CoT branch with token-level likelihoods. 

\captionsetup{hypcap=false}
\begin{figure}[t]
\centering
\begin{minipage}[t]{0.49\linewidth}
\begin{algorithm}[H]
\caption{AUGUstus-GRPO: Stage I}
\label{alg:augustus_grpo}
\begin{algorithmic}[1]
\REQUIRE SFT policy and contexts.
\ENSURE Progress-tuned policy $\pi_{\mathrm I}^{*}$.
\STATE Sample paired action and CoT rollouts.
\STATE Compute $R_{\mathrm{pred},i}$; set $R_{\mathrm{plan},i}=\mathrm{PDMS}_i$.
\STATE Normalize the shared group advantage $A_i$.
\STATE Jointly update JaT-GRPO and CoT-GRPO.
\end{algorithmic}
\end{algorithm}
\end{minipage}\hfill
\begin{minipage}[t]{0.50\linewidth}
\begin{algorithm}[H]
\caption{AUGUstus-GRPO: Stage II}
\label{alg:augustus_grpo_stage2}
\begin{algorithmic}[1]
\REQUIRE Stage-I policy $\pi_{\mathrm I}^{*}$ and contexts.
\ENSURE Safety-refined policy $\pi_{\mathrm{II}}^{*}$.
\STATE Sample paired action and CoT rollouts.
\STATE Keep $R_{\mathrm{pred},i}$; set $R_{\mathrm{plan},i}=R_{\mathrm{end},i}+R_{\mathrm{safe},i}$.
\STATE Normalize the shared group advantage $A_i$.
\STATE Jointly update JaT-GRPO and CoT-GRPO.
\end{algorithmic}
\end{algorithm}
\end{minipage}
\end{figure}

\textbf{Unified GRPO objective.} For each context, the behavior policy samples $G$ paired rollouts. In stage $s\in\{1,2\}$, $R_{\mathrm{pred},i}$ evaluates the CoT and $R_{\mathrm{plan},i}^{(s)}$ denotes the corresponding planning reward. We define the shared reward and normalized advantage as
\begin{equation}
R_i^{(s)}=R_{\mathrm{pred},i}+w_{\mathrm{plan}}R_{\mathrm{plan},i}^{(s)},
\qquad
A_i=\frac{R_i^{(s)}-\mathrm{mean}_{j}(R_j^{(s)})}{\mathrm{std}_{j}(R_j^{(s)})+\epsilon_0}.
\label{eq:reward}
\end{equation}

For each branch, we compute the importance sampling ratio between $\pi_\theta$ and $\pi_{\mathrm{old}}$:
\begin{equation}
r_{i,q}^{(\mathrm{action})}
=
\frac{\pi_\theta(x_{i,t-\Delta t}\mid x_{i,t},c_i)}
     {\pi_{\mathrm{old}}(x_{i,t-\Delta t}\mid x_{i,t},c_i)},
\qquad
r_{i,q}^{(\mathrm{CoT})}
=
\frac{\pi_\theta(y_{i,k}\mid y_{i,<k},c_i,x_i)}
     {\pi_{\mathrm{old}}(y_{i,k}\mid y_{i,<k},c_i,x_i)}.
\label{eq:main_ratio}
\end{equation}
Both branches are optimized with the same clipped GRPO objective,
\begin{equation}
\mathcal J_m=\mathbb E\!\left[\tfrac{1}{G}\sum_{i=1}^{G}\tfrac{1}{|\mathcal I_i^m|}
\sum_{q\in\mathcal I_i^m}\min\!\left(r_{i,q}^{(m)}A_i,
\operatorname{clip}(r_{i,q}^{(m)},1-\epsilon,1+\epsilon)A_i\right)\right],
\label{eq:main_clip}
\end{equation}
where \(m\in\{\mathrm{action},\mathrm{CoT}\}\), and \(\mathcal I_i^{\mathrm{action}}\) and \(\mathcal I_i^{\mathrm{CoT}}\) index the denoising steps and CoT token positions, respectively. Combining the two branches, the overall objective is

\begin{equation}
\mathcal L_{\mathrm{AUGUstus\text{-}GRPO}}
=\underbrace{-\mathcal J_{\mathrm{action}}+\beta_{\mathrm{action}}\mathrm{KL}_{\mathrm{act}}}_{\text{JaT-GRPO on }x_0}
+\alpha_{\mathrm{CoT}}\underbrace{\left(-\mathcal J_{\mathrm{CoT}}+\beta_{\mathrm{CoT}}\mathrm{KL}_{\mathrm{CoT}}\right)}_{\text{CoT-GRPO on language}}.
\label{eq:unify}
\end{equation}

\textbf{Closed-form JaT-GRPO.} The action-policy derivation starts from the linear conditional-flow path $x_t=(1-t)x_0+t\varepsilon$, with $\varepsilon\sim\mathcal N(0,I)$. Following Flow-GRPO~\citep{flow_grpo}, we adopt its reverse transition:
\begin{equation}
x_{t-\Delta t}=x_t-\left[v_\theta+\frac{g(t)^2}{2t}\bigl(x_t+(1-t)v_\theta\bigr)\right]\Delta t
  +g(t)\sqrt{\Delta t}\,z,\qquad z\sim\mathcal N(0,I).
\label{eq:kernel}
\end{equation}
Here $v_\theta$ is computed by our predicted clean trajectory $\hat x_0^\theta(x_t,t,c)$ through:
\begin{equation}
v_\theta(x_t,t,c)=\frac{x_t-\hat x_0^\theta(x_t,t,c)}{t}.
\label{eq:rl_path}
\end{equation}
Using Eq.~(\ref{eq:rl_path}) and $g(t)^2=\lambda t$ in Eq.~(\ref{eq:kernel}) yields a transition with mean $\mu_\theta=c_xx_t+c_0\hat x_0^\theta$ and variance $\sigma_s^2I$. Only $\hat x_0^\theta$ is policy-dependent, and the covariance is fixed. Thus, the $\pi_\theta$, $\pi_{\mathrm{old}}$, and $\pi_{\mathrm{ref}}$ policies induce Gaussian transitions with the same covariance $\sigma_s^2I$ and means $\mu_\theta$, $\mu_{\mathrm{old}}$, and $\mu_{\mathrm{ref}}$, respectively. With $\delta_{\mathrm{old}}=\hat x_0^\theta-\hat x_0^{\mathrm{old}}$ and $\delta_{\mathrm{ref}}=\hat x_0^\theta-\hat x_0^{\mathrm{ref}}$, the importance ratio and KL simplify to:
\begin{equation}
\begin{aligned}
\log r_t^{(\mathrm{action})}
&=\tfrac{c_0}{\sigma_s}\langle\delta_{\mathrm{old}},z\rangle
 -\tfrac{c_0^2}{2\sigma_s^2}\|\delta_{\mathrm{old}}\|_2^2,
&\mathrm{KL}_{\mathrm{act}}
&=\tfrac{c_0^2}{2\sigma_s^2}\|\delta_{\mathrm{ref}}\|_2^2.
\end{aligned}
\label{eq:main_closed_form}
\end{equation}
With the closed-form solution, we plug resulting ratios and KL terms into Eq.~(\ref{eq:main_clip}), enabling optimization with few-step and single-step denoising. The complete derivation is in Appendix~\ref{app:derivation}.

\textbf{Two-stage RL.} \oursgrpo uses a two-stage training procedure inspired by Run-then-Walk scheduling~\citep{ye2026gottarunwalkrunthenwalk}. As summarized by Algorithms~\ref{alg:augustus_grpo} and~\ref{alg:augustus_grpo_stage2}, Stage I starts from the SFT policy and, following Eq.~\ref{eq:reward}, sets the planning reward \(R_{\mathrm{plan},i}^{(1)}=\mathrm{PDMS}_i\) while keeping \(R_{\mathrm{pred},i}\). This encourages assertive progress and aligns the policy with the benchmark objective. Stage II starts from the Stage-I checkpoint and sets \(R_{\mathrm{plan},i}^{(2)}=R_{\mathrm{end},i}+R_{\mathrm{safe},i}\), preserving the same \(R_{\mathrm{pred},i}\). It therefore retains the progress learned in Stage I while refining the trajectory toward accurate endpoints and safe behavior. In both stages, the shared reward and advantage in Eq.~\ref{eq:reward} jointly update JaT-GRPO and CoT-GRPO; only planning reward is different. 

\begin{table}[t]
\caption{\textbf{Performance comparison on NAVSIMv1 benchmark.}}
\label{tab:main}
\small
\setlength{\tabcolsep}{4.2pt}
\renewcommand{\arraystretch}{1.05}
\resizebox{1.0\linewidth}{!}{
\begin{tabular}{l c c c c c c c >{\columncolor{tablered}}c}
\toprule
\textbf{Method} & \textbf{Ref} & \textbf{RL} & \textbf{NC$\uparrow$} & \textbf{DAC$\uparrow$} & \textbf{EP$\uparrow$} & \textbf{TTC$\uparrow$} & \textbf{Comf.$\uparrow$} & \textbf{PDMS$\uparrow$} \\
\midrule
Human & - & \ding{55} & 100.0 & 100.0 & 87.5 & 100.0 & 100.0 & 94.8 \\
\midrule
\multicolumn{9}{c}{\textbf{End-to-end small models}} \\
\midrule
TransFuser~\citep{transfuser} & TPAMI'23 & \ding{55} & 97.8 & 92.6 & 78.9 & 92.9 & 99.9 & 83.9 \\
PARA-Drive~\citep{paradrive} & CVPR'24 & \ding{55} & 97.9 & 92.4 & 79.3 & 93.0 & 99.8 & 84.0 \\
Hydra-MDP~\citep{hydra_mdp} & CVPR'24 & \ding{55} & 98.3 & 96.0 & 78.7 & 94.6 & \textbf{100.0} & 86.5 \\
WoTE~\citep{wote} & ICCV'25 & \ding{55} & 98.5 & 96.8 & 81.9 & 94.9 & \textbf{100.0} & 88.3 \\
\midrule
\multicolumn{9}{c}{\textbf{Regression-based VLA planners}} \\
\midrule
ImagiDrive~\citep{imagidrive} & ICRA'26 & \ding{55} & 97.9 & 95.5 & 80.7 & 93.1 & 99.9 & 86.4 \\
PWM~\citep{pwm} & NeurIPS'25 & \ding{55} & 98.6 & 95.9 & 81.8 & 95.4 & \textbf{100.0} & 88.1 \\
DriveVLA-W0~\citep{drivevla_w0} & ICLR'26 & \ding{52} & 98.7 & 99.1 & 83.3 & 95.3 & 99.3 & 90.2 \\
\midrule
\multicolumn{9}{c}{\textbf{Autoregressive VLA planners}} \\
\midrule
AutoVLA~\citep{autovla} & NeurIPS'25 & \ding{52} & 98.4 & 97.5 & 81.9 & 98.0 & 99.9 & 89.1 \\
AdaThinkDrive~\citep{adathinkdrive} & ICRA'26 & \ding{52} & 98.4 & 97.8 & 84.4 & 95.2 & \textbf{100.0} & 90.3 \\
AutoDrive-R$^2$~\citep{yuan2026autodrive} & ICLR'26 & \ding{52} & 98.5 & 95.9 & 82.7 & 95.4 & \textbf{100.0} & 89.1 \\
AutoDrive-P$^3$~\citep{autodrive_p3} & ICLR'26 & \ding{52} & 98.9 & 97.7 & 83.7 & 96.6 & 99.9 & 90.2 \\
VGGDrive~\citep{vggdrive} & CVPR'26 & \ding{55} & 98.6 & 96.3 & 82.9 & 95.6 & \textbf{100.0} & 88.8 \\
\midrule
\multicolumn{9}{c}{\textbf{Flow-matching VLA planners}} \\
\midrule
ReCogDrive~\citep{recogdrive} & ICLR'26 & \ding{52} & 98.1 & 97.7 & 86.5 & 94.9 & \textbf{100.0} & 90.6 \\
\midrule
\midrule
\textbf{AUGUstus (Ours, 1-step denoising)} & - & \ding{52} & 98.6 & 98.7 & 91.4 & 96.6 & \textbf{100.0} & 94.0 \\
\textbf{AUGUstus (Ours, 3-step denoising)} & - & \ding{52} & \textbf{98.9} & \textbf{98.9} & \textbf{91.9} & \textbf{97.1} & \textbf{100.0} & \textbf{94.6} \\
\bottomrule
\end{tabular}}
\end{table}

\begin{table}[t]
\centering
\caption{\textbf{Performance comparison on NAVSIMv2 benchmark.}}
\label{tab:epdms}
\small
\setlength{\tabcolsep}{3.4pt}
\resizebox{1.0\linewidth}{!}{
\begin{tabular}{l c c c c c c c c c >{\columncolor{tablered}}c}
\toprule
\multicolumn{1}{c}{\textbf{Method}} & \textbf{NC}$\uparrow$ & \textbf{DAC}$\uparrow$ & \textbf{DDC}$\uparrow$ & \textbf{TLC}$\uparrow$ & \textbf{EP}$\uparrow$ & \textbf{TTC}$\uparrow$ & \textbf{LK}$\uparrow$ & \textbf{HC}$\uparrow$ & \textbf{EC}$\uparrow$ & \textbf{\begin{tabular}[c]{@{}c@{}}\cellcolor{tablered}EPDMS$\uparrow$\\ \cellcolor{tablered}False / True\end{tabular}} \\
\midrule
Human & 100.0 & 100.0 & 99.8 & 100.0 & 87.4 & 100.0 & 100.0 & 98.1 & 90.1 & \cellcolor{tablered}90.3 / 94.5 \\
\midrule
Ego Status MLP & 93.1 & 77.9 & 92.7 & 99.6 & 86.0 & 91.5 & 89.4 & \underline{98.3} & 85.4 & \cellcolor{tablered}64.0 / -- \\
Transfuser~\citep{transfuser} & 96.9 & 89.9 & 97.8 & 99.7 & 87.1 & 95.4 & 92.7 & \underline{98.3} & \underline{87.2} & \cellcolor{tablered}76.7 / 84.0 \\
Hydra-MDP++~\citep{hydra_mdp} & 97.2 & 97.5 & 99.4 & 99.6 & 83.1 & 96.5 & 94.4 & 98.2 & 70.9 & \cellcolor{tablered}81.4 / -- \\
DiffusionDrive~\citep{diffusiondrive} & 98.2 & 96.2 & \underline{99.5} & 99.8 & \underline{87.4} & 97.3 & \underline{96.9} & \textbf{98.4} & \textbf{87.7} & \cellcolor{tablered}84.7 / 88.2 \\
WoTE~\citep{wote} & 98.5 & 96.8 & 98.8 & 99.8 & 86.1 & 97.9 & 95.5 & \underline{98.3} & 82.9 & \cellcolor{tablered}84.2 / 87.7 \\
ReCogDrive~\citep{recogdrive} & 98.2 & 97.7 & 98.4 & \textbf{100.0} & \textbf{89.9} & 97.4 & 90.8 & 97.7 & 26.8 & \cellcolor{tablered}82.7 / -- \\
AutoDrive-P\textsuperscript{3}~\citep{autodrive_p3} & \underline{98.9} & \underline{97.6} & 98.9 & 99.8 & 86.8 & \textbf{98.5} & 95.4 & \underline{98.3} & 80.6 & \cellcolor{tablered}\underline{85.2} / \underline{88.7} \\
\midrule
\textbf{AUGUstus (Ours, 1-step denoising)} & \textbf{99.0} & \textbf{97.9} & \textbf{99.7} & \underline{99.9} & 87.3 & \underline{98.4} & \textbf{98.0} & \underline{98.3} & \underline{87.2} & \cellcolor{tablered}\textbf{87.1} / \textbf{90.7} \\
\bottomrule
\end{tabular}}
\end{table}

\section{Experiments}
\label{sec:exp}

\subsection{Benchmarks and Baselines}
\label{sec:experimental_setup}
\label{sec:sftproto}

\textbf{Public benchmarks.} Following \citep{autovla, recogdrive, autodrive_p3}, we evaluate on the NAVSIM~\citep{navsim} and nuScenes~\citep{caesar2020nuscenes} benchmarks. For both benchmarks, the input is a 2-second, 4-frame clip from the front, front-left, and front-right cameras at \(2048{\times}512\). \ours predicts 8 waypoints over a 4-second horizon for NAVSIM, and 6 waypoints over a 3-second horizon for nuScenes. For NAVSIM, we report the official PDMS on NAVSIMv1 and EPDMS on NAVSIMv2. For nuScenes, following the ST-P3 protocol~\citep{hu2022st}, we report L2 error and collision rate.

\textbf{In-house benchmarks and driving test.} To validate the practicality of \ours, we train it on in-house dataset without using any public dataset, using 512 NVIDIA H800 GPUs for approximately two days with a global batch size of 2048. We first evaluate it on in-house 1,000 km driving set covering safety, comfort, efficiency, and compliance metrics, and then deploy the model on a real vehicle for a preliminary driving test in Guangzhou, China.

\textbf{Baselines.} We use Qwen3.5-2B~\citep{qwen3.5} as the base model. On public datasets, we compare against mainstream approaches, including end-to-end small planners and representative autoregressive, regression-based, and diffusion-based VLA planners. On the in-house dataset, we compare \ours with diffusion-based and regression-based architectures. Note that, considering the real-time requirements for deployment, unless otherwise specified, \textbf{inference in this paper refers to 1-step denoising without the subsequent CoT generation}.

\subsection{Main Results} \label{sec:main}

\label{sec:public_results}
\textbf{Public benchmark results.} As shown in Table~\ref{tab:main}, \ours reaches \textbf{94.0/94.6 PDMS} with 1/3-step denoising on NAVSIMv1, only 0.2 below human performance (94.8), while only generating a single trajectory without any scorer module. On NAVSIMv2, as shown in Table~\ref{tab:epdms}, \ours obtains the SOTA performance, \textbf{90.7 EPDMS} with the human-penalty filter (87.1 without it) with only 1-step denoising. On nuScenes, as shown in Table~\ref{tab:nuscenes_zeroshot}, our method with only 1-step denoising achieves \textbf{0.27\,m} average L2 error and \textbf{0.05\%} average collision rate.

\begin{figure}[t]
\centering

\begin{minipage}[t]{0.48\textwidth}
    \centering
    \includegraphics[width=\linewidth]{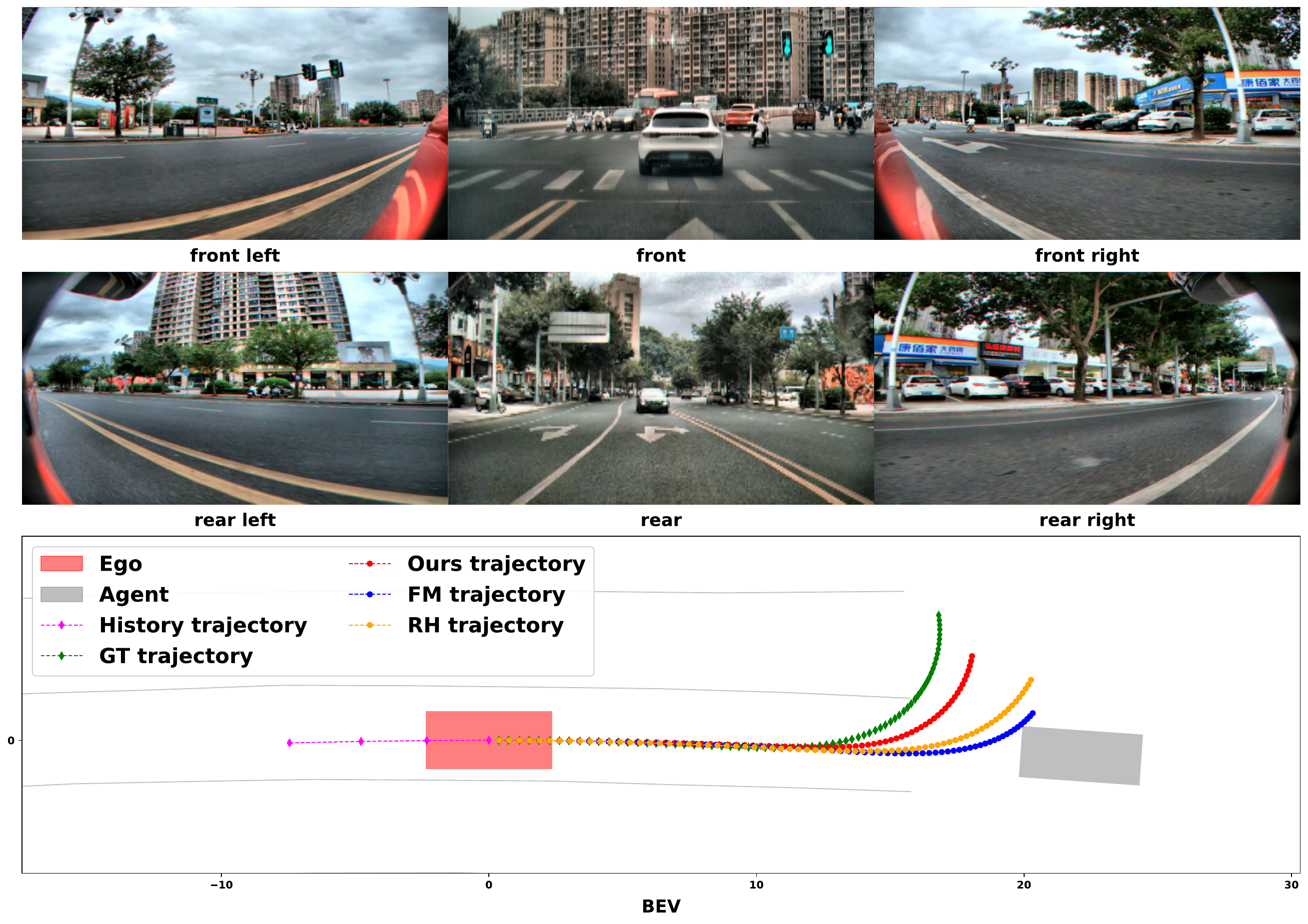}
    
    \captionsetup{justification=centering,singlelinecheck=false}
    \captionof{figure}{\textbf{In-house evaluation visualization.}}
    \label{fig:demo_internal}
\end{minipage}
\hfill
\begin{minipage}[t]{0.48\textwidth}
    \centering
    \includegraphics[width=\linewidth]{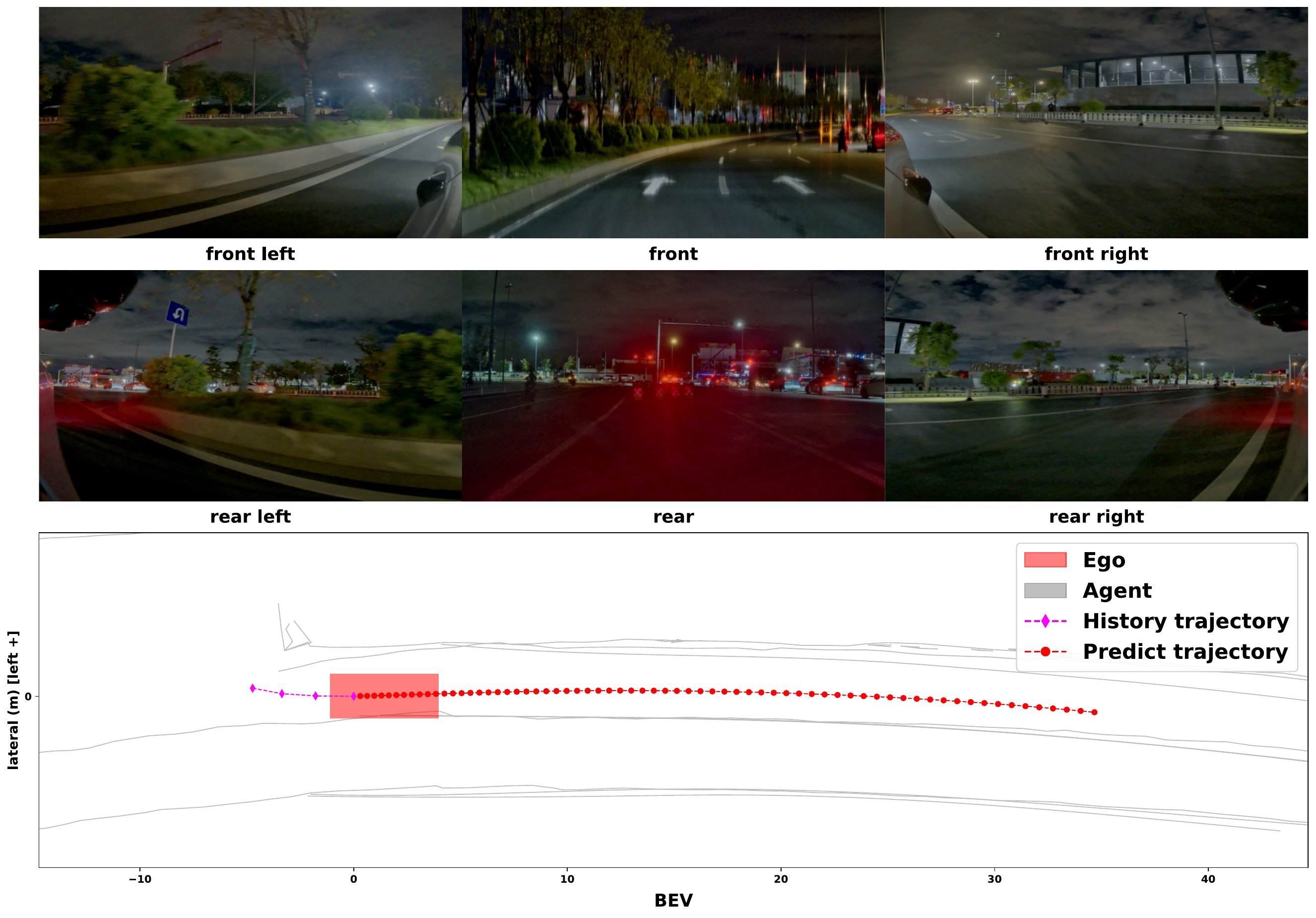}
    
    \captionsetup{justification=centering,singlelinecheck=false}
    \captionof{figure}{\textbf{Real driving test visualization.}}
    \label{fig:demo_real}
\end{minipage}

\end{figure}

\label{sec:internal_eval}
\textbf{In-house benchmarks and real driving test.} Beyond public benchmarks, we evaluate VLM+flow matching, VLM+regression, and \ours on an internal 1,000 km driving dataset. We report macro-averages across safety, comfort, efficiency, and compliance, where positive values indicate worse performance and \ours is fixed at 0\%. \ours performs best in safety, comfort, and efficiency, with the largest gaps in safety and comfort; the flow-matching baseline is slightly better in compliance (Table~\ref{tab:internal}). As shown in Table~\ref{tab:latency}, \ours achieves 13.6\,Hz on a single H800 and 12.0\,Hz on the real vehicle in our preliminary assisted driving tests. Figs.~\ref{fig:demo_internal} and~\ref{fig:demo_real} further visualize its qualitative behavior on the in-house evaluation and real driving test. Further details and supplementary videos are provided in the supplementary material.

\begin{table}[t]
\centering
\caption{\textbf{Performance comparison on nuScenes benchmark.}}
\label{tab:nuscenes_zeroshot}
\small
\setlength{\tabcolsep}{4.0pt}
\resizebox{1.0\linewidth}{!}{
\begin{tabular}{l c c c c >{\columncolor{tablered}}c c c c >{\columncolor{tablered}}c}
\toprule
\multicolumn{1}{c}{\multirow{2}{*}[-1ex]{\textbf{Method}}} & \multicolumn{1}{c}{\multirow{2}{*}[-1ex]{\textbf{Ref}}} & \multicolumn{4}{c}{\textbf{L2 (m)} $\downarrow$} & \multicolumn{4}{c}{\textbf{Collision} (\%) $\downarrow$} \\
\cmidrule(lr){3-6} \cmidrule(lr){7-10}
 & & 1s & 2s & 3s & Avg. & 1s & 2s & 3s & Avg. \\
\midrule
ST-P3~\citep{hu2022st} & ECCV'22 & 1.33 & 2.11 & 2.90 & 2.11 & 0.23 & 0.62 & 1.27 & 0.71 \\
VAD~\citep{vad} & ICCV'23 & 0.17 & 0.34 & 0.60 & 0.37 & 0.07 & 0.10 & 0.24 & 0.14 \\
UniAD~\citep{uniad} & CVPR'23 & 0.44 & 0.67 & 0.96 & 0.69 & 0.04 & 0.08 & 0.23 & 0.12 \\
\midrule
GPT-Driver~\citep{mao2023gpt} & arXiv'23 & 0.20 & 0.40 & 0.70 & 0.44 & 0.04 & 0.12 & 0.36 & 0.17 \\
DriveVLM~\citep{tian2024drivevlm} & CoRL'25 & 0.18 & 0.34 & 0.68 & 0.40 & 0.10 & 0.22 & 0.45 & 0.27 \\
RDA-Driver~\citep{huang2024making} & arXiv'24 & 0.17 & 0.37 & 0.69 & 0.40 & \underline{0.01} & 0.05 & 0.26 & 0.10 \\
OmniDrive~\citep{wang2024omnidrive} & CVPR'25 & \textbf{0.14} & \underline{0.29} & \underline{0.55} & \underline{0.33} & \underline{0.01} & \underline{0.04} & 0.27 & 0.11 \\
OpenDriveVLA~\citep{opendrivevla} & AAAI'26 & \textbf{0.14} & 0.30 & \underline{0.55} & \underline{0.33} & 0.02 & 0.07 & 0.22 & 0.10 \\
AutoVLA ~\citep{autovla} & NeurIPS'25 & 0.25 & 0.46 & 0.73 & 0.48 & 0.07 & 0.07 & 0.26 & 0.13 \\
AutoDrive-R$^2$~\citep{yuan2026autodrive} & ICLR'26 & 0.35 & 0.49 & 0.62 & 0.49 & 0.13 & 0.44 & 0.87 & 0.48 \\
AutoDrive-P$^3$~\citep{autodrive_p3} & ICLR'26 & 0.16 & 0.31 & 0.56 & 0.34 & \textbf{0.00} & \underline{0.04} & \underline{0.20} & \underline{0.08} \\
\midrule
\textbf{AUGUstus (Ours, 1-step denoising)} & - & \textbf{0.14} & \textbf{0.24} & \textbf{0.42} & \textbf{0.27} & \underline{0.01} & \textbf{0.02} & \textbf{0.13} & \textbf{0.05} \\
\bottomrule
\end{tabular}}
\end{table}

\begin{table}[t]
\centering

\begin{minipage}[t]{0.48\textwidth}
\centering
\captionof{table}{\textbf{In-house 1,000 km evaluation.}}
\label{tab:internal}
\small
\setlength{\tabcolsep}{1.8pt}

\resizebox{\linewidth}{!}{\begin{tabular}{@{}lrrr@{}}
\toprule
\textbf{Category} & \textbf{VLM+Flow} & \textbf{VLM+Reg.} & \cellcolor{tablered}\textbf{AUGUstus} \\
\midrule
Safety     & +18.4\% & +15.1\% & \cellcolor{tablered}\textbf{0.0\%} \\
Comfort    & +8.2\%  & +8.2\%  & \cellcolor{tablered}\textbf{0.0\%} \\
Efficiency & +3.0\%  & +3.9\%  & \cellcolor{tablered}\textbf{0.0\%} \\
Compliance & \textbf{-0.5\%} & +0.1\% & \cellcolor{tablered}0.0\% \\
\bottomrule
\end{tabular}}
\end{minipage}\hfill \begin{minipage}[t]{0.48\textwidth}
\centering
\captionof{table}{\textbf{Inference time.}}
\label{tab:latency}
\small
\setlength{\tabcolsep}{3.4pt}

\resizebox{\linewidth}{!}{\begin{tabular}{@{}lr@{}}
\toprule
\textbf{Module/system} & \textbf{Time (ms)} \\
\midrule
Vision Encoder Stage & 40.2 \\
VLA Stage (1-step denoising) & 33.2 \\
\rowcolor{tablered}
\textbf{Total (Locally, no CoT)} & \textbf{73.6 (13.6\,Hz)} \\
\midrule
\rowcolor{tablered}
\textbf{Real vehicle total} & \textbf{83.0 (12.0\,Hz)} \\
\bottomrule
\end{tabular}}
\end{minipage}
\end{table}

\subsection{Ablation Study}

\begin{table}[t]
\centering
\caption{\textbf{Ablation Study of CoT and AUGUstus-mode.}}
\label{tab:cot_main}
\small
\setlength{\tabcolsep}{3.2pt}
\resizebox{\linewidth}{!}{\begin{tabular}{l >{\columncolor{tablered}}c c c c c c c c c}
\toprule
\textbf{Setting} & \multicolumn{1}{>{\columncolor{tablered}}c}{\textbf{PDMS$\uparrow$}} & \textbf{NC$\uparrow$} & \textbf{DAC$\uparrow$} & \textbf{DDC$\uparrow$} & \textbf{EP$\uparrow$} & \textbf{TTC$\uparrow$} & \textbf{Comf.$\uparrow$} & \textbf{ADE (m)$\downarrow$} & \textbf{Time (ms)$\downarrow$} \\
\midrule
No CoT, Only Action & 89.2 & 98.8 & 96.8 & 98.4 & 83.5 & 95.8 & \textbf{100.0} & 0.530 & \textbf{73.60} \\
CoT first, Action second & \textbf{90.6} & 99.1 & 97.8 & \textbf{98.6} & \textbf{84.2} & \textbf{96.6} & \textbf{100.0} & 0.521 & 4587 \\
\textbf{AUGUstus-mode (Ours)} & \textbf{90.6} & \textbf{99.0} & \textbf{97.9} & 98.5 & \textbf{84.2} & 96.5 & \textbf{100.0} & \textbf{0.520} & \textbf{73.60} \\
\bottomrule
\end{tabular}}
\end{table}

\begin{table}[t]
\centering
\caption{\textbf{Ablation Study of AUGUstus-GRPO.}}
\label{tab:rl_main}
\small
\setlength{\tabcolsep}{3.0pt}
\resizebox{0.9\linewidth}{!}{\begin{tabular}{l >{\columncolor{tablered}}c c c c c c}
\toprule
\textbf{Setting} & \multicolumn{1}{>{\columncolor{tablered}}c}{\textbf{PDMS$\uparrow$}} & \textbf{NC$\uparrow$} & \textbf{DAC$\uparrow$} & \textbf{EP$\uparrow$} & \textbf{TTC$\uparrow$} & \textbf{Comf.$\uparrow$} \\
\midrule
SFT with AUGUstus-mode  & 90.6 & \textbf{99.0} & 97.9 & 84.2 & 96.4 & \textbf{100.0} \\
+ AUGUstus-GRPO Stage I  & 94.0 & 98.6 & 98.7 & 91.4 & 96.6 & \textbf{100.0} \\
+ AUGUstus-GRPO Stage II  & \textbf{94.6} & 98.9 & \textbf{98.9} & \textbf{91.9} & \textbf{97.1} & \textbf{100.0} \\
+ JaT-GRPO, without CoT-GRPO  & 92.6 & 98.0 & 97.6 & 91.8 & 94.1 & \textbf{100.0} \\
\bottomrule
\end{tabular}}
\end{table}

\begin{figure}[t]
\centering
\includegraphics[width=\linewidth]{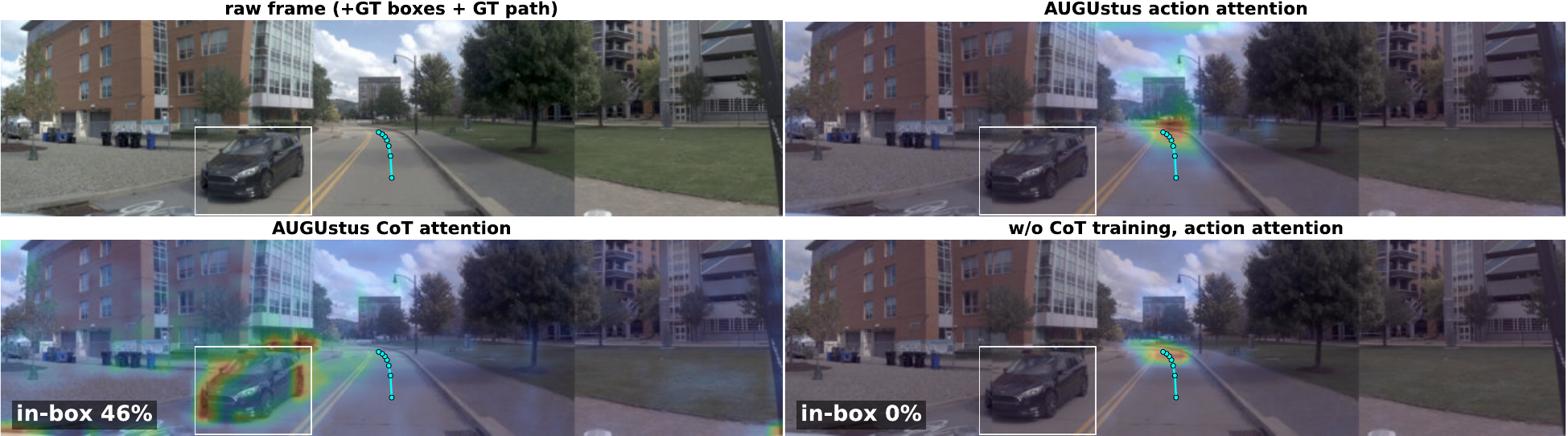}
\caption{\textbf{CoT attention visualization.} CoT supervision anchors the action-relevant attention on the nearby risk object (46\% in-box), whereas the no-CoT model misses it (0\%).}
\label{fig:cot_ablation}
\end{figure}

\begin{figure*}[t]
\centering
\begin{minipage}[t]{0.49\textwidth}
\centering
\includegraphics[width=\linewidth]{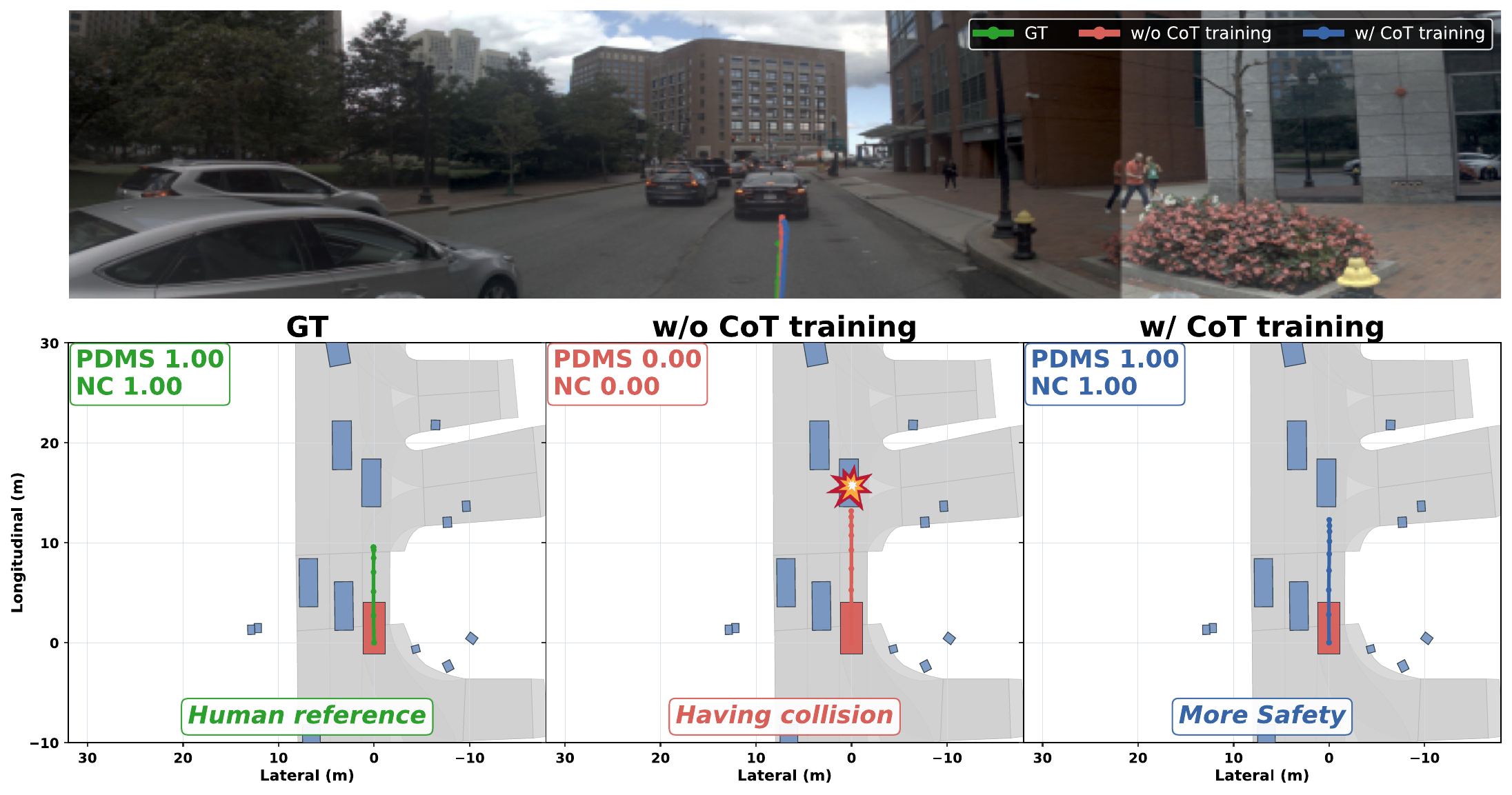}
\end{minipage}\hfill
\begin{minipage}[t]{0.49\textwidth}
\centering
\includegraphics[width=\linewidth]{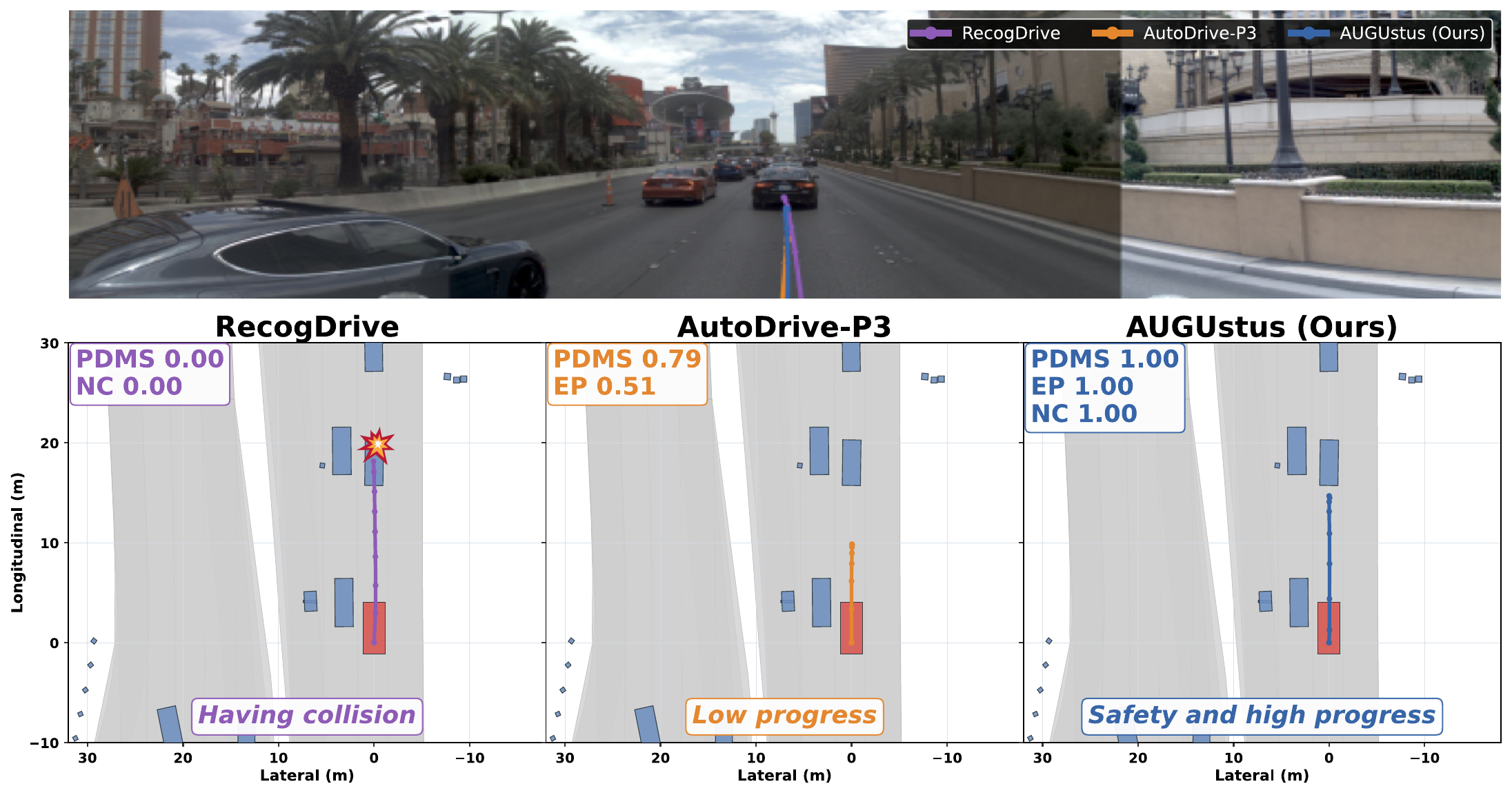}
\end{minipage}

\caption{\textbf{Visualization.} \emph{Left:} Trajectory visualization with and without CoT. \emph{Right:} Trajectory visualization compared with other methods.}
\label{fig:qualitative_ablation}
\end{figure*}

\begingroup
\titleformat*{\subsubsection}{\normalsize\sffamily\mdseries}
\subsubsection{Ablation Study of AUGUstus-mode and CoT}
\label{sec:ablation_cot}
As shown in Table~\ref{tab:cot_main}, we ablate CoT and \oursmode on SFT with NAVSIMv1. With \oursmode, adding CoT to training makes driving safer and further improves performance. Compared with CoT-first, \oursmode achieves comparable performance while preserving real-time inference, making the architecture deployable. Overall, our proposed \oursmode and unified CoT training prove effective.

\subsubsection{Ablation Study of AUGUstus-GRPO}
\label{sec:ablation_rl}
As shown in Table~\ref{tab:rl_main}, with our proposed \oursgrpo, jointly optimized with JaT-GRPO and CoT-GRPO, stage I improves PDMS by +4 and stage II further improves it by +0.6, achieving a SOTA PDMS of 94.6. These results demonstrate the effectiveness of our proposed unified \oursgrpo algorithm.
\endgroup

\subsection{Visualization}
As shown in Fig.~\ref{fig:qualitative_ablation}, the no-CoT model fails to yield to the nearby agent and collides, whereas CoT supervision produces a safer trajectory. Compared with other methods on the same scene, \ours avoids collision while maintaining high progress.

\section{Conclusion and Future Work}
\label{sec:conclusion}
In this work, we propose \ours, a deployable unified VLA that integrates diffusion-based action generation and autoregressive CoT understanding in a single transformer, with \oursmode ordering for low-latency inference and \oursgrpo for joint optimization. Experiments on NAVSIMv1/v2 and nuScenes show state-of-the-art performance: \ours achieves 94.6 PDMS on NAVSIMv1 and 90.7 EPDMS on NAVSIMv2 without a scorer module, and on nuScenes it achieves an average L2 error of 0.27\,m and an average collision rate of 0.05\%, both the best among the compared methods. To validate its practicality, we train it on our in-house dataset with a large-scale GPU cluster, then evaluate it on an internal driving dataset and preliminarily deploy it on a real vehicle, achieving real-time operation. In future work, we will conduct more extensive real-vehicle tests to further validate the architecture and attempt to apply it to robot VLA.

\subsection*{Acknowledgments}
Yuqi Ye completed this project as the first author during an internship at XPeng. We extend special thanks to Caigao Jiang, the project lead, for his leadership, invaluable guidance, and sustained support throughout the project. We also thank Zhuonan Wang, Ph.D., and Prof.~Chao Qu for their insightful advice, Congyuan Pei for his assistance with on-vehicle deployment, and Prof.~Wei Gao for academic mentorship. We appreciate the contributions of all other collaborators. This work is solely for non-commercial academic research.

\bibliographystyle{assets/plainnat}
\bibliography{paper}

\newpage
\appendix

\section{Supplementary Architecture Details}
\label{app:architecture}

\subsection{Backbone and action projections}

\textbf{Action-module data path.} Given a normalized noisy trajectory \(x_t\in\mathbb{R}^{8\times3}\) and flow time \(t\in[0,1]\), the module produces a clean trajectory estimate \(\hat{x}_0\in\mathbb{R}^{8\times3}\) through three stages: an Action Projector, the shared transformer action-token block, and an Action Decoder. The eight rows always correspond to the eight future waypoints; all linear maps are shared across waypoint positions, while positional information is carried by the backbone's continued token positions.

\begin{figure}[t]
\centering
\includegraphics[width=\textwidth]{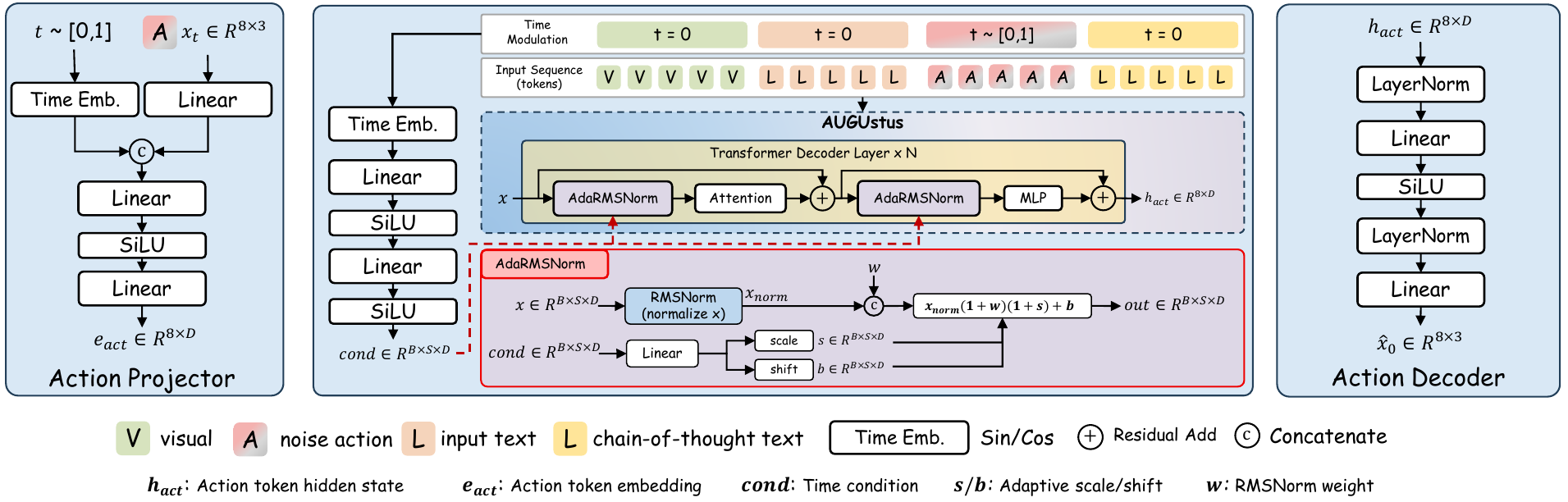}
\caption{\textbf{Detailed AUGUstus module.} \emph{Left: Action Projector.} A normalized noisy trajectory \(x_t\in\mathbb{R}^{8\times3}\) and a sinusoidal time embedding are projected, concatenated, and passed through Linear--SiLU layers to form \(e_{\mathrm{act}}\in\mathbb{R}^{8\times D}\). \emph{Center: shared decoder.} Visual tokens \(V\), input text \(L\), noisy action tokens \(A\), and CoT text share the transformer; the action block is bidirectional, and the segment-specific time modulation uses \(t{=}0\) for context/CoT and the sampled \(t\) for action tokens. Each layer contains attention, MLP, residual additions, and AdaRMSNorm. The inset expands the adaptive scale/shift \(s,b\) and the pretrained RMSNorm weight \(w\). \emph{Right: Action Decoder.} LayerNorm--Linear--SiLU--LayerNorm--Linear decoding maps \(h_{\mathrm{act}}\in\mathbb{R}^{8\times D}\) to \(\hat{x}_0\in\mathbb{R}^{8\times3}\).}
\label{fig:model}
\end{figure}

\textbf{Action Projector.} In the left block of Fig.~\ref{fig:model}, each waypoint is first mapped from three coordinates to the hidden width by a linear layer. In parallel, the scalar flow time is encoded with a sinusoidal sine/cosine embedding. The two representations are concatenated row-wise and passed through a Linear--SiLU--Linear MLP, yielding action-token embeddings \(e_{\mathrm{act}}\in\mathbb{R}^{8\times D}\). Thus the time condition is available at the input of every action token without changing the dimensionality of the pretrained language-vision stream.

\textbf{Shared action-token block.} The embeddings \(e_{\mathrm{act}}\) are inserted as an eight-token action block after the multimodal context. In AUGUstus-mode, the action block reads the cached video, ego-state, route, and language context and is bidirectional within itself; the subsequent causal CoT block may read the action states, but action tokens cannot read CoT tokens. The action states therefore receive scene information from the shared trunk while remaining independent of the optional autoregressive decoding path. During training, this is also the route by which the CoT loss can shape the action representation; during deployment, only the action block is evaluated.

\textbf{Segmented input sequence and transformer layer.} The center of Fig.~\ref{fig:model} expands one forward pass into its token segments: visual tokens \(V\), input-language tokens \(L\), noisy action tokens \(A\), and causal chain-of-thought tokens \(L\). The action segment is a single bidirectional block, while the language and CoT segments retain their autoregressive masks. Every decoder layer applies the same residual pattern, \(\mathrm{AdaRMSNorm}\!\to\!\mathrm{Attention}\!\to\!\mathrm{Add}\!\to\!\mathrm{AdaRMSNorm}\!\to\!\mathrm{MLP}\!\to\!\mathrm{Add}\), to the shared hidden states. The time condition is segment-specific: visual and input-language tokens use \(t{=}0\), the noisy action tokens use the sampled \(t\in[0,1]\), and the CoT tokens again use \(t{=}0\). This keeps the diffusion state from contaminating the language stream while allowing every action layer to adapt to the denoising point.

\textbf{Action Decoder.} The right block of Fig.~\ref{fig:model} maps the final action hidden states \(h_{\mathrm{act}}\in\mathbb{R}^{8\times D}\) back to waypoint coordinates. It applies LayerNorm, a linear projection, SiLU, a second LayerNorm, and a final linear projection to three channels, producing \(\hat{x}_0\in\mathbb{R}^{8\times3}\). The output is inverse-normalized to metric \(x\), \(y\), and heading coordinates before being sent to the planner. At \(t=1\), the \(x_0\)-parameterization makes this decoder output the deployed one-step trajectory directly; for a multi-step ODE/SDE rollout, the output is re-embedded as the next noisy action state and the same shared module is applied again. No separate action transformer or scorer is introduced.

Flow time is injected with per-segment AdaRMSNorm. For hidden state \(h\) and condition \(q\), we use
\begin{equation}
\operatorname{AdaRMSNorm}(h;q)=\operatorname{RMSNorm}(h)\odot(1+w)\odot(1+s(q))+b(q),
\qquad [s(q),b(q)]=W_q q,
\label{eq:appada}
\end{equation}
where \(w\) is the pretrained RMSNorm weight. Context, language, and boundary tokens use a zero-time condition, while action tokens use the sampled flow time. The positional indices continue from the multimodal context into the action and CoT blocks, so the pretrained positional statistics are preserved.

\section{Algorithmic Details for JaT-GRPO}
\label{app:derivation}
\label{app:proofs}

\subsection{Reverse-time SDE kernel in x0-parameterization}

With the convention $t{=}1$ pure noise, $t{=}0$ data, the forward CFM interpolation $x_t=(1-t)x_0+t\varepsilon$ defines the probability-flow ODE $\dot x_t = v_\theta(x_t,t)$ in the direction of decreasing $t$. Adding Langevin diffusion with coefficient $g(t)$ (the Flow-GRPO construction~\citep{flow_grpo}) gives the reverse-time SDE; its Euler--Maruyama step of size $\Delta t>0$ (from $t$ to $t-\Delta t$) is
\begin{equation}
x_{t-\Delta t}=x_t-\Big[\,v_\theta+\frac{g(t)^2}{2t}\big(x_t+(1-t)v_\theta\big)\Big]\Delta t+g(t)\sqrt{\Delta t}\,z,
\label{app:sde}
\end{equation}
where $-\frac{1}{t}(x_t+(1-t)v)$ is the score $\nabla\log p_t$ of the linear interpolation path. Substituting the $x_0$-parameterized velocity $v_\theta=(x_t-\hat x_0^\theta)/t$ and collecting $x_t$ and $\hat x_0^\theta$ yields the kernel of Eq.~(\ref{eq:kernel}) with
\[
c_x=1-\frac{\Delta t}{t}\Big(1+\frac{g(t)^2}{2t}\Big),\qquad
c_0=\frac{\Delta t}{t}\Big(1+\frac{g(t)^2\,(1-t)}{2t}\Big),\qquad
\sigma_s=g(t)\sqrt{\Delta t}.
\]
For $g(t)^2=\lambda t$ these reduce to $c_x=1-\frac{\Delta t}{t}(1+\frac\lambda2)$, $c_0=\frac{\Delta t}{t}(1+\frac{\lambda(1-t)}2)$, $\sigma_s=\sqrt{\lambda t\Delta t}$; at $t{=}1,\Delta t{=}1$: $c_x=-\lambda/2$, $c_0=1$, $\sigma_s=\sqrt\lambda$, hence Eq.~(\ref{eq:onestep}).

The action-transition schedules used by GRPO are summarized in Table~\ref{tab:schedules}. The proposed schedule is finite at the pure-noise endpoint and vanishes at the clean endpoint, so it supports both one-step deployment and a clean terminal trajectory. The schedule enters the exact ratio and KL only through $c_0$ and $\sigma_s$, so the same implementation covers all three alternatives. 

\begin{table}[t]
\caption{\textbf{Kernel coefficients per schedule} (Eq.~(\ref{eq:kernel}); one reverse step $t\to t-\Delta t$). The first row names each method and the second row gives its diffusion schedule $g(t)^2$; the remaining rows give the induced kernel coefficients and endpoint behavior. $\dagger$Flow-GRPO implementations substitute a clamp $\sigma_{\max}$ for $\sigma_s$ near $t{=}1$.}
\label{tab:schedules}
\label{tab:coefficients}
\centering\small
\setlength{\tabcolsep}{3.4pt}
\resizebox{\linewidth}{!}{\begin{tabular}{lccc}
\toprule
 & Flow-GRPO~\citep{flow_grpo} & DanceGRPO~\citep{dance_grpo} & JaT-GRPO (ours) \\
\midrule
$g(t)^2$ & $a^2\tfrac{t}{1-t}$ & $\eta^2$ & $\lambda t$ \\
\midrule
$c_x$ & $1-\frac{\Delta t}{t}\big(1+\frac{a^2}{2(1-t)}\big)$ & $1-\frac{\Delta t}{t}\big(1+\frac{\eta^2}{2t}\big)$ & $1-\frac{\Delta t}{t}\big(1+\frac{\lambda}{2}\big)$ \\
$c_0$ & $\frac{\Delta t}{t}\big(1+\frac{a^2}{2}\big)$ & $\frac{\Delta t}{t}\big(1+\frac{\eta^2(1-t)}{2t}\big)$ & $\frac{\Delta t}{t}\big(1+\frac{\lambda(1-t)}{2}\big)$ \\
$\sigma_s$ & $a\sqrt{\frac{t\,\Delta t}{1-t}}$ & $\eta\sqrt{\Delta t}$ & $\sqrt{\lambda t\,\Delta t}$ \\
\midrule
$t{=}1,\ \Delta t{=}1$ & $c_x\!\to\!-\infty$, $\sigma_s\!\to\!\infty$ $^\dagger$ & $c_x{=}-\frac{\eta^2}{2},\ c_0{=}1,\ \sigma_s{=}\eta$ & $c_x{=}-\frac{\lambda}{2},\ c_0{=}1,\ \sigma_s{=}\sqrt\lambda$ \\
single-step policy & \emph{undefined} & $\mathcal N(\hat x_0-\frac{\eta^2}{2}x_1,\eta^2I)$ & $\mathcal N(\hat x_0-\frac{\lambda}{2}x_1,\lambda I)$ \\
terminal state $t{=}0$ & clean & $+\mathcal N(0,\eta^2\Delta t\,I)$ corruption & clean \\
\bottomrule
\end{tabular}}
\vspace{-4pt}
\end{table}

\subsection{Closed-form log-ratio and KL}

Both kernels share variance $\sigma_s^2 I$, so with $z=(x_{t-\Delta t}-\mu_{\mathrm{old}})/\sigma_s$:
\begin{align}
\log\rho_t &= -\frac{1}{2\sigma_s^2}\Big(\big\|x_{t-\Delta t}-\mu_\theta\big\|^2-\big\|x_{t-\Delta t}-\mu_{\mathrm{old}}\big\|^2\Big)\\
&= \frac{1}{\sigma_s^2}\langle \mu_\theta-\mu_{\mathrm{old}},\,x_{t-\Delta t}-\mu_{\mathrm{old}}\rangle-\frac{1}{2\sigma_s^2}\big\|\mu_\theta-\mu_{\mathrm{old}}\big\|^2
= \frac{c_0}{\sigma_s}\langle \delta,z\rangle-\frac{c_0^2}{2\sigma_s^2}\|\delta\|^2 .
\label{eq:logratio}
\end{align}
Since $\mathbb E[z]=0$, $\mathbb E[\log\rho_t]=-\frac{c_0^2}{2\sigma_s^2}\|\delta\|^2=-\mathrm{KL}(\pi_{\mathrm{old}}\|\pi_\theta)$ (equal-covariance Gaussian KL). The same derivation applies per coordinate $d$ when $\sigma_s$ and $z$ are diagonal per-dimension quantities -- the form used in our implementation with multivariate noise levels $\lambda_{x,y,\psi}$.

The corresponding reference-policy KL and deployed one-step policy are
\begin{equation}
\mathrm{KL}_{\mathrm{act}}=\frac{c_0^2}{2\sigma_s^2}\|\hat x_0^\theta-\hat x_0^{\mathrm{ref}}\|_2^2.
\label{eq:kl}
\end{equation}
\begin{equation}
\pi_\theta(x_0\mid x_1,c)=\mathcal N\!\left(\hat x_0^\theta-\frac{\lambda}{2}x_1,\lambda I\right)\quad(t=1,\Delta t=1).
\label{eq:onestep}
\end{equation}

\textbf{Scalar parametric form.} Writing $m=\delta/t$ (the velocity-space prediction difference) and defining the schedule coefficient $\kappa\triangleq c_0\,t/\sigma_s$, Eq.~(\ref{eq:logratio}) factorizes per element into the compact quadratic form
\[
\log\rho=\kappa\langle m,z\rangle-\tfrac12\kappa^2\|m\|^2,\qquad \mathrm{KL}=\tfrac12\kappa^2\|m\|^2,\qquad \rho\le\exp\big(\|z\|^2/2\big),
\]
with equality of the bound at $m=-z/\kappa$ -- the quadratic dependence of $\rho_{\max}$ on the exploration noise is a second, independent motivation for two-sided PPO clipping. For the constant-$\eta$ schedule, $\kappa=\sqrt{\Delta t}\,(\,\eta^{-1}+\tfrac{\eta(1-t)}{2t}\,)$; at $t{=}1$ this stays finite ($\kappa=\sqrt{\Delta t}/\eta$).

\subsection{Gradient decomposition and comparison}

Writing $\phi(\theta)\triangleq\log\rho_t$ and differentiating Eq.~(\ref{eq:logratio}):
\begin{equation}
\nabla_\theta\phi = \frac{c_0}{\sigma_s}\Big(\langle \nabla_\theta\hat x_0^\theta,\,z\rangle-\frac{c_0}{\sigma_s}\langle\nabla_\theta\hat x_0^\theta,\,\delta\rangle\Big),
\end{equation}
i.e.\ an advantage-weighted \emph{noise-following} term (move $\hat x_0$ along $z$ when $A_i>0$) and a \emph{trust-region-like} term pulling $\hat x_0$ back toward $\hat x_0^{\mathrm{old}}$, which vanishes at the behavior policy; on-policy, $\mathbb E_z\nabla\phi=A_i\frac{c_0}{\sigma_s}z$, equivalent to minimizing $\mathcal L_{\mathrm{equiv}}=-\sum_t A_i\frac{c_0}{\sigma_s}\langle z,\hat x_0^\theta\rangle$. This contrasts with DiffusionNFT~\citep{diffusion_nft}, whose gradient follows the \emph{velocity difference} direction with a built-in trust region of radius $\propto A_i/\beta$; Flow-GRPO coincides direction-wise with our first term but has no closed-form KL and a diverging $t{=}1$ schedule.

\subsection{Necessity of \texorpdfstring{$\lambda>0$}{lambda>0} for Single-Step GRPO}

When \(N=1\), all rollouts in a group share the same initial condition \(x_1\), so they can differ only in the noise variable \(z\). If \(\lambda=0\), the kernel collapses to a point mass. Consequently, all \(G\) samples are identical, \(\mathrm{std}_G(R)=0\), and the advantages are undefined or zero. Therefore, \(\lambda>0\) is necessary. Sufficiency follows because \(\mathcal N(\mu,\lambda I)\) has full support.

\subsection{Exploration Noise Engineering}
\label{app:noise}

\textbf{Per-dimension levels and trajectory smoothing.} Exploration noise is applied in the quantile-normalized action space, with per-dimension levels \((\lambda_x,\lambda_y,\lambda_\psi)\). The injected noise is smoothed along the trajectory, keeping perturbed trajectories kinematically plausible. The group shares the initial \(x_1\) state so that advantages compare like-for-like trajectories.

\textbf{$\sigma$-sensitivity of heteroscedastic noise.} For fixed prediction shift $\delta_d$ in dimension $d$, Eq.~(\ref{eq:kl}) gives per-step KL $\propto\delta_d^2\lambda_d^{-1}$, while the induced update drift of dimension $d$ scales as $\nabla\propto\delta_d\lambda_d^{-1}$ (gradient) and the post-update KL as $\propto\lambda_d^{-2}$; with per-dimension noise levels spanning ${\sim}2.5\times$ and time-scaled reweighting across waypoints, the effective step-size ratio across dimensions blows up to ${\sim}10^3$ in drift terms, collapsing the effective sample size of the group to a few near-deterministic slots. As a result, we set noise level $5{\times}10^{-2}$ and groups with $G{=}8$.

\section{More detail about CoT}
\label{app:cot}

This appendix complements the main-text CoT ablation with the complete CoT-content metrics and supervision templates (Table~\ref{tab:cot}) and additional CoT-vs-no-CoT attention statistics. To avoid duplicate reporting, the metric table and the format table are merged: each CoT variant is listed once with its supervision template and full PDMS sub-metric breakdown.

\begin{table*}[t]
\centering
\caption{\textbf{CoT content ablation (complete metrics and supervision formats).} All variants share the Section~\ref{sec:sftproto} protocol and are evaluated with the deterministic one-step evaluator on the full navtest split; only the CoT content changes. PDMS is reported together with its full sub-metric breakdown: NC, DAC, DDC, EP, TTC, Comfort, and ADE. NC: no-collision, DAC: drivable-area compliance, DDC: driving-direction compliance, EP: ego progress, TTC: time-to-collision, Comfort: comfort, ADE: average L2 displacement. The default prediction CoT (risk-object boxes + future positions) is \textbf{bolded}; the last row is the CoT-first ordering control (identical recipe and prediction content, reverse token order). Inference time is reported once for the shared 73.60\,ms action-only path and separately for the CoT-first path.}
\label{tab:cot}
\centering
\small
\setlength{\tabcolsep}{1.7pt}
\renewcommand{\arraystretch}{1.15}
\resizebox{\textwidth}{!}{\begin{tabular}{@{}
>{\centering\arraybackslash}m{0.10\textwidth}
>{\centering\arraybackslash}m{0.20\textwidth}
c c c c c c c c c@{}}
\toprule
\textbf{CoT type} & \textbf{Template} & \cellcolor{tablered}\textbf{PDMS}$\uparrow$ & \textbf{NC}$\uparrow$ & \textbf{DAC}$\uparrow$ & \textbf{DDC}$\uparrow$ & \textbf{EP}$\uparrow$ & \textbf{TTC}$\uparrow$ & \textbf{Comf.}$\uparrow$ & \textbf{ADE}$\downarrow$ & \textbf{Time (ms)} \\
\midrule

No CoT & action-only training & \cellcolor{tablered}89.2 & 98.8 & 96.8 & 98.4 & 83.5 & 95.8 & 100.0 & 0.530 & \\

\cmidrule{1-10}

Planning & \texttt{slow, keep lane, pass after crossing} & \cellcolor{tablered}90.1 & 98.9 & 97.4 & 98.5 & 83.8 & 96.1 & 100.0 & 0.522 & \\

\cmidrule{1-10}

Reasoning & perception + future bins + maneuver & \cellcolor{tablered}89.6 & 98.7 & 96.9 & 98.4 & 83.4 & 96.1 & 100.0 & 0.534 & \\

\cmidrule{1-10}

Perception & \texttt{car [x1,y1,x2,y2]; ped. [x1,y1,x2,y2]} & \cellcolor{tablered}90.3 & 98.9 & 97.5 & \textbf{98.6} & \textbf{84.3} & 96.1 & 100.0 & 0.517 & \textbf{73.60} \\

\cmidrule{1-10}

Perception + Prediction & boxes + future bins, no maneuver sentence & \cellcolor{tablered}90.3 & 98.9 & 97.6 & \textbf{98.6} & 84.1 & 96.3 & 100.0 & \textbf{0.513} & \\

\cmidrule{1-10}

Short-prediction & atomic vocabulary encoding & \cellcolor{tablered}89.1 & 98.7 & 96.7 & 98.3 & 83.1 & 95.4 & 100.0 & 0.559 & \\

\cmidrule{1-10}

\textbf{Prediction} & \texttt{car [box] -> near-front; ped. [box] -> far-left} & \cellcolor{tablered}\textbf{90.6} & \textbf{99.0} & \textbf{97.9} & 98.5 & 84.2 & \textbf{96.5} & \textbf{100.0} & 0.520 & \\

\cmidrule{1-11}

\quad CoT first & same as Prediction, CoT tokens precede action & \cellcolor{tablered}90.6 & 99.1 & 97.8 & 98.6 & 84.2 & 96.6 & 100.0 & 0.521 & 4597 \\

\bottomrule
\end{tabular}}
\end{table*}

\textbf{CoT content (Table~\ref{tab:cot}).} Removing CoT costs 1.4 PDMS (89.2 vs.\ 90.6), concentrated in drivable-area compliance ($-1.1$ DAC) and ego progress ($-0.7$ EP). This ablation directly tests the VLA-to-VA hypothesis: CoT is useful only when its content is action-relevant and its supervision reaches the shared action representation. Among CoT types, \emph{prediction} beats \emph{perception} ($+0.3$), \emph{planning} ($+0.5$), and their union ($+0.3$): future-position labels carry the operative information, and stacking perception on top adds tokens without information. Free-form \emph{reasoning} is worse (89.6), and a 2573-token-vocabulary atomic \emph{short-prediction} encoding (${\sim}14$ tokens per scene, $7\times$ shorter than the default) even falls below the no-CoT baseline (89.1 vs.\ 89.2) -- aggressive CoT compression trades away the planning gain entirely. Order analysis confirms the ``pre $>$ per/perpre $>$ pla'' ranking. The CoT-first ordering control (last row) reaches the same 90.6 PDMS as the action-first default, confirming that the quality gain comes from the CoT supervision itself rather than the ordering; the ordering decides only the deployment cost, as reported in Fig.~\ref{fig:cot_ablation} (left).

\subsection{CoT content and supervision formats}
\label{app:cot_query_loss}

\subsubsection{What each CoT teaches}
Our CoT ablation is a comparison of three \emph{alternative} supervision targets, not a three-stage pipeline: every variant is trained on the identical scene and action, and only the content of the single CoT block differs. We take the three task-oriented CoT \emph{definitions} from AutoDrive-P$^3$~\citep{autodrive_p3} -- \emph{Perception CoT} (the 2-D boxes of relevant agents), \emph{Prediction CoT} (each box plus a discretized future-position label, such as \texttt{near-front} or \texttt{far-left}), and \emph{Planning CoT} (textual ego waypoints or maneuvers) -- and ask which one makes the action head plan best, since only one of them is emitted per sample. We do not chain them, and the model never consumes its own perception output to produce a prediction or a plan; the three streams are competing ways of describing the same scene, and Table~\ref{tab:cot} reports the outcome (prediction CoT wins). A fourth variant, \emph{reasoning CoT}, concatenates the three signals into a compact approximately 100-token explanation: it first names the relevant agents, then states their future bins, and finally gives the resulting ego maneuver. This is intentionally shorter and more structured than unconstrained free-form reasoning.

\begin{figure*}[t]
\centering
\includegraphics[width=0.92\textwidth]{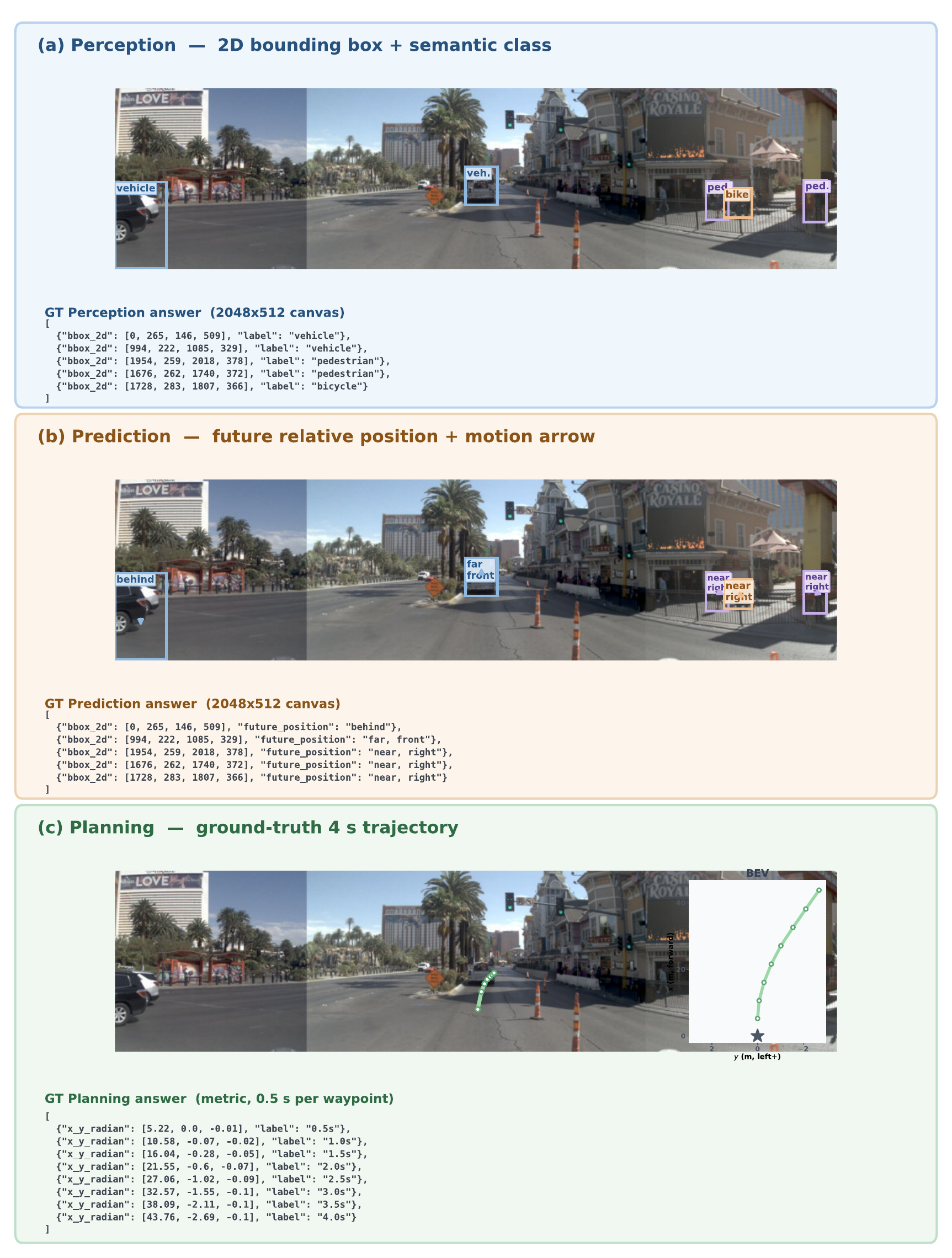}
\caption{\textbf{The three CoT contents compared on one NAVSIM right-turn scene (ground-truth annotations).} The three rows are three \emph{alternative} supervision targets, not a serial pipeline: each is trained alone on the same scene and action, and the model never consumes one row's output to produce another's. One pastel outer frame encloses the title, annotated scene, and JSON answer of each CoT variant. \textbf{(a)~Perception CoT}: a 2-D box and a class label for every salient agent (vehicle: blue, bicycle: orange, pedestrian: violet) on the $2048\times512$ deployment canvas. \textbf{(b)~Prediction CoT}: each box is annotated with its future relative position and a motion arrow; this is the default target. \textbf{(c)~Planning CoT}: the ground-truth $4$\,s trajectory (green) projected onto the camera strip, with a BEV inset (ego-left on the left) showing the top-down right-turn arc.}
\label{fig:app_cot_demo}
\end{figure*}

Figure~\ref{fig:app_cot_demo} instantiates the three variants on one NAVSIM right-turn scene, using the ground-truth annotation of the released evaluation split. The three panels share the identical scene and the identical agent ordering, so the figure isolates exactly what each CoT content contributes. All coordinates below are reported on the deployment canvas ($2048\times512$), matching the model input coordinate system.

\textbf{(a) Perception CoT: \emph{what is where}.}
This variant emits a 2-D bounding box and a semantic class for every agent that can influence ego motion. Here the annotation contains five agents---two vehicles (one in the left lane, one far ahead in the center) and three vulnerable road users (two pedestrians and one cyclist on the right sidewalk)---each drawn with a class-colored box and a label tag in Figure~\ref{fig:app_cot_demo}(a). The complete answer field is
\begin{lstlisting}[basicstyle=\fontsize{7.2pt}{7.8pt}\ttfamily\selectfont]
<perception>[{"bbox_2d":[0,265,146,509],"label":"vehicle"},
 {"bbox_2d":[994,222,1085,329],"label":"vehicle"},
 {"bbox_2d":[1954,259,2018,378],"label":"pedestrian"},
 {"bbox_2d":[1676,262,1740,372],"label":"pedestrian"},
 {"bbox_2d":[1728,283,1807,366],"label":"bicycle"}]</perception>
\end{lstlisting}
All five boxes are the salient agents that can influence ego motion: the two vehicles lie along the ego path, and the three vulnerable road users stand on the sidewalk that the right turn sweeps past. The ego vehicle is moving at $10.35$\,m/s, so the pedestrians and the cyclist are the agents whose distance the maneuver changes most rapidly. This variant carries location but no motion: it tells the planner where the conflicts currently are, not where they will be.

\textbf{(b) Prediction CoT: \emph{where each agent will be}.}
This variant keeps the same boxes and assigns each of them a coarse future relative position over the next 4\,s (\emph{e.g.}\ \emph{near, right} / \emph{far, front} / \emph{behind}), drawn as a directional arrow and a text tag on every box in Figure~\ref{fig:app_cot_demo}(b):
\begin{lstlisting}[basicstyle=\fontsize{7.2pt}{7.8pt}\ttfamily\selectfont]
<prediction>[{"bbox_2d":[0,265,146,509],"future_position":"behind"},
 {"bbox_2d":[994,222,1085,329],"future_position":"far, front"},
 {"bbox_2d":[1954,259,2018,378],"future_position":"near, right"},
 {"bbox_2d":[1676,262,1740,372],"future_position":"near, right"},
 {"bbox_2d":[1728,283,1807,366],"future_position":"near, right"}]</prediction>
\end{lstlisting}
The boxes share the same order as the perception variant. The vehicle in the adjacent left lane drops to \emph{behind}, the lead vehicle stays \emph{far, front}, and all three vulnerable road users remain \emph{near, right}---the right-side road users are the safety-critical agents for this maneuver. This variant is the one our default model uses: it adds the future bearing of each conflict to its current location, and Table~\ref{tab:cot} shows that this extra content, and not merely the presence of language, is what lifts PDMS from $89.2$ to $90.6$.

\textbf{(c) Planning CoT: \emph{what the ego does}.}
This variant instead describes the ego maneuver itself, as eight waypoints $(x,y,\theta)$ at $0.5$\,s intervals over a $4$\,s horizon. Figure~\ref{fig:app_cot_demo}(c) projects the ground-truth trajectory (green) back onto the camera strip and shows the complete top-down arc in a BEV inset; note that the inset is drawn with ego-left on the left, so the right turn bends to the right of the star:
\begin{lstlisting}[basicstyle=\fontsize{7.2pt}{7.8pt}\ttfamily\selectfont]
<planning>[{"x_y_radian":[5.22,0.0,-0.01],"label":"0.5s"},
 {"x_y_radian":[10.58,-0.07,-0.02],"label":"1.0s"},
 {"x_y_radian":[16.04,-0.28,-0.05],"label":"1.5s"},
 {"x_y_radian":[21.55,-0.6,-0.07],"label":"2.0s"},
 {"x_y_radian":[27.06,-1.02,-0.09],"label":"2.5s"},
 {"x_y_radian":[32.57,-1.55,-0.1],"label":"3.0s"},
 {"x_y_radian":[38.09,-2.11,-0.1],"label":"3.5s"},
 {"x_y_radian":[43.76,-2.69,-0.1],"label":"4.0s"}]</planning>
\end{lstlisting}
The eight waypoints trace a gentle, high-speed right curve that advances from $(5.2,0.0)$\,m to $(43.8,-2.7)$\,m while the heading rotates from $-0.01$ to $-0.10$\,rad, keeping a lateral offset from the sidewalk agents on the right. Because this variant describes the answer rather than the evidence, it is farther from the visual input than (a) and (b): its $90.1$ PDMS sits below prediction CoT, consistent with the view that the most useful CoT content is the one that supplies the action head with the operative scene information.

Prediction CoT is the most useful target because it supplies the action head with both the current conflict location and its likely future bearing. Perception alone gives the object but not its motion, while planning text is farther from the visual evidence. Free reasoning is less stable under teacher forcing, and aggressive atomic compression removes useful spatial context. The supervision templates and the complete PDMS sub-metric breakdown (NC, DAC, DDC, EP, TTC, Comfort, and ADE) are reported together in Table~\ref{tab:cot}.

\FloatBarrier
\subsection{CoT attention visualization}
\label{app:cot_attention}

\begin{figure*}[!htbp]
\centering
\includegraphics[width=0.70\textwidth]{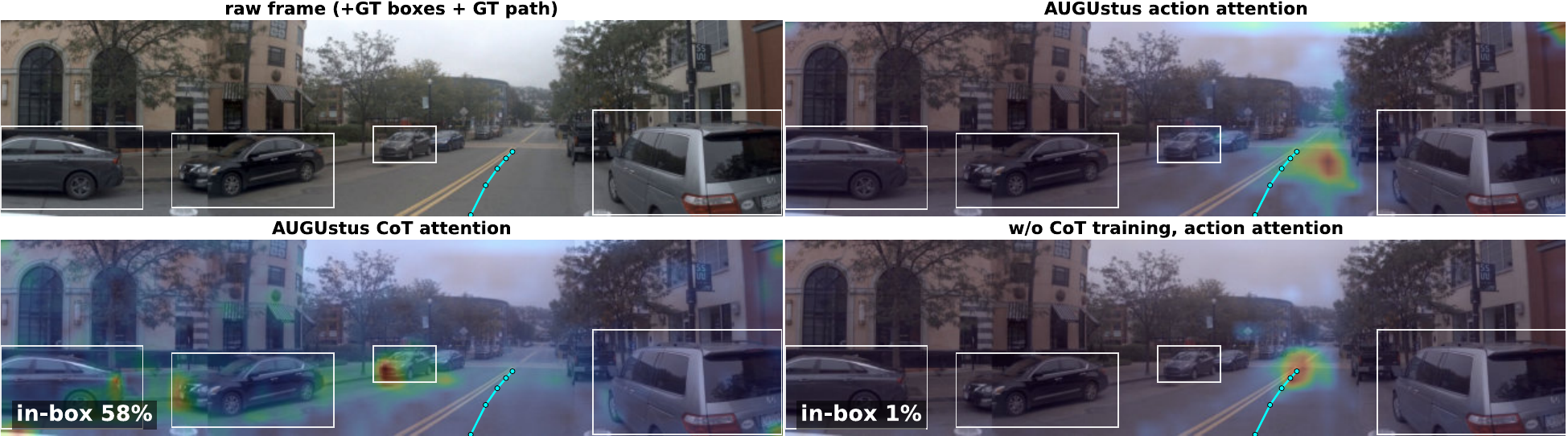}
\vspace{3pt}
\includegraphics[width=0.70\textwidth]{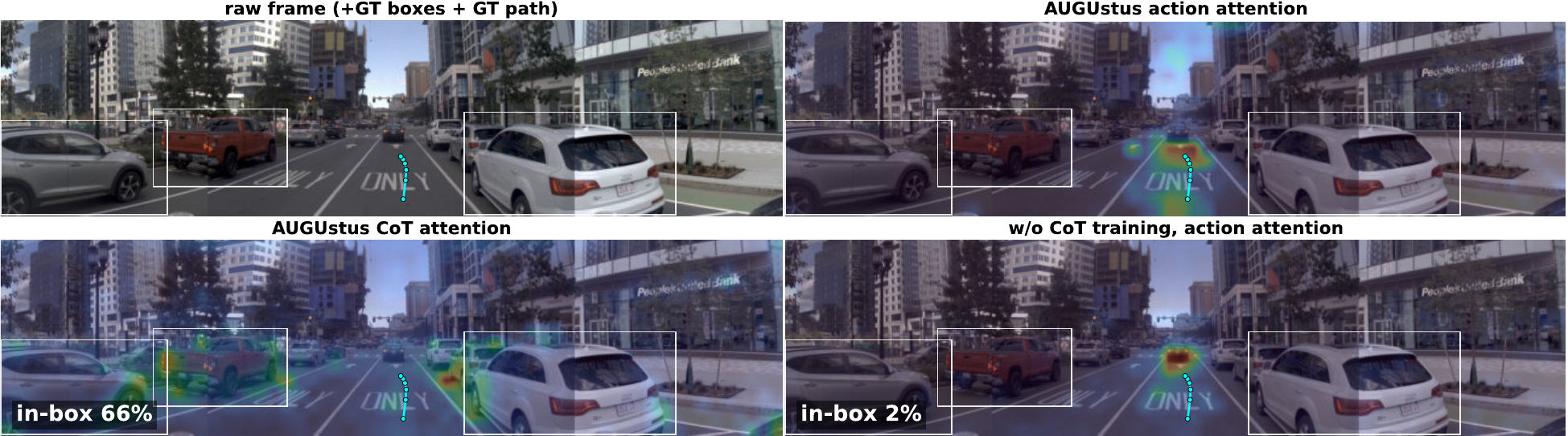}
\vspace{3pt}
\includegraphics[width=0.70\textwidth]{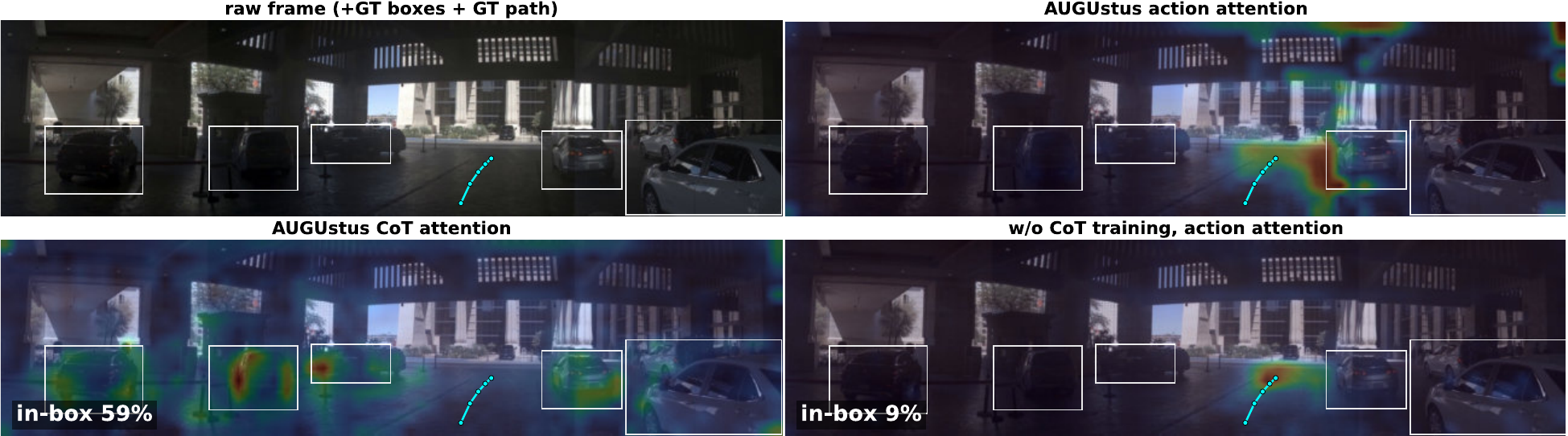}
\vspace{3pt}
\includegraphics[width=0.70\textwidth]{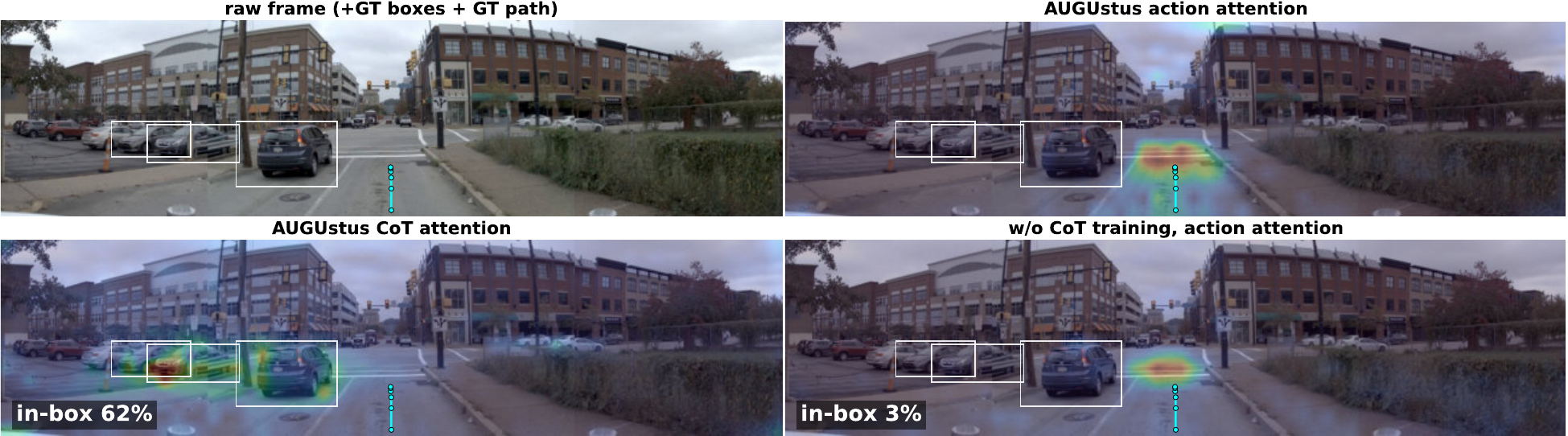}
\vspace{3pt}
\includegraphics[width=0.70\textwidth]{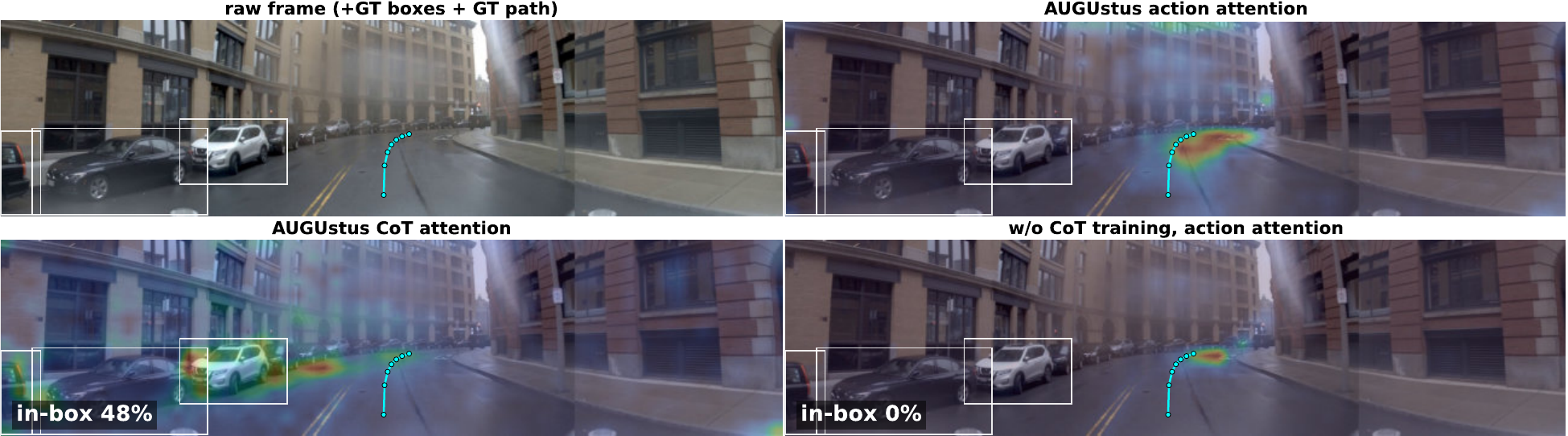}
\vspace{3pt}
\includegraphics[width=0.70\textwidth]{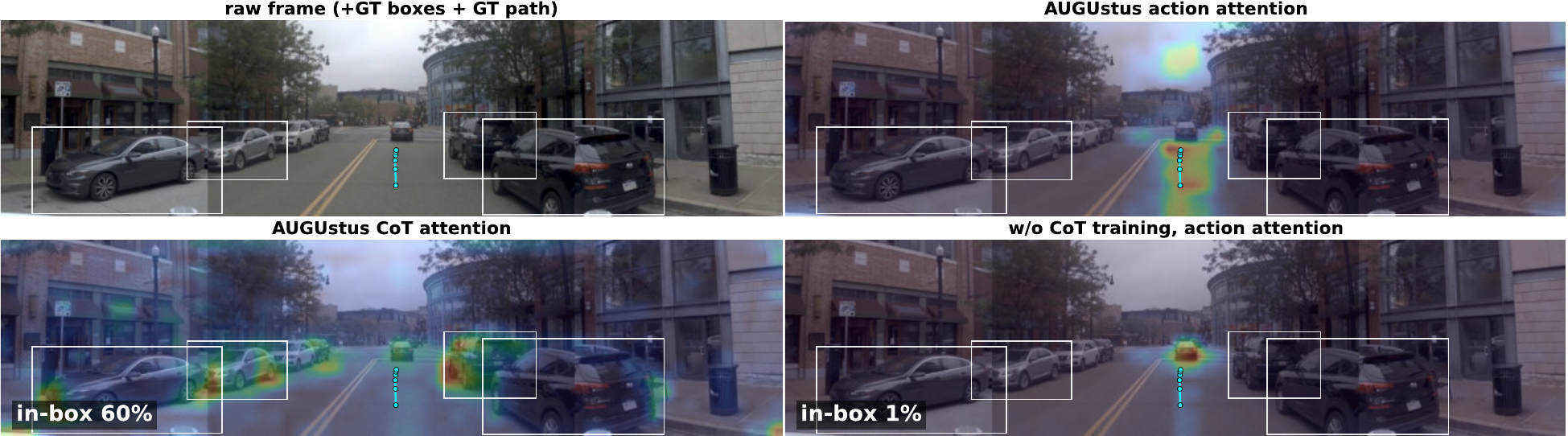}
\caption{\textbf{CoT attention visualization.} After prediction-CoT supervision, action tokens attend more to key objects across these scenes.}
\label{fig:app_attention_cases}
\end{figure*}

We visualize the last-frame attention from the final full-attention layer, as shown in Figure~\ref{fig:app_attention_cases}. With prediction-CoT supervision, the action queries consistently concentrate on nearby vehicles and vulnerable road users that constrain the maneuver, rather than spreading attention over the whole image. More importantly, we observe a clear qualitative difference between models trained with and without CoT. Without CoT supervision, the attention is dominated by the lead vehicle that the ego car is following, while surrounding vehicles receive little attention. This narrow focus becomes problematic in cluttered or safety-critical scenes: nearby hazardous vehicles can be overlooked, which in turn prevents the policy from reacting safely. In contrast, adding CoT supervision broadens the attention of the VLA to a larger set of task-relevant objects, including adjacent vehicles, pedestrians, and cyclists that may affect the maneuver. As a result, the model acquires more comprehensive spatial awareness and produces safer driving strategies. These observations are consistent with the quantitative gains in NC and DAC reported in Table~\ref{tab:cot}.

\FloatBarrier
\FloatBarrier
\subsection{Trajectory visualization: collision reduction}
\label{app:cot_collision_vis}

The attention evidence is reflected in the trajectories. Training CoT reduces collisions and improves the safety terms in Table~\ref{tab:cot}: NC rises from $98.8$ to $99.0$, DAC from $96.8$ to $97.9$, and TTC from $95.8$ to $96.5$. The following three collision cases show the qualitative counterpart: the no-CoT policy enters the conflict region, whereas the CoT-trained policy identifies the nearby agent and keeps a feasible clearance, as shown in \ref{fig:app_cot_collision}.

\begin{center}
\centering
\includegraphics[width=0.70\textwidth]{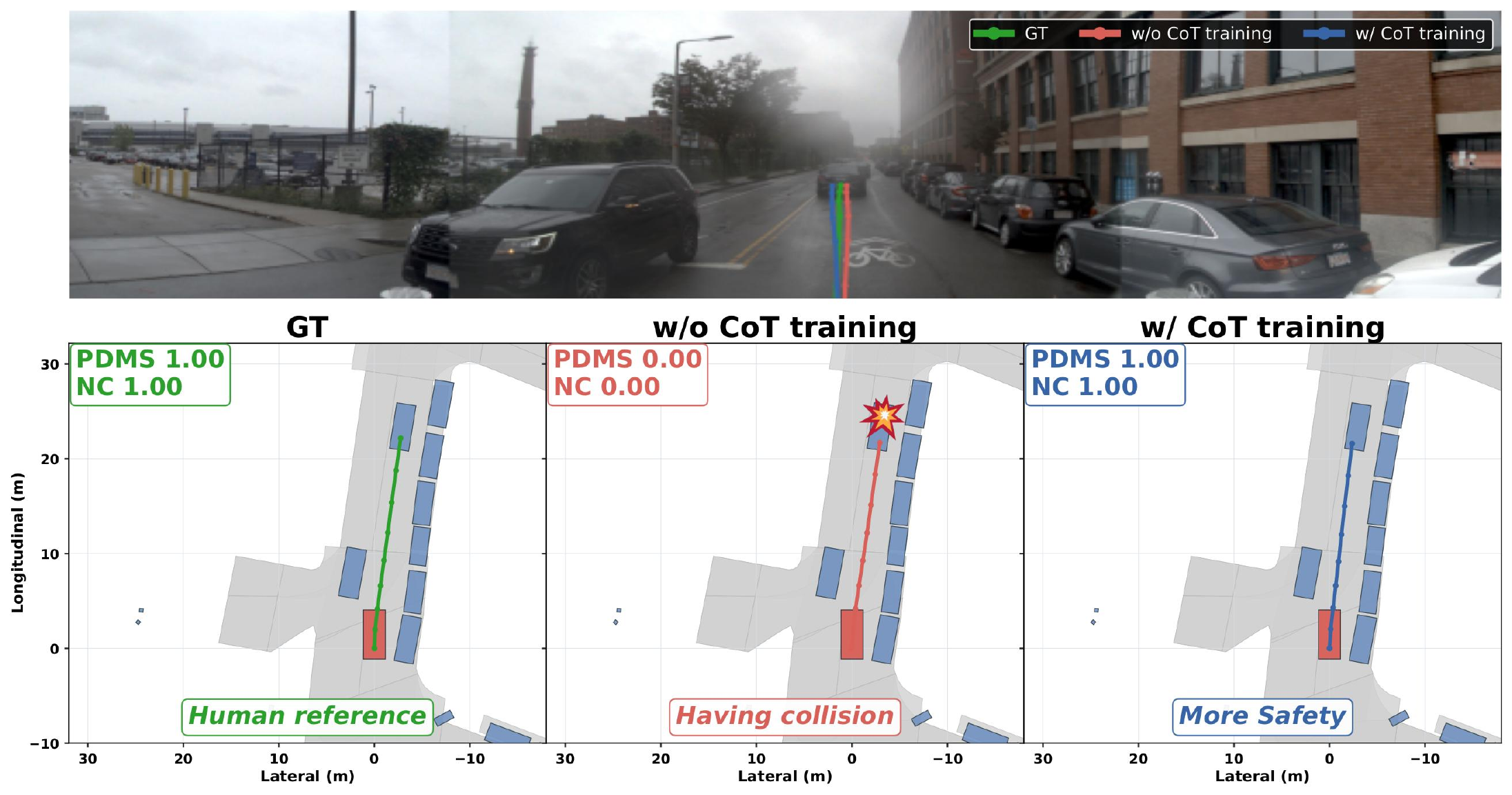}
\vspace{8pt}
\includegraphics[width=0.70\textwidth]{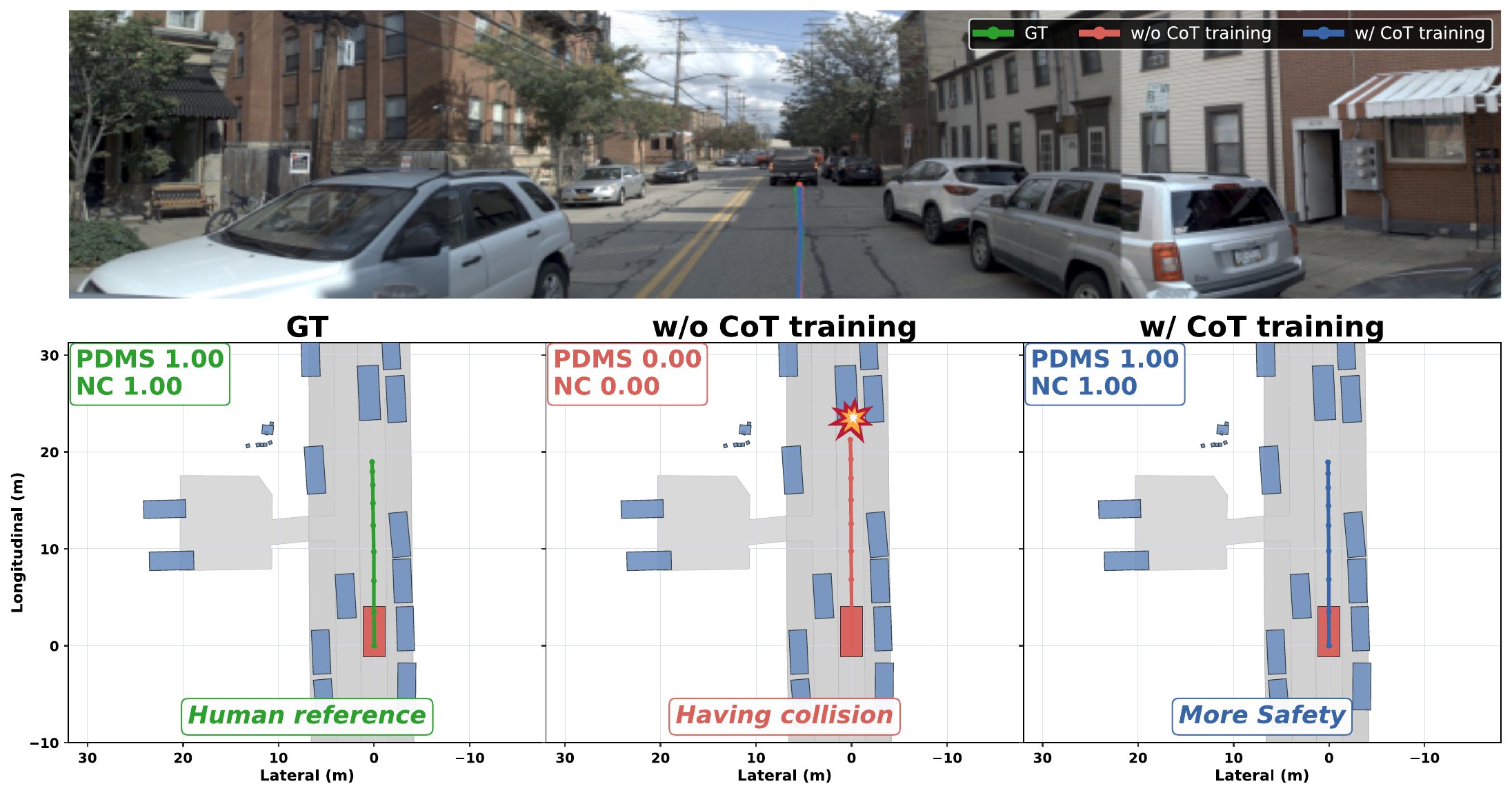}
\vspace{8pt}
\includegraphics[width=0.70\textwidth]{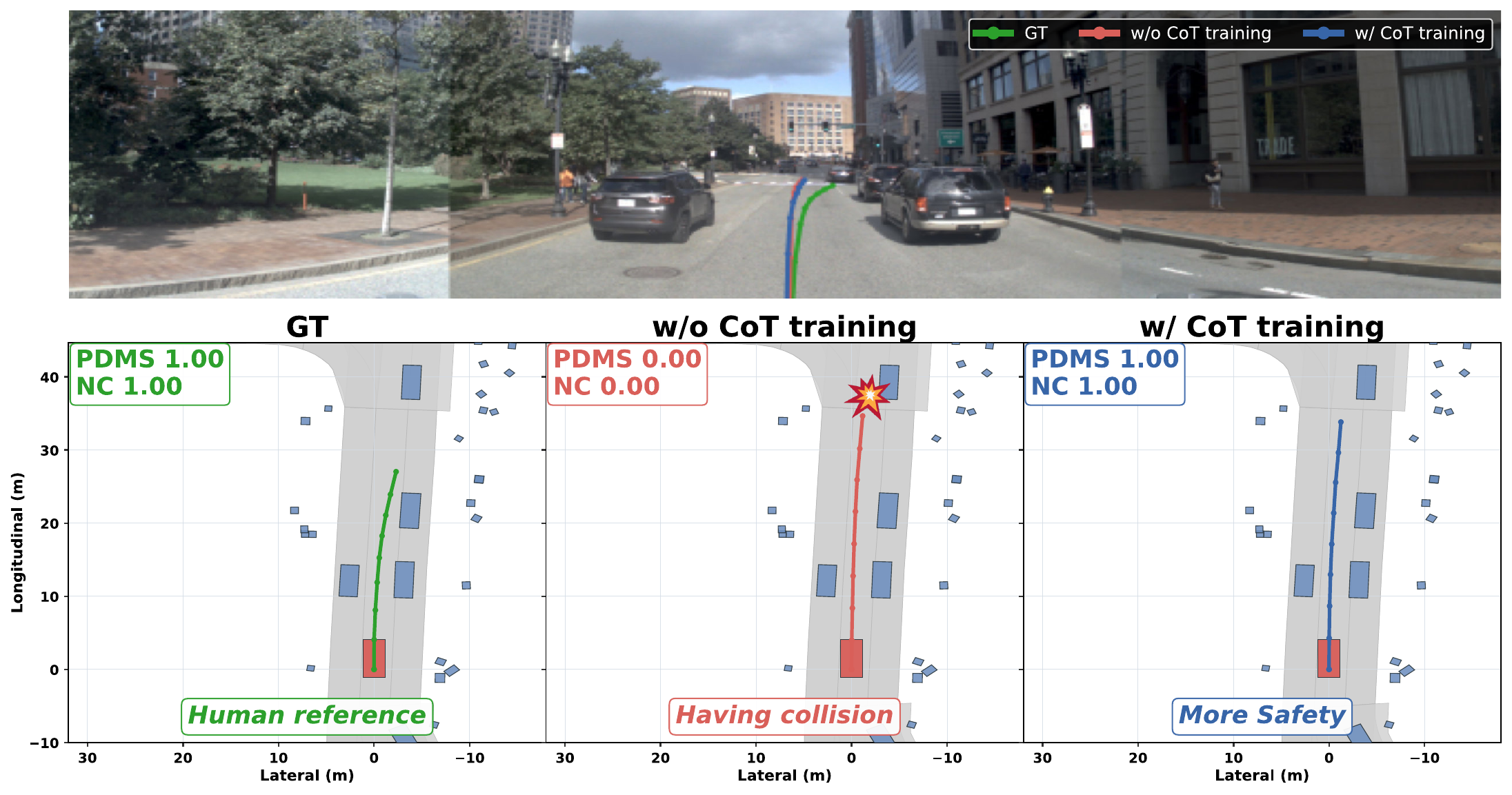}
\captionof{figure}{\textbf{Collision cases.} Each row compares the no-CoT and prediction-CoT trajectories in the same scene. CoT redirects the action away from the conflict agent, consistent with the improved NC, DAC, and TTC values in Table~\ref{tab:cot}.}
\label{fig:app_cot_collision}
\end{center}
\FloatBarrier
\subsection{Trajectory visualization: progress preservation}
\label{app:cot_progress_vis}

Safety does not come from stopping. The prediction-CoT model also improves ego progress (EP $83.5\!\to\!84.2$) and the aggregate PDMS ($89.2\!\to\!90.6$) in Table~\ref{tab:cot}, while retaining full comfort. The three progress cases, as shown in \ref{fig:app_cot_progress},  below show that the CoT policy avoids the conflict and continues along the route instead of replacing unsafe motion with an overly conservative halt.

\begin{center}
\centering
\includegraphics[width=0.70\textwidth]{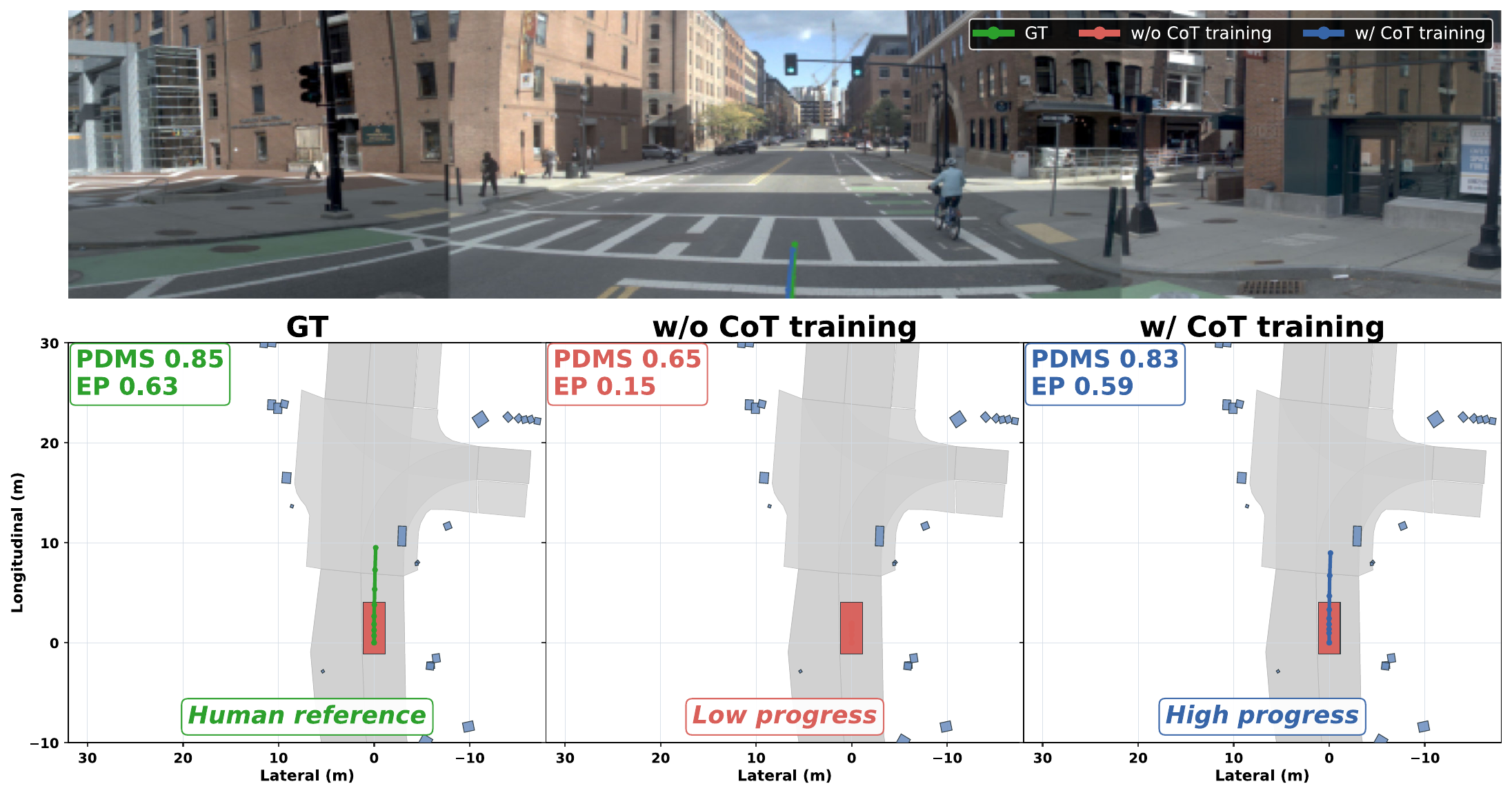}
\vspace{8pt}
\includegraphics[width=0.70\textwidth]{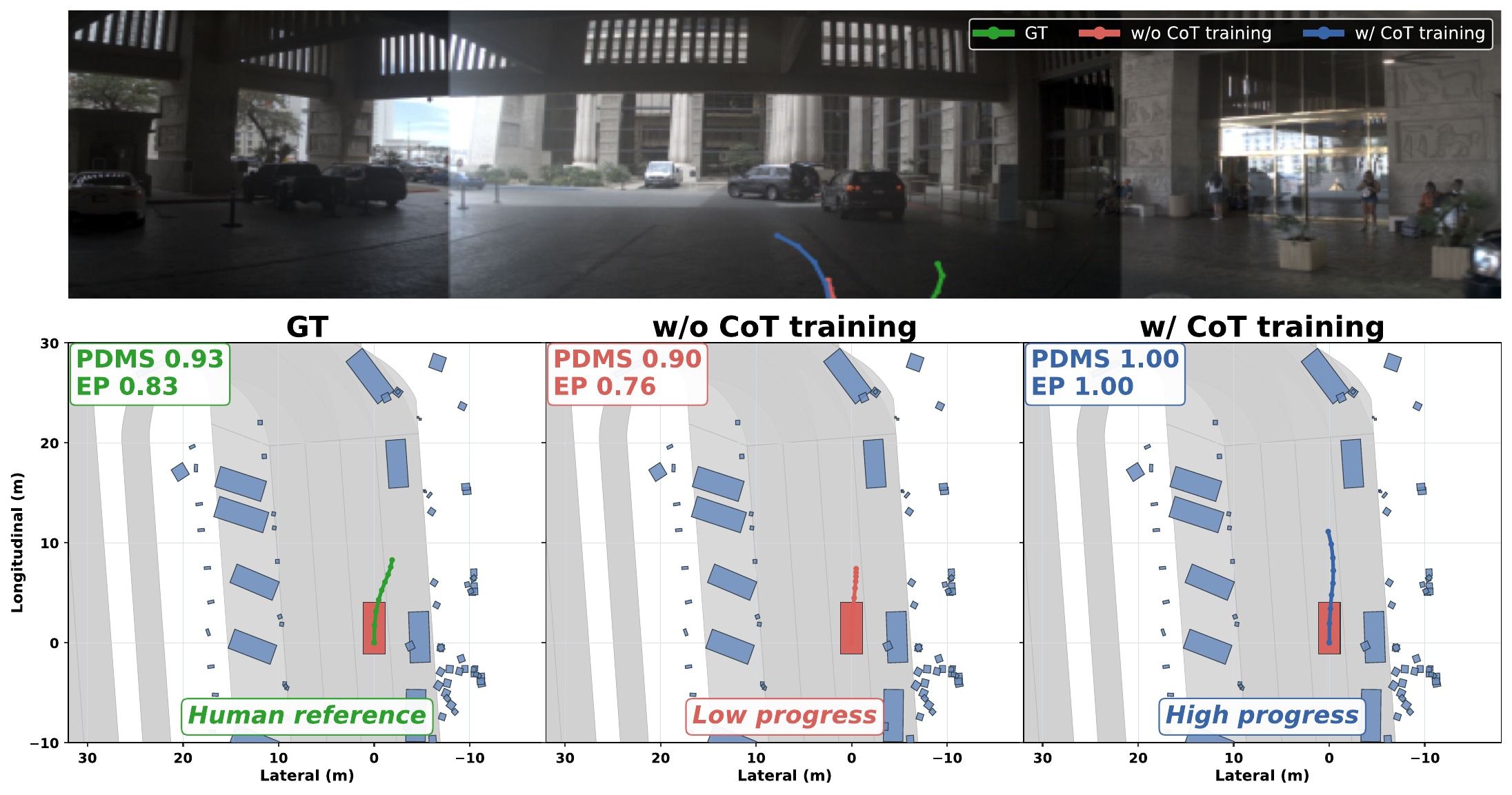}
\vspace{8pt}
\includegraphics[width=0.70\textwidth]{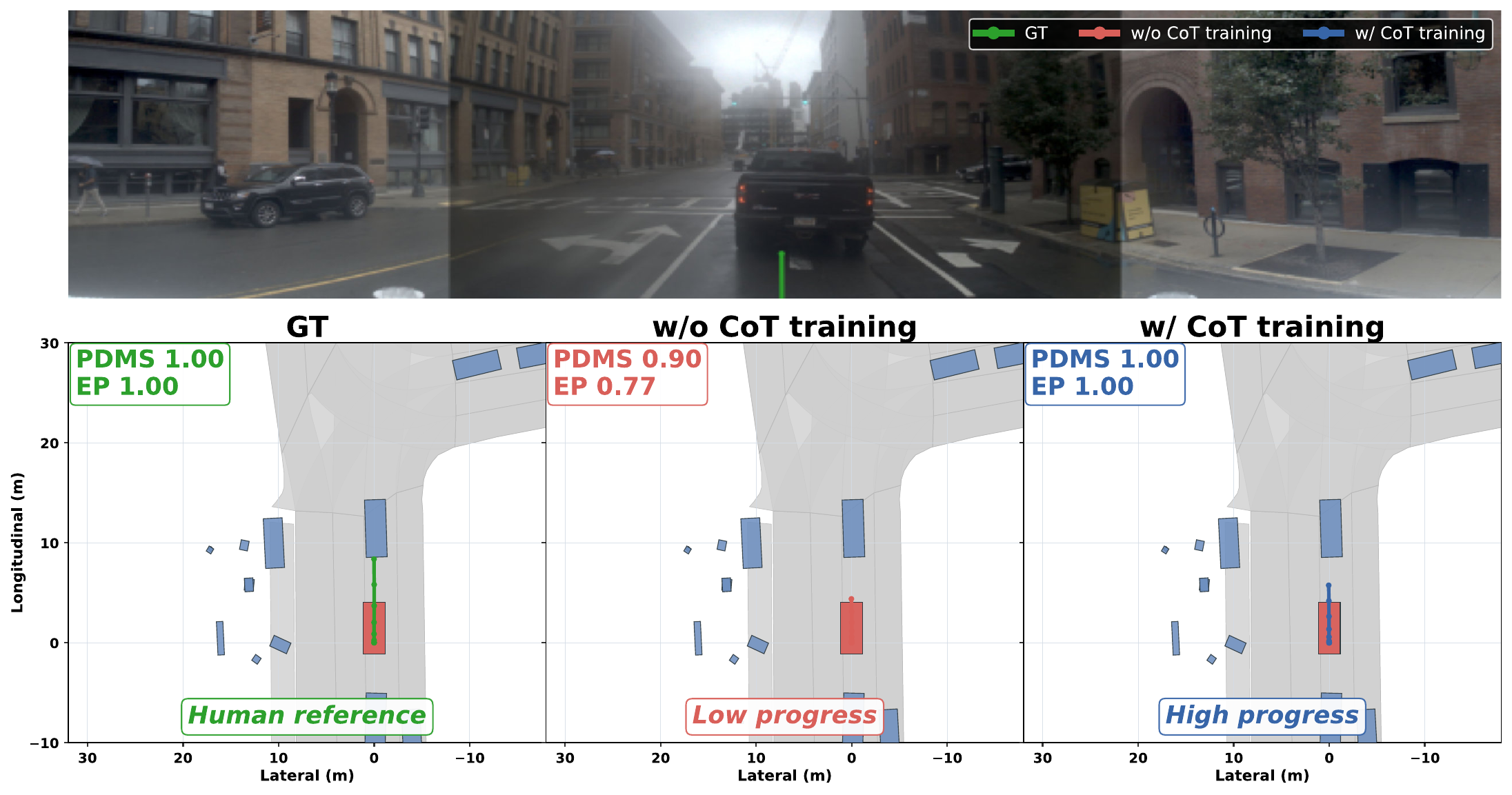}
\captionof{figure}{\textbf{Progress-preserving cases.} Prediction CoT keeps the route-directed motion while avoiding the nearby agent. The qualitative cases complement the EP and PDMS gains reported in Table~\ref{tab:cot}; they are representative examples rather than a pointwise guarantee.}
\label{fig:app_cot_progress}
\end{center}
\FloatBarrier
\section{Algorithmic Details: SFT, AUGUstus-GRPO}
\subsection{Reward Design Details and Curriculum Rationale}
\label{app:rewards}

\textbf{Prediction (CoT) reward.} Predicted boxes are greedily matched to ground-truth boxes by descending IoU at threshold $0.5$; $r_{\mathrm{det}}=\sum\mathrm{IoU}/\max(\#\mathrm{pred},\#\mathrm{gt})$ is a soft F1 (penalizing both misses and hallucinations), and $r_{\mathrm{fp}}$ is the fraction of matched boxes whose future-position label agrees; $r_{\mathrm{pred}}=r_{\mathrm{det}}\!\cdot\!r_{\mathrm{fp}}$. A \emph{strict format gate} requires exactly one \texttt{<prediction>...</prediction>} block with clean JSON content and no surrounding text or code fences; violations zero $r_{\mathrm{pred}}$ (and only $r_{\mathrm{pred}}$).

\textbf{Planning reward.} Stage I uses $R=r_{\mathrm{pred}}+5\cdot\mathrm{PDMS}$. Stage II uses $R=r_{\mathrm{pred}}+5\,(r_{\mathrm{endpoint}}+r_{\mathrm{safe}})$, where $r_{\mathrm{endpoint}}$ maps the final-waypoint deviation to $\{1,.8,.6,.4,.2\}$ over progressively looser L1 thresholds (the softest of four tested ``soft-error'' shapings worked best), and
\[
\begin{aligned}
r_{\mathrm{safe}}&=\begin{cases}\mathrm{mean}\big(\{\mathrm{TTC}, \mathrm{DAC}\}\big),& \min_{m\in\mathcal G}\,\mathrm{subscore}_m>0\\ 0 & \text{otherwise},\end{cases}\\[-2pt]
\mathcal G&=\{\mathrm{NC},\mathrm{DAC},\mathrm{TTC},\mathrm{EP}\}.
\end{aligned}
\]
with the hard gate additionally zeroing the whole planning term. PDMS is deliberately removed from the stage-II reward for two reasons: (i) late-stage PDMS becomes near-constant within a group (std${\sim}0.02$), so its group-relative advantage is dominated by noise, a measured failure mode of dense-score rewards; (ii) direct PDMS optimization slightly trades NC/TTC for EP. The endpoint-plus-shield shaping preserves a dense informative signal (waypoint geometry) but re-imposes safety ceilings; PDMS remains the validation/model-selection metric throughout, so there is no metric/reward mismatch risk in checkpoint choice.

\subsection{Implementation details}
\label{app:impl}

\subsubsection{SFT Training Detail} 
We adopt Qwen3.5-2B backbone network. The action sequence consists of 8 action tokens with hidden dimension \(H{=}2048\). The ActionEncoder is a 3-layer SiLU MLP with concatenated sinusoidal time embedding \(t\). The ActionDecoder follows the structure \(\mathrm{LN}\to\mathrm{Lin}\to\mathrm{SiLU}\to\mathrm{LN}\to\mathrm{Lin}(3)\) and uses a near-zero final initialization. Each layer applies per-norm modulation projections that map \(H\) to \(2H\), corresponding to a DiT variant with approximately \(8\)M additional parameters. We apply quantile normalization with \((q_{01},q_{99})\) per action dimension. Training uses DeepSpeed ZeRO-2 with bf16 precision. The loss weights are \(\lambda_{\mathrm{diff}}{=}10\) and \(\lambda_{\mathrm{lm}}{=}1\). The main model uses dimension weights \((1,1,1)\).

For every training example we sample a flow time \(t\) and noise \(\varepsilon\), construct \(x_t=(1-t)x_0+t\varepsilon\), and run one joint AUGUstus-mode forward. The action head predicts \(\hat x_0\) while the language head teacher-forces the structured CoT \(y\). The complete objective is
\begin{equation}
\mathcal L_{\mathrm{SFT}}=\lambda_{\mathrm{diff}}\mathcal L_{x_0}+\lambda_{\mathrm{lm}}\mathcal L_{\mathrm{CE}},
\label{eq:app_sft_objective}
\end{equation}
where \(\mathcal L_{x_0}\) is the dimension-normalized flow-matching L1 loss. We use \(\lambda_{\mathrm{diff}}{=}10\) and \(\lambda_{\mathrm{lm}}{=}1\) in the default recipe. The CoT labels are generated from the same scene annotation as the action target. Teacher forcing is used only during SFT: the action tokens are visible to the CoT stream, whereas the action tokens never read CoT tokens. Consequently, \(\mathcal L_{\mathrm{CE}}\) back-propagates through the action states and the shared trunk, but does not create an inference-time dependency on generated text.

\subsubsection{RL Training Detail} 
We perform synchronous rollouts based on Verl, where a unified actor serves both rollout generation and policy updates. Each iteration uses 32 prompts with \(G{=}8\) rollouts per prompt, resulting in 256 sequences. The rollout uses a 3-step SDE with per-dimension diffusion coefficients \(g_d(t)^2=a_d^2 t\), where \(a_d\in(0.05,0.05,0.02)\). All rollouts in a group share the same initial state \(x_1\). During rollout, the prefix KV is frozen, and the behaviour policy output \(\hat x_0\) as well as the CoT token log-probabilities are recorded. For the update, we use a single PPO mini-batch that is fully on-policy, with a learning rate of \(10^{-6}\), a clipping threshold of \(0.2\) for both heads, and a gradient clipping norm of \(1.0\). 

For a context group \(c\) we draw one shared initial noise state \(x_1\) and \(G\) independent transition-noise sequences. Each rollout returns an action trajectory \(x_i\) and a CoT sequence \(y_i\). Its scalar reward is
\begin{equation}
R_i=R_{\mathrm{pred}}(y_i)+\lambda R_{\mathrm{plan}}(x_i),\qquad
A_i=\frac{R_i-\operatorname{mean}_{j}(R_j)}{\operatorname{std}_{j}(R_j)+10^{-6}},
\label{eq:app_shared_adv}
\end{equation}
where we set \(\lambda=5\). The same outcome signal trains both modalities. Stage~I sets \(R_{\mathrm{plan}}=\mathrm{PDMS}\); Stage~II replaces this saturating score by the endpoint-plus-safety shaping in Appendix~\ref{app:rewards}.The action objective is the clipped JaT-GRPO surrogate, and the language objective is token-level clipped CoT-GRPO:
\begin{align}
\mathcal J_a&=\mathbb E\left[\min(\rho_a A,\operatorname{clip}(\rho_a,1-\epsilon,1+\epsilon)A)\right],\\
\mathcal J_c&=\mathbb E\left[\frac{1}{|y|}\sum_k\min(\rho_{c,k} A,\operatorname{clip}(\rho_{c,k},1-\epsilon,1+\epsilon)A)\right].
\label{eq:app_joint_grpo}
\end{align}
The update maximizes $\mathcal J_a+\alpha_c\mathcal J_c$ while adding the exact action KL in Eq.~(\ref{eq:kl}) and a token-level reference KL for CoT. Thus CoT-GRPO is ordinary PPO on the discrete suffix, but its advantage is tied to the same simulator outcome as the continuous action head.

\begin{algorithm}[htbp]
\caption{AUGUstus supervised fine-tuning: complete procedure}
\label{alg:augustus_sft_full}
\small
\begin{algorithmic}[1]
\REQUIRE Pretrained parameters $\theta$; training samples $\mathcal D=\{(c,x_0,y)\}$; loss weights and optimizer.
\ENSURE Cold-start policy $\pi_{\mathrm{SFT}}$ with action and prediction-CoT outputs.
\FOR{each supervised training iteration}
    \STATE Sample a minibatch of contexts, target trajectories, and prediction CoTs from $\mathcal D$.
    \STATE Normalize target trajectories using the fixed action statistics.
    \STATE Sample flow times $t$ and independent noise $\varepsilon\sim\mathcal N(0,I)$.
    \STATE Form $x_t=(1-t)x_0+t\varepsilon$ for each training sample.
    \STATE Encode context $c=(S,E,C,H)$, noisy action $x_t$ with time $t$, and shifted CoT tokens $y_{<k}$.
    \STATE Run one shared-transformer forward pass with AUGUstus-mode hybrid attention.
    \STATE Read the clean-trajectory prediction $\hat x_0$ and the token distribution $p_\theta(y_k\mid y_{<k},c,x_t)$.
    \STATE Compute the joint flow-matching and teacher-forced CoT loss in Eq.~(\ref{eq:sftloss}) and include the auxiliary module losses.
    \STATE Backpropagate both losses jointly through the shared transformer and update $\theta$.
    \STATE Validate with action-only denoising.
\ENDFOR
\RETURN The selected cold-start policy $\pi_{\mathrm{SFT}}$.
\end{algorithmic}
\end{algorithm}

\begin{algorithm}[htbp]
\caption{AUGUstus deployment: complete AUGUstus-mode inference}
\label{alg:augustus_inference_full}
\small
\begin{algorithmic}[1]
\REQUIRE Trained policy $\pi_\theta$; context $c=(S,E,C,H)$; schedule $1=t_0>\cdots>t_N=0$; optional CoT flag.
\ENSURE Future ego trajectory $x^{\mathrm{out}}$ and, only if requested, prediction CoT $y$.
\STATE Encode the causal context prefix once and cache its states.
\STATE Draw $x\sim\mathcal N(0,I)$ in normalized action space.
\FOR{$n=0$ to $N-1$}
    \STATE Evaluate the bidirectional action block against the cached context to obtain $\hat x_0$.
    \STATE Apply the deterministic flow step: $x\leftarrow x-(t_n-t_{n+1})(x-\hat x_0)/t_n$.
\ENDFOR
\STATE Invert the fixed action normalization to obtain $x^{\mathrm{out}}$.
\IF{prediction CoT is requested}
    \STATE Append the denoised action block to the context with zero-time conditioning.
    \STATE Autoregressively decode $y_k\sim p_\theta(\cdot\mid y_{<k},c,x)$ with causal CoT attention.
    \RETURN $(x^{\mathrm{out}},y)$.
\ELSE
    \RETURN $x^{\mathrm{out}}$ without CoT decoding.
\ENDIF
\end{algorithmic}
\end{algorithm}

\subsection{Complete Training and Inference Algorithms}
\label{app:algorithms}

Algorithms~\ref{alg:augustus_sft_full} and~\ref{alg:augustus_inference_full} give the full cold-start and deployment procedures corresponding to the contrasts in Algorithms~\ref{alg:augustus} and~\ref{alg:augustus_inference}. Both use the same AUGUstus-mode attention (action before CoT): context is causal, actions are bidirectional, and causal CoT tokens can read context and actions but not vice versa. Training uses noisy ground-truth actions and teacher-forced prediction CoT; deployment uses initial Gaussian noise and deterministic action denoising, with no ground-truth action or CoT required. Action normalization, time conditioning, and auxiliary module losses follow Appendix~\ref{app:architecture} and Appendix~\ref{app:impl}.

\subsection{Complete two-stage unified GRPO}

Algorithm~\ref{alg:augustus_grpo_full} specifies the complete two-stage post-training procedure. The reference is frozen at SFT throughout both stages; the behavior policy is snapshotted before each on-policy rollout batch. Within each context group, initial flow noise is shared, whereas transition exploration noise and CoT sampling distinguish the rollouts. The standard run uses the noise settings in Appendix~\ref{app:noise}, the reward and format gates in Appendix~\ref{app:rewards}, and the optimization settings in Appendix~\ref{app:impl}. The same prediction reward is used in both stages. Stage I uses $R_{\mathrm{pred}}+5\,\mathrm{PDMS}$, whereas Stage II uses $R_{\mathrm{pred}}+5(R_{\mathrm{endpoint}}+R_{\mathrm{safe}})$ with the planning safety gate.

\begin{algorithm}[!htbp]
\caption{AUGUstus-GRPO: complete two-stage unified optimization}
\label{alg:augustus_grpo_full}
\small
\begin{algorithmic}[1]
\REQUIRE Cold-start policy $\pi_{\mathrm{SFT}}$; contexts $\mathcal D$; group size $G$; denoising schedule; clip and KL weights.
\ENSURE Two-stage action--CoT policy $\pi_{\mathrm{II}}^{*}$.
\STATE Freeze $\pi_{\mathrm{ref}}\leftarrow\pi_{\mathrm{SFT}}$ and initialize $\pi_\theta\leftarrow\pi_{\mathrm{SFT}}$.
\FOR{stage $s\in\{1,2\}$}
    \IF{$s=2$}
        \STATE Warm-start $\pi_\theta\leftarrow\pi_{\mathrm I}^{*}$, the best validation-PDMS Stage-I checkpoint.
    \ENDIF
    \FOR{each on-policy RL iteration}
        \STATE Snapshot the behavior policy $\pi_{\mathrm{old}}\leftarrow\pi_\theta$ and sample a context minibatch.
        \FOR{each context $c$ in the minibatch}
            \STATE Draw one initial $x_1\sim\mathcal N(0,I)$ shared by the $G$ rollouts.
            \FOR{$i=1$ to $G$}
                \STATE Initialize $x_{i,t_0}\leftarrow x_1$.
                \FOR{$n=0$ to $N-1$}
                    \STATE Predict $\hat x_0^{\mathrm{old}}=f_{\mathrm{old}}(x_{i,t_n},t_n,c)$.
                    \STATE Draw kernel noise $z$ and sample $x_{i,t_{n+1}}\leftarrow c_xx_{i,t_n}+c_0\hat x_0^{\mathrm{old}}+\sigma_s z$ via Eq.~(\ref{eq:kernel}).
                    \STATE Record both states, $\hat x_0^{\mathrm{old}}$, $z$, and the kernel coefficients for re-scoring.
                \ENDFOR
                \STATE Set $x_i\leftarrow x_{i,0}$; decode action-conditioned CoT $y_i$ and record its token log probabilities.
                \STATE Compute $R_{\mathrm{pred},i}$ with the same prediction scoring and format gate in both stages.
                \IF{$s=1$}
                    \STATE Set $R_{\mathrm{plan},i}\leftarrow\mathrm{PDMS}_i$.
                \ELSE
                    \STATE Set $R_{\mathrm{plan},i}\leftarrow R_{\mathrm{ep},i}+R_{\mathrm{safe},i}$ and apply the planning safety gate.
                \ENDIF
                \STATE Set $R_i\leftarrow R_{\mathrm{pred},i}+5R_{\mathrm{plan},i}$.
            \ENDFOR
            \STATE Normalize group rewards into the shared advantages $A_i$ using Eq.~(\ref{eq:reward}).
        \ENDFOR
        \STATE Re-evaluate recorded action states and CoT prefixes under $\pi_\theta$ and the frozen $\pi_{\mathrm{ref}}$.
        \STATE Evaluate JaT-GRPO ratios directly from $\hat x_0^\theta-\hat x_0^{\mathrm{old}}$ using Eq.~(\ref{eq:logratio}).
                \STATE Evaluate the action and CoT ratios and their native KL terms; keep each rollout's shared $A_i$.
                \STATE Compute the two clipped objectives $\mathcal J_a$ and $\mathcal J_c$.
        \STATE Jointly update $\theta$ with $\mathcal L=-\mathcal J_a+\beta_a\mathrm{KL}_{\mathrm{act}}+\alpha_c(-\mathcal J_c+\beta_c\mathrm{KL}_{\mathrm{CoT}})$.
        \STATE Synchronize the updated policy for the next rollout batch and retain the best validation checkpoint.
    \ENDFOR
    \STATE Save the selected stage checkpoint as $\pi_{\mathrm I}^{*}$ or $\pi_{\mathrm{II}}^{*}$.
\ENDFOR
\RETURN $\pi_{\mathrm{II}}^{*}$.
\end{algorithmic}
\end{algorithm}

\FloatBarrier

\subsection{Three-stage training trace}
\begin{figure*}[!htbp]
\centering
\includegraphics[width=0.64\textwidth]{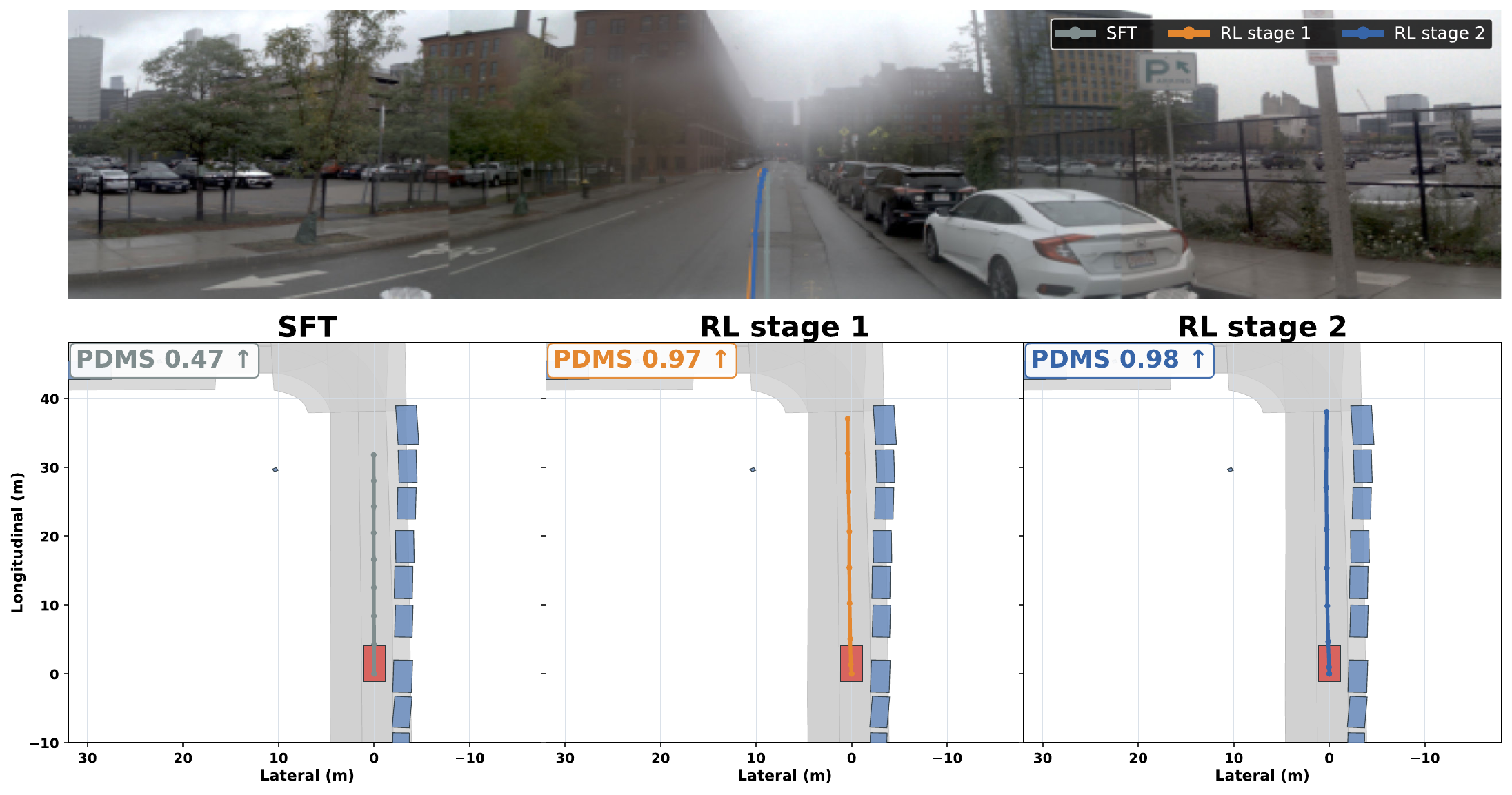}

\vspace{4pt}
\includegraphics[width=0.64\textwidth]{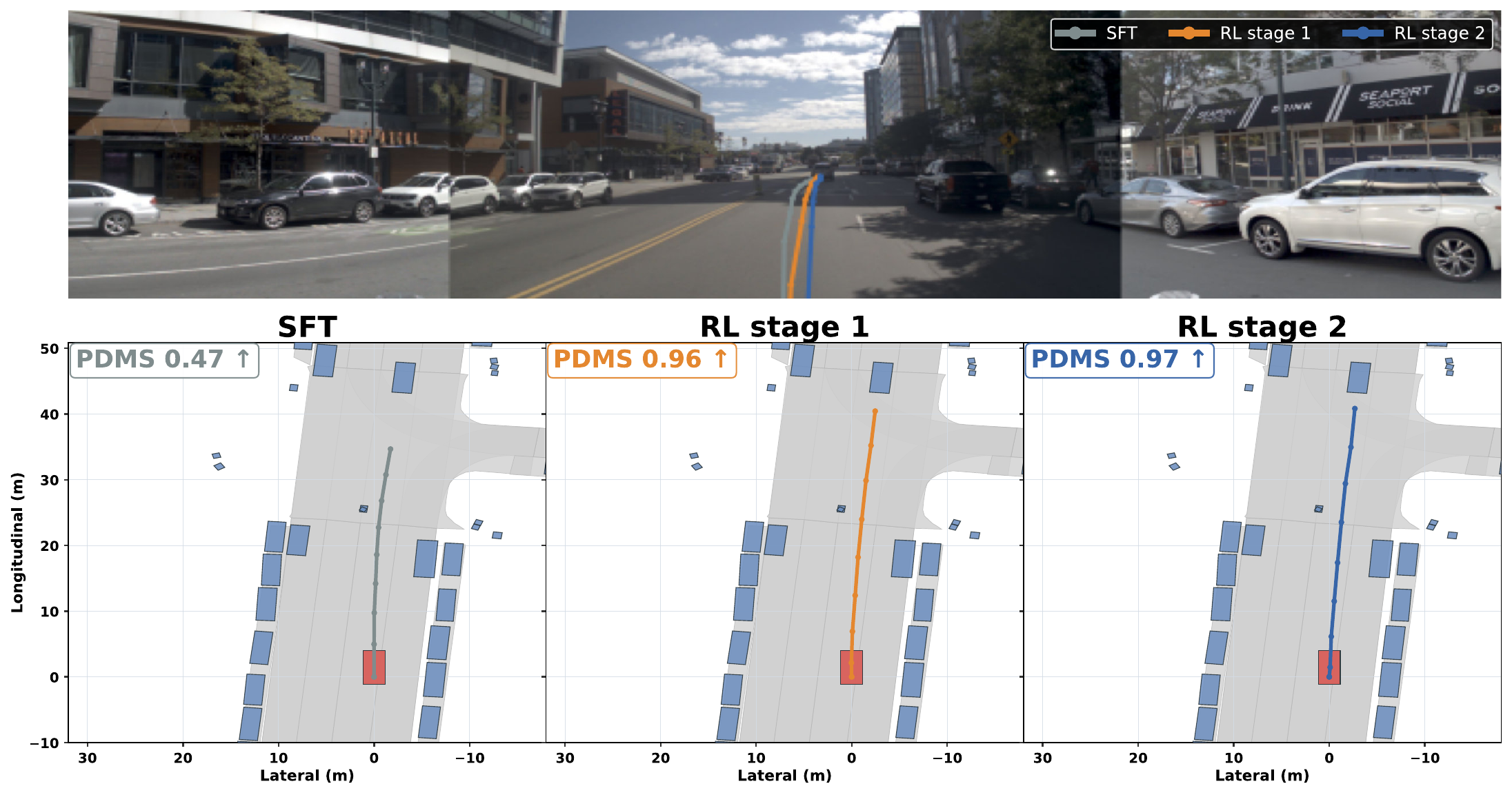}

\vspace{4pt}
\includegraphics[width=0.64\textwidth]{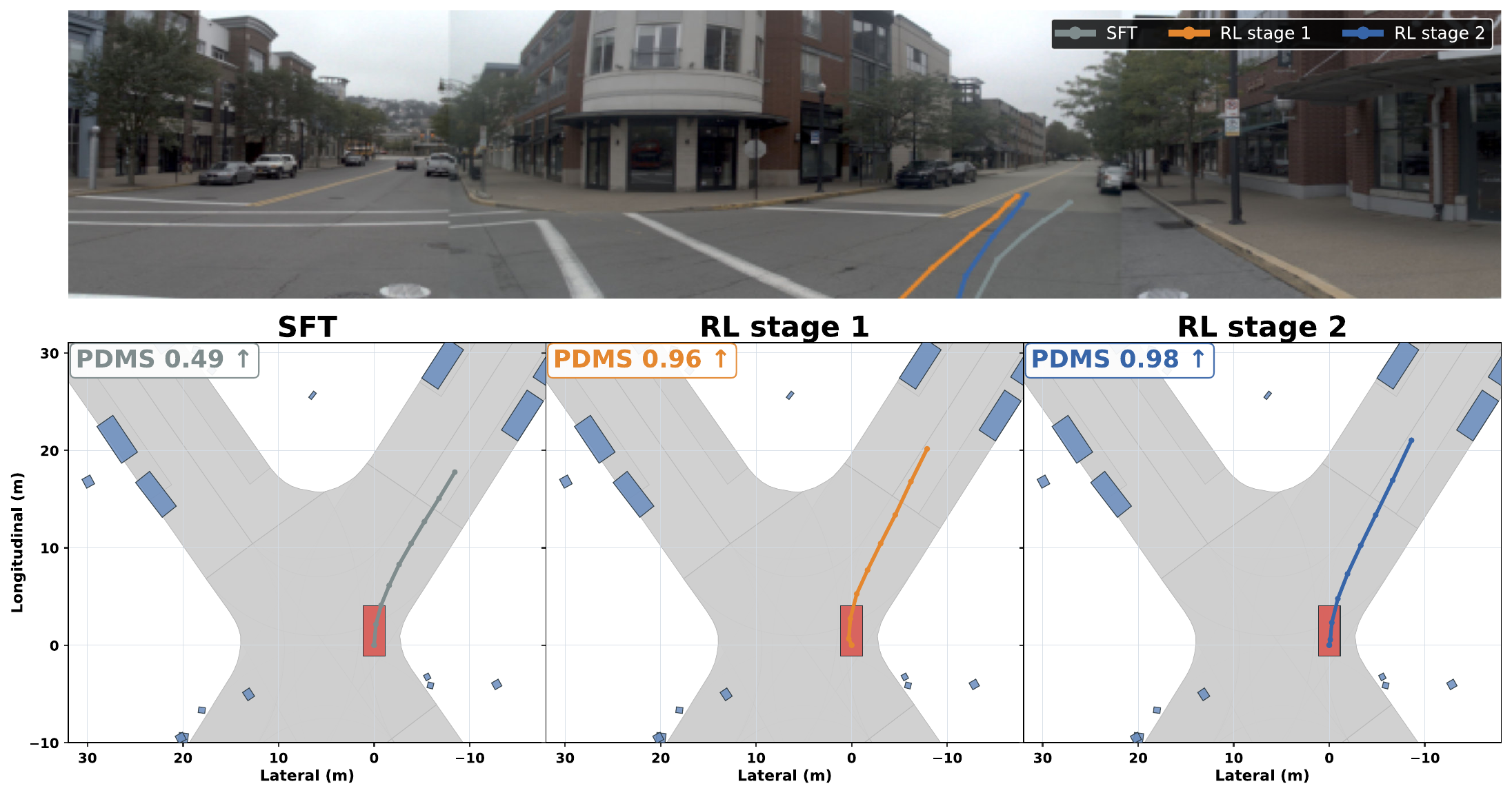}
\caption{\textbf{Three-stage RL visualization.} Each case shows the same scene under SFT, RL Stage~I, and RL Stage~II. Stage~I moves the path toward the desired progress profile; Stage~II retains that progress while increasing the safety margin. The displayed case-level PDMS values are illustrative qualitative traces, whereas the aggregate values are in Table~\ref{tab:rl_main}.}
\label{fig:app_rl_stages}
\end{figure*}

The SFT checkpoint provides scene grounding and a stable action prior. Stage~I uses PDMS to improve progress and establishes the first RL checkpoint. Stage~II switches to endpoint and safety shaping to recover within-group variance after PDMS saturates. The qualitative traces in Fig.~\ref{fig:app_rl_stages} show the same progression: the predicted path becomes more compliant at Stage~I and then adds a larger safety margin at Stage~II. Meanwhile, the CoT prediction accuracy improves from 0.75 at SFT to 0.85 after RL, indicating increasingly accurate identification of critical objects and thereby enhancing the model's action planning capability.

\FloatBarrier
\subsection{Deployment Complexity}
\label{app:complexity}
\textbf{Latency.} The deployed path is a causal visual prefix followed by one bidirectional eight-token action suffix. The vision encoder remains eager because the multimodal preprocessing contains host synchronization; the prefix and denoising suffix are captured in one CUDA graph. Table~\ref{tab:app_latency} reports the steady-state batch-one measurements from the profiling script in \texttt{fast\_inference.md}.

\begin{table}[t]
\centering\small
\caption{\textbf{Measured batch-one latency on one H800.}}
\label{tab:app_latency}
\begin{tabular}{lrr}
\toprule
Path or module & Mean (ms) & Fraction \\
\midrule
Vision encoder and M-RoPE & 40.20 & 54.6\% \\
Context prefix, eager & 46.04 & 34.8\% \\
One-step denoising, eager & 45.70 & 34.6\% \\
LLM stage, CUDA-graph replay & 33.19 & 45.1\% \\
\textbf{End-to-end, eager} & 131.88 & -- \\
\textbf{End-to-end, CUDA graph} & \textbf{73.67} & \textbf{13.6 Hz} \\
\bottomrule
\end{tabular}
\end{table}

The eager profile launches about $3.7$k small kernels and performs roughly 200 pageable host-to-device copies per sample. Graph replay removes this CPU dispatch overhead and reduces the LLM stage from $91.83$ to $33.19$\,ms (2.77$\times$), while leaving the visual stage unchanged. The action suffix contributes only eight tokens on top of a roughly 2,257-token context. Optional post-hoc CoT is therefore deliberately disabled in the speed-critical configuration: a measured 166-token explanation takes about 4.5\,s at batch one, whereas the action-only path is below 100\,ms.

\textbf{Scorer Module.}
Let $C_{\mathrm{vis}}$ be the visual encoder cost, $C_{\mathrm{pre}}(L)$ the prefix cost for $L$ context tokens, and $C_{\mathrm{act}}(T)$ the eight-token action suffix. Our single-trajectory path is
\begin{equation}
T_{\mathrm{ours}}=C_{\mathrm{vis}}+C_{\mathrm{pre}}(L)+C_{\mathrm{act}}(T),\qquad T=8,
\end{equation}
and is independent of the number of candidates. A VLA+scorer system that generates $N$ trajectories and scores them costs approximately
\begin{equation}
T_{\mathrm{scorer}}(N)=N\big(C_{\mathrm{pre}}(L)+C_{\mathrm{act}}(T)\big)+C_{\mathrm{vis}}+N C_{\mathrm{score}}.
\end{equation}
Thus, $N{=}32$--$64$ multiplies the expensive VLM pass even if the scorer itself is lightweight. As a result, this VLA+scorer design is difficult to run in real time on vehicle hardware. We therefore generate one trajectory in one network evaluation; the deployed path takes 73.6\,ms on an H800 (13.6\,Hz) and 83\,ms on the real vehicle (12\,Hz), while retaining SOTA public-benchmark performance.

\section{Qualitative Results on Public, Private Data, and Driving Test}
\label{app:qualitative}

\subsection{Public benchmark examples}
\begin{figure}[p]
\centering
\begin{subfigure}{0.85\linewidth}
\includegraphics[width=\linewidth]{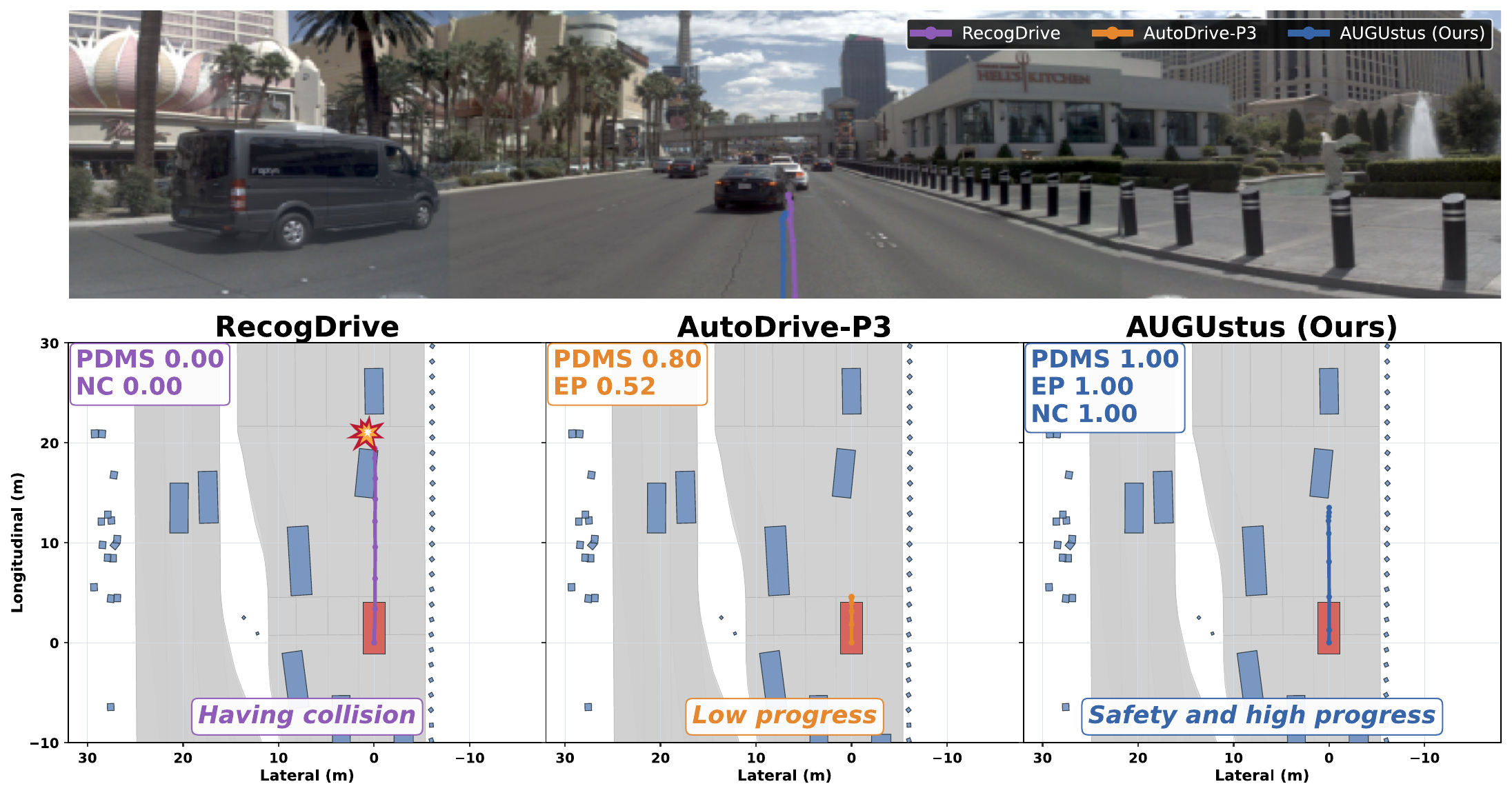}
\end{subfigure}

\vspace{2.5mm}
\begin{subfigure}{0.85\linewidth}
\includegraphics[width=\linewidth]{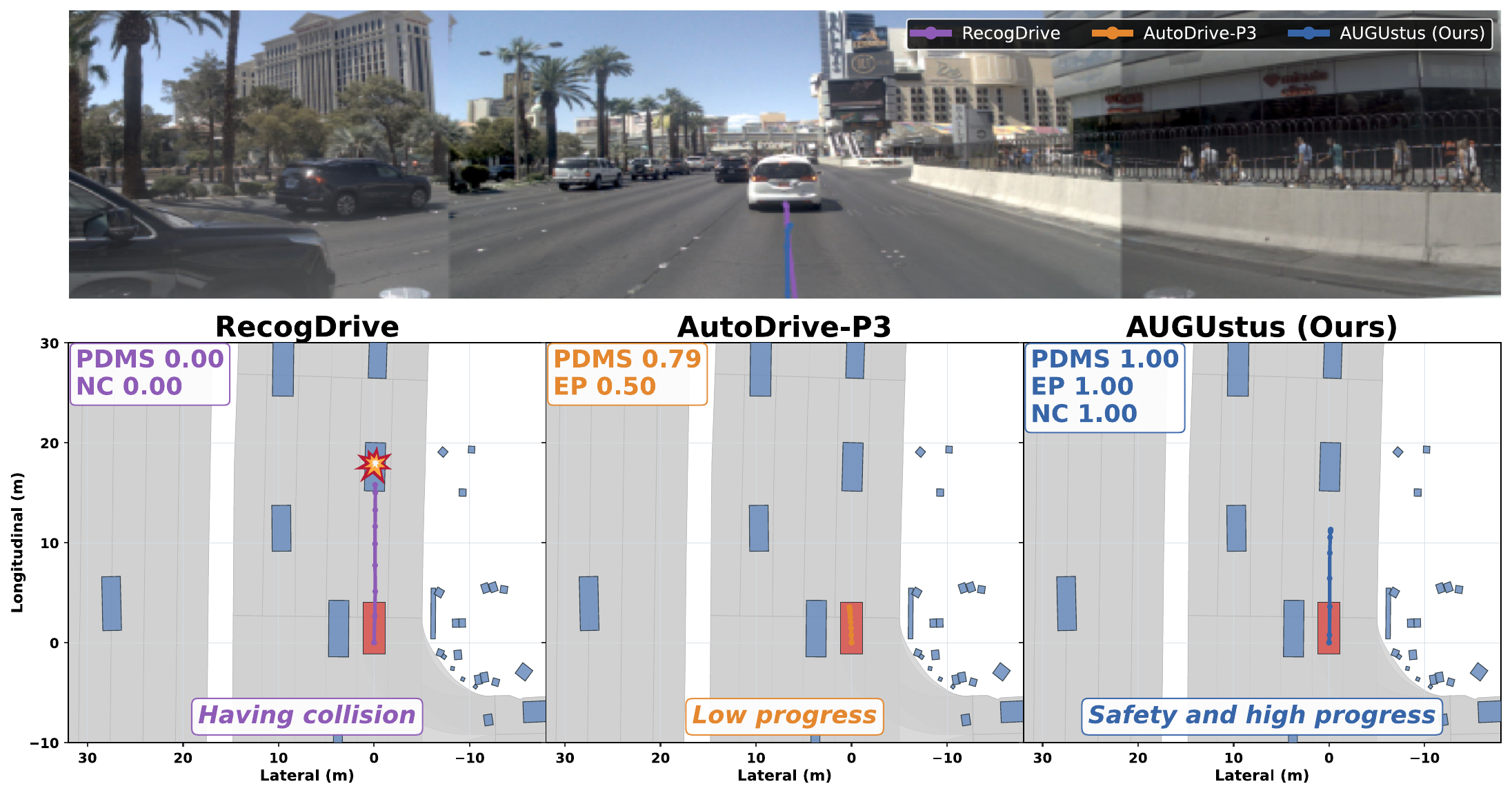}
\end{subfigure}

\vspace{2.5mm}
\begin{subfigure}{0.85\linewidth}
\includegraphics[width=\linewidth]{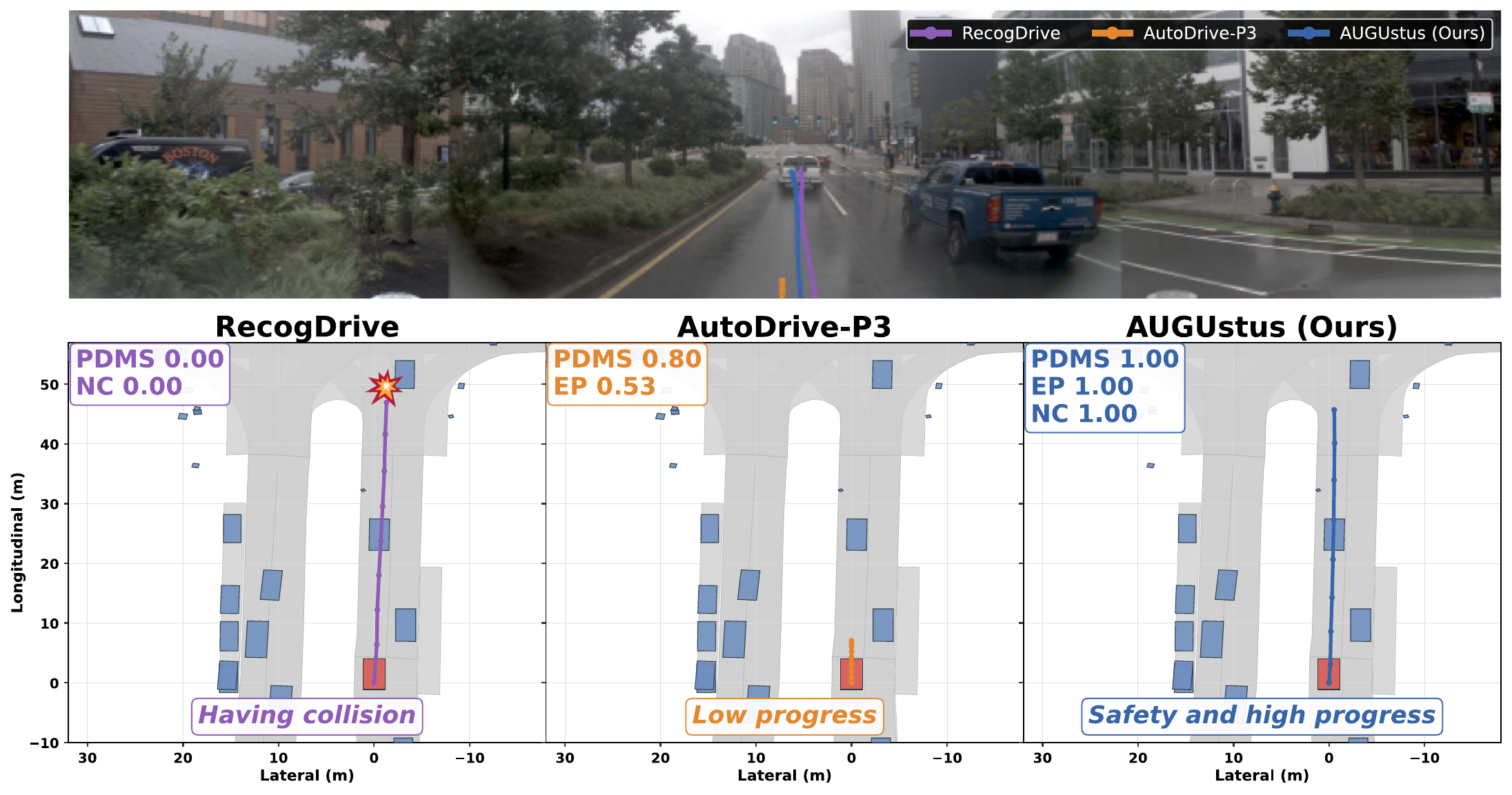}
\end{subfigure}
\caption{\textbf{Public-set qualitative comparison (part 1/2).} Each panel contrasts ReCogDrive, AutoDrive-P$^3$, and AUGUstus on the same public scene. The single AUGUstus trajectory is both collision-free and route compliant in these examples.}
\label{fig:app_public_visuals}
\end{figure}

\begin{figure}[p]
\centering
\begin{subfigure}{0.85\linewidth}
\includegraphics[width=\linewidth]{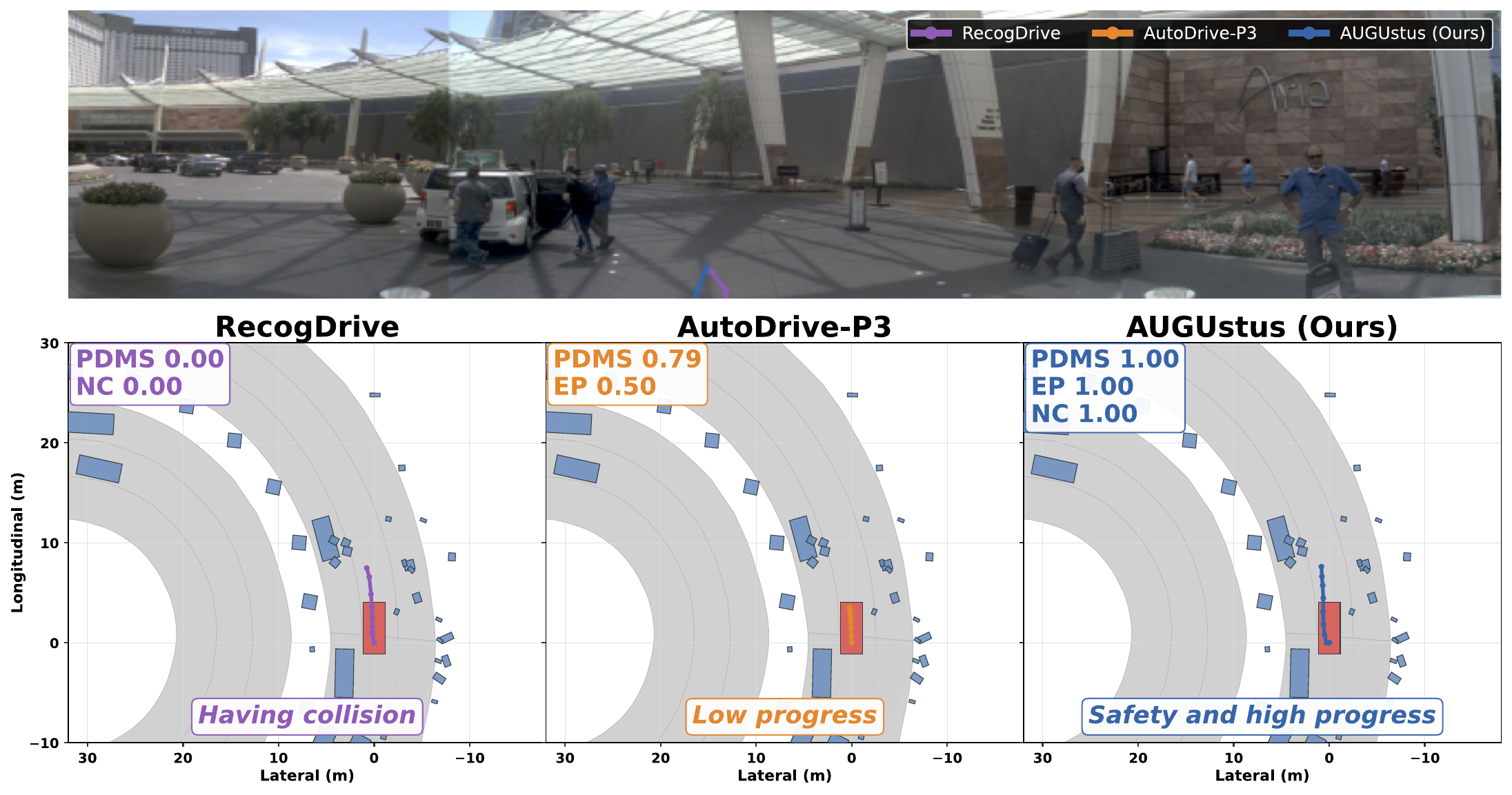}
\end{subfigure}

\vspace{2.5mm}
\begin{subfigure}{0.85\linewidth}
\includegraphics[width=\linewidth]{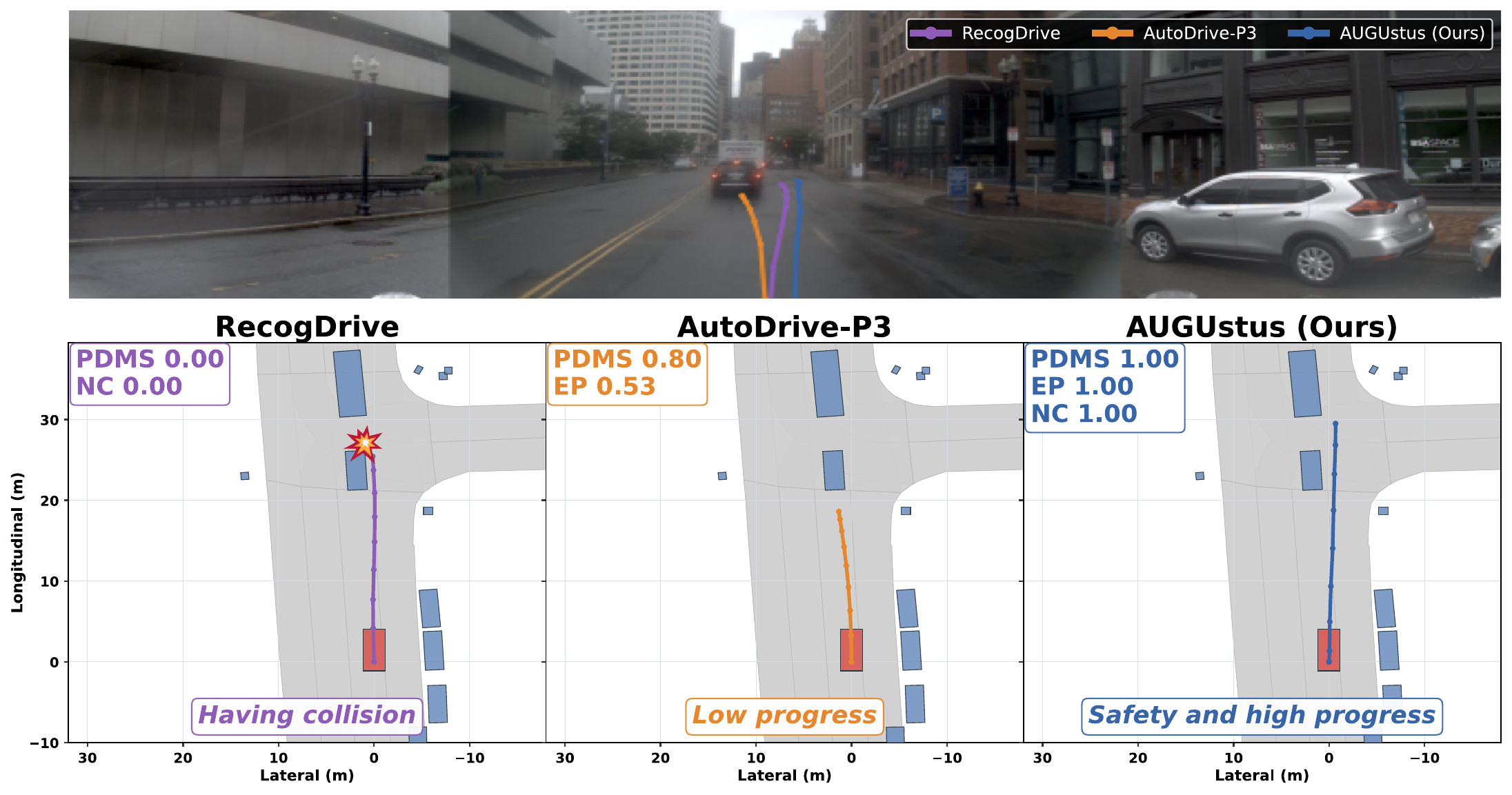}
\end{subfigure}

\vspace{2.5mm}
\begin{subfigure}{0.85\linewidth}
\includegraphics[width=\linewidth]{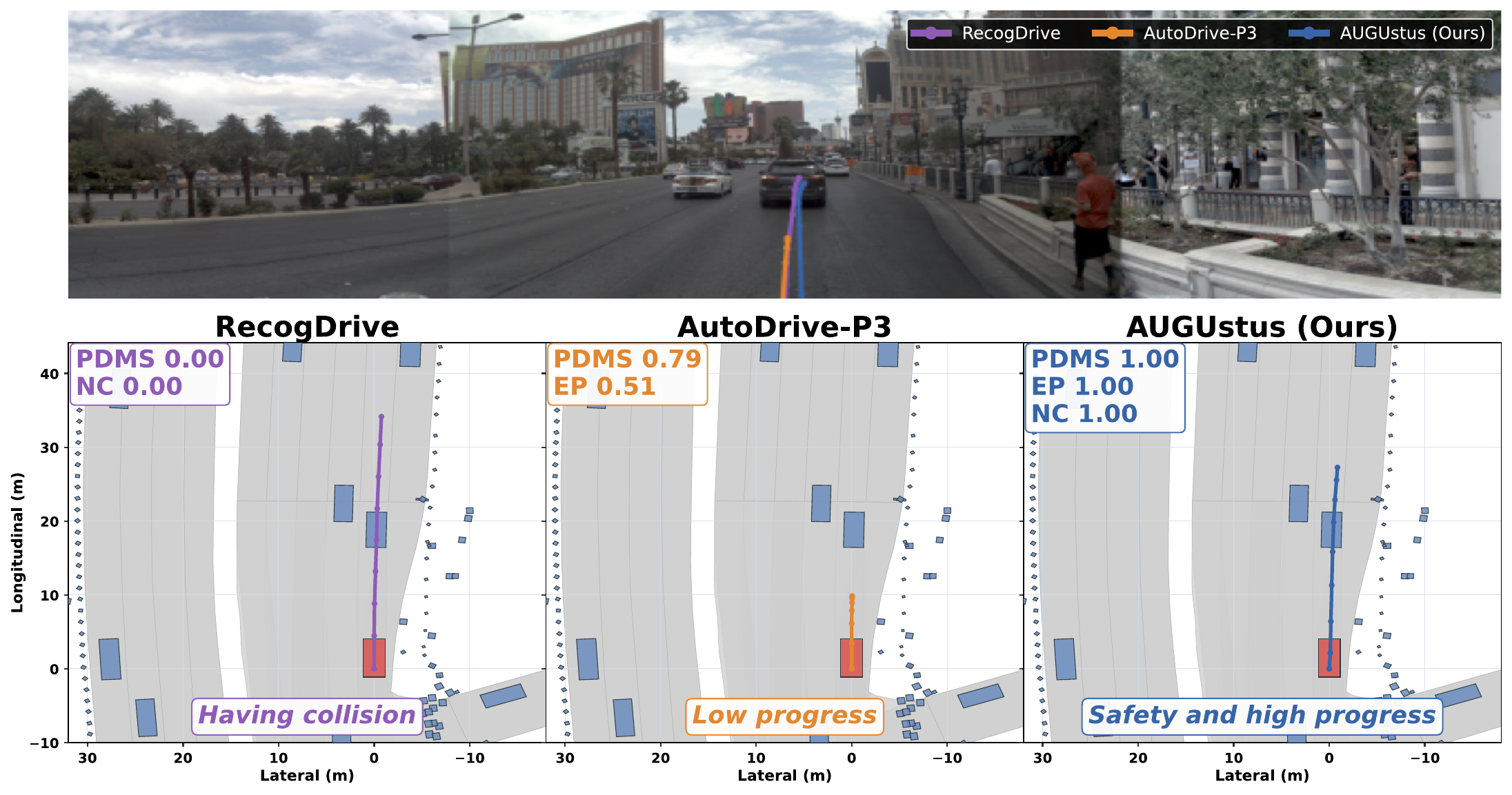}
\end{subfigure}
\caption{\textbf{Public-set qualitative comparison (part 2/2).} Three additional public scenes comparing ReCogDrive, AutoDrive-P$^3$, and AUGUstus under the same visualization protocol.}
\label{fig:app_public_visuals_b}
\end{figure}

The public examples compare the same scene across ReCogDrive \citep{recogdrive}, AutoDrive-P$^3$ \cite{autodrive_p3}, and AUGUstus. The twelve cases in Fig.~\ref{fig:app_public_visuals} and Fig.~\ref{fig:app_public_visuals_b} provides representative qualitative comparisons.

\subsection{Internal 1,000 km data}
\begin{figure}[p]
\centering

\begin{subfigure}{0.72\linewidth}
\centering
\scalebox{0.98}[0.90]{\includegraphics[width=\linewidth]{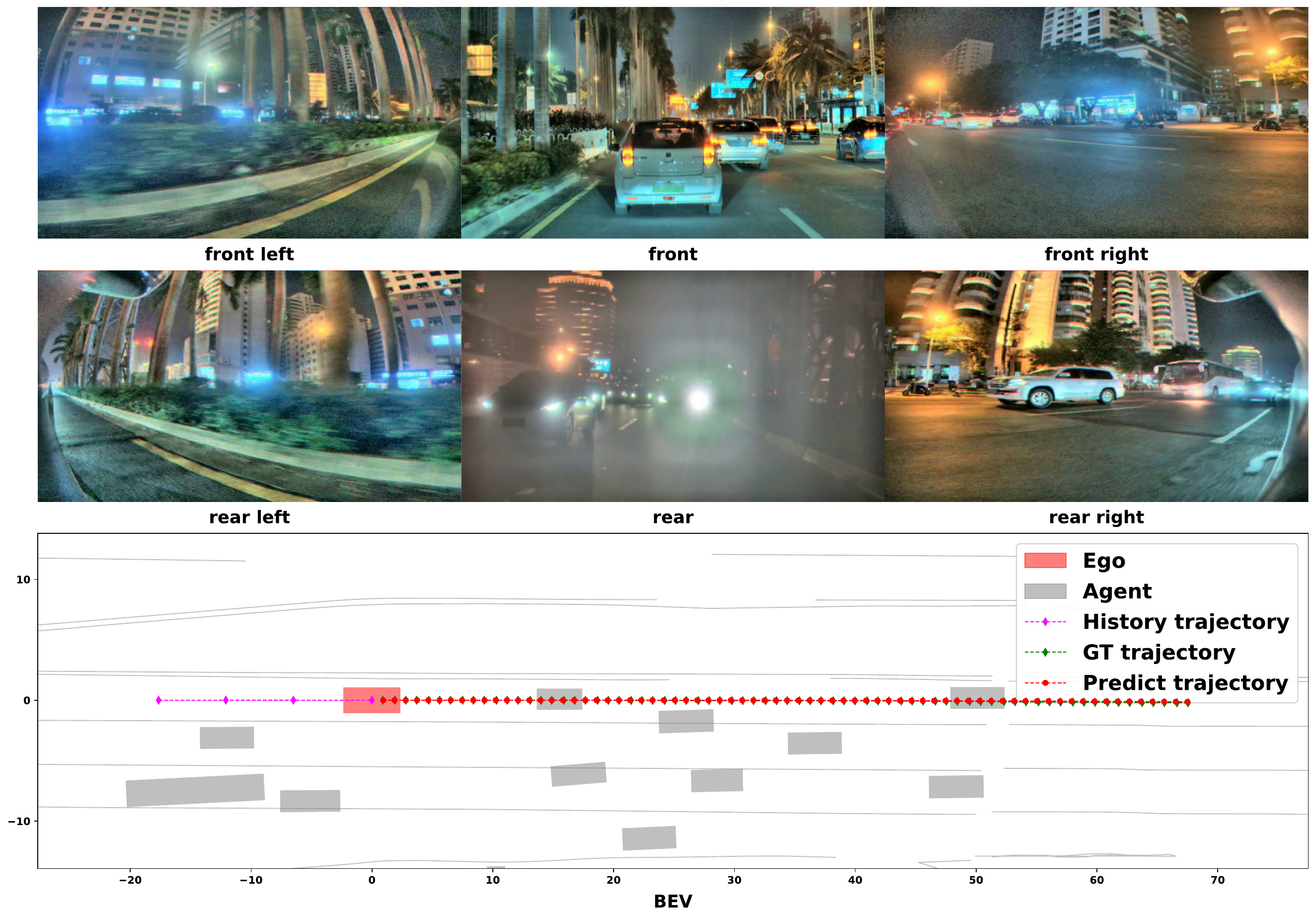}
}
\end{subfigure}

\vspace{1mm}

\begin{subfigure}{0.72\linewidth}
\centering
\scalebox{0.98}[0.90]{\includegraphics[width=\linewidth]{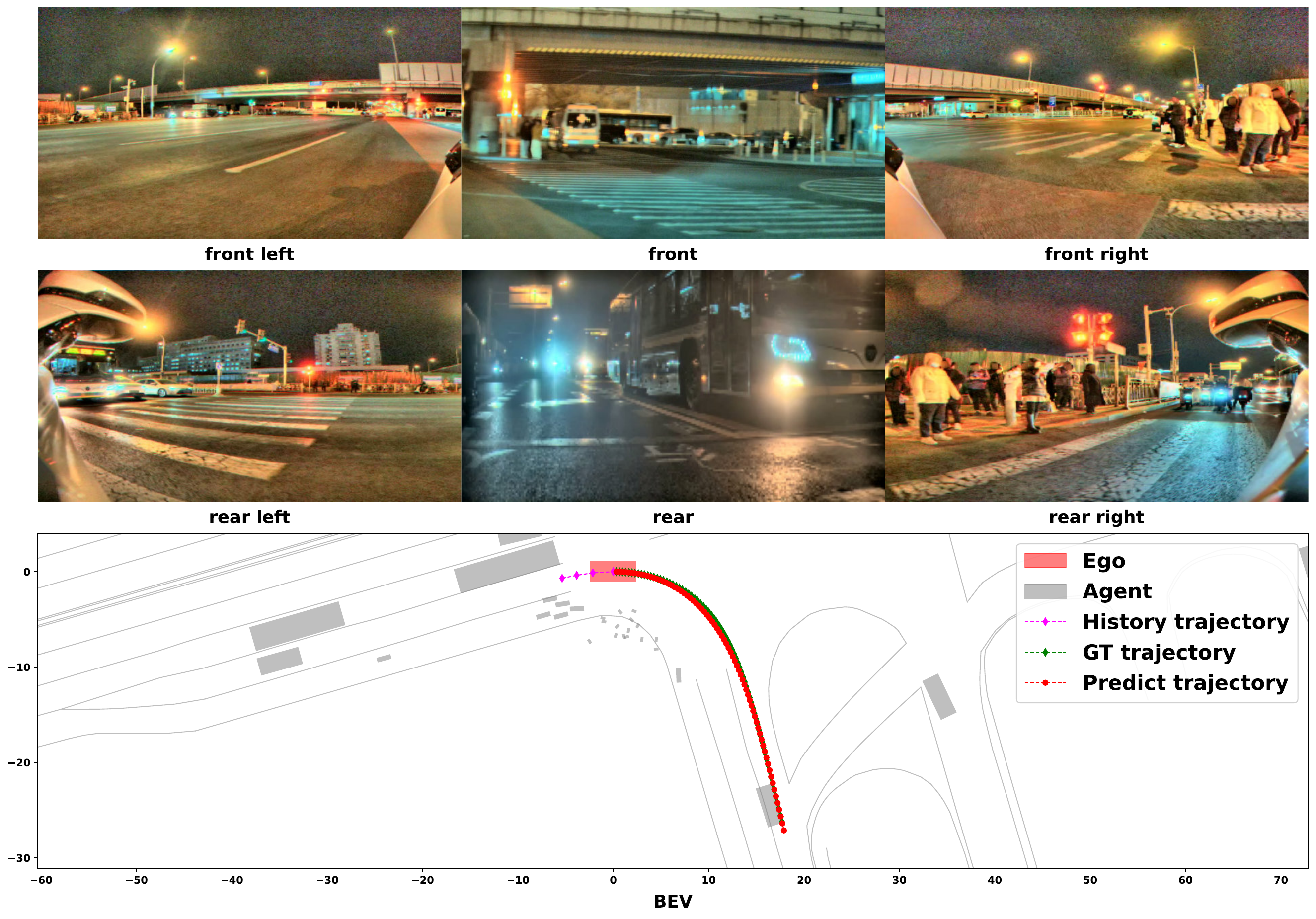}
}
\end{subfigure}

\vspace{1mm}

\begin{subfigure}{0.72\linewidth}
\centering
\scalebox{0.98}[0.90]{\includegraphics[width=\linewidth]{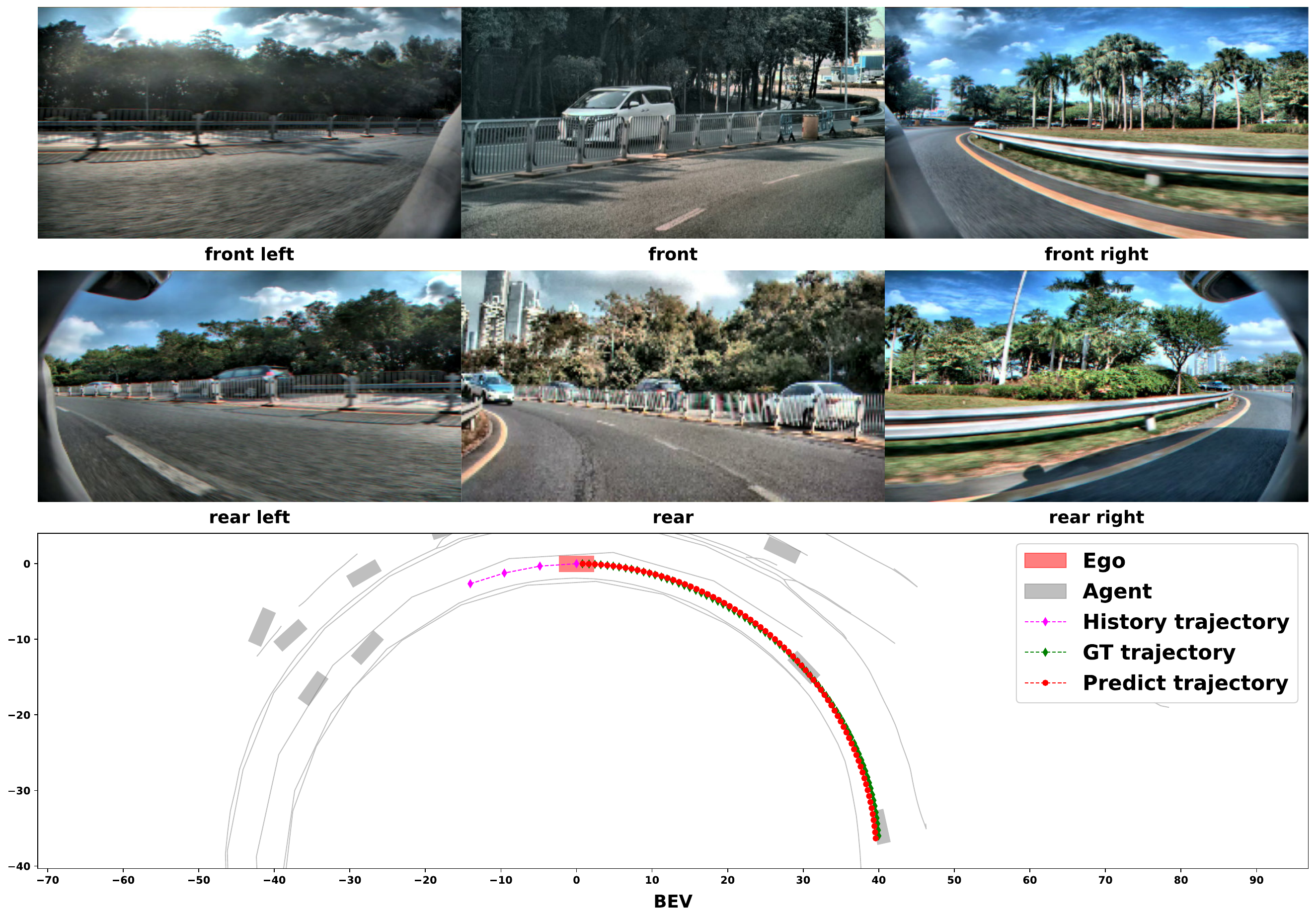}
}
\end{subfigure}

\captionsetup{skip=4pt}
\caption{\textbf{Internal 1,000\,km qualitative examples.}
Representative long-tail interactions and camera conditions
from the private evaluation set.}
\label{fig:app_internal}

\end{figure}

\begin{figure}[p]
\centering
\begin{subfigure}{0.72\linewidth}
\includegraphics[width=\linewidth]{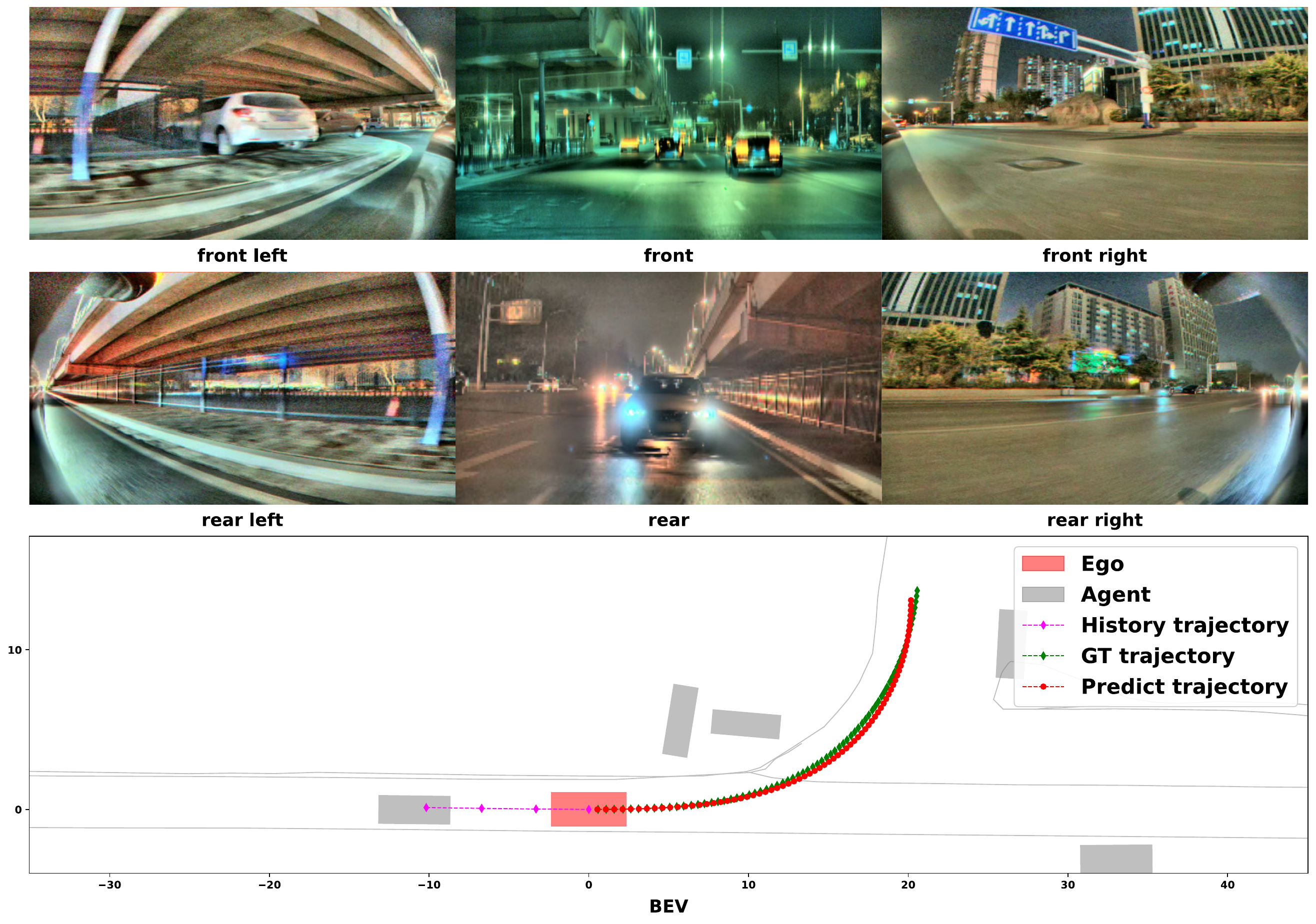}
\end{subfigure}

\vspace{2.5mm}
\begin{subfigure}{0.72\linewidth}
\includegraphics[width=\linewidth]{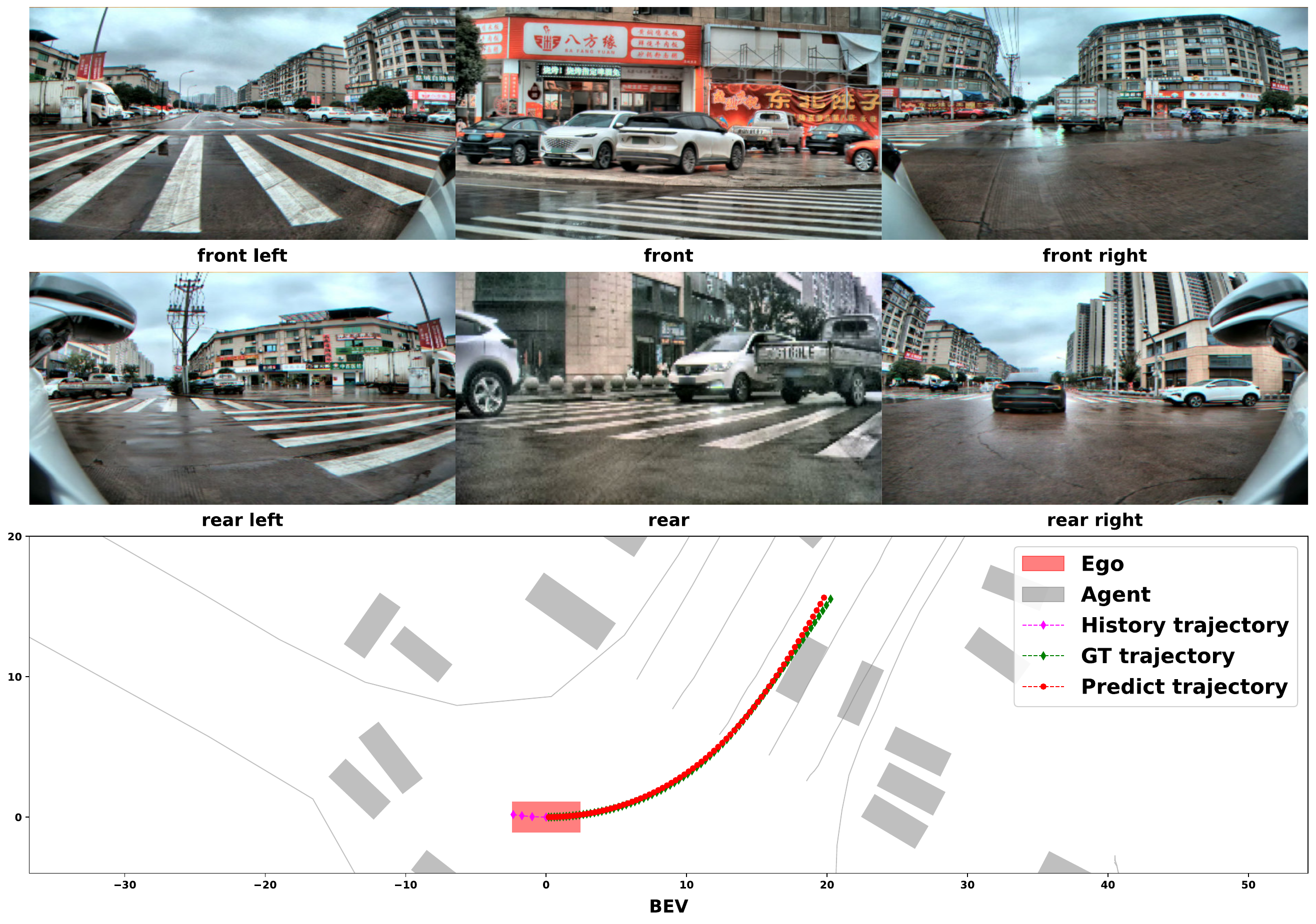}
\end{subfigure}

\vspace{2.5mm}
\begin{subfigure}{0.72\linewidth}
\includegraphics[width=\linewidth]{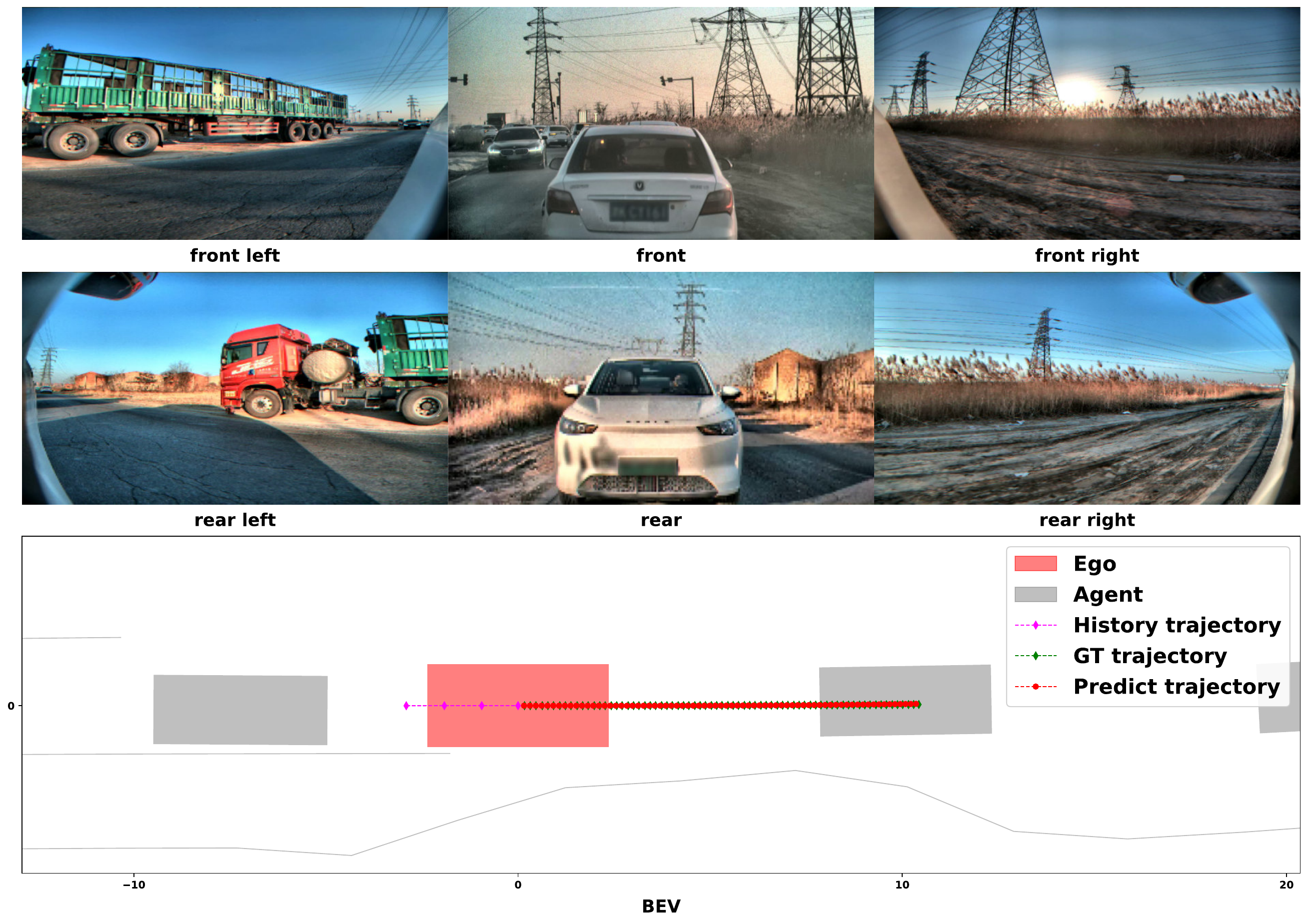}
\end{subfigure}
\caption{\textbf{Internal 1,000\,km qualitative examples.}
Representative long-tail interactions and camera conditions
from the private evaluation set. (part 2/2).}
\label{fig:app_internal_b}
\end{figure}

The internal set contains long-tail interactions and camera conditions not present in NAVSIM. Figures~\ref{fig:app_internal} and Figures~\ref{fig:app_internal_b} shows twelve representative cases from the 1,000 km evaluation; the category-level aggregate remains the one reported in Table~\ref{tab:internal}.

\subsection{Regression and flow-matching baselines}
\begin{figure}[p]
\centering

\begin{subfigure}{0.72\linewidth}
\centering
\scalebox{0.98}[0.90]{\includegraphics[width=\linewidth]{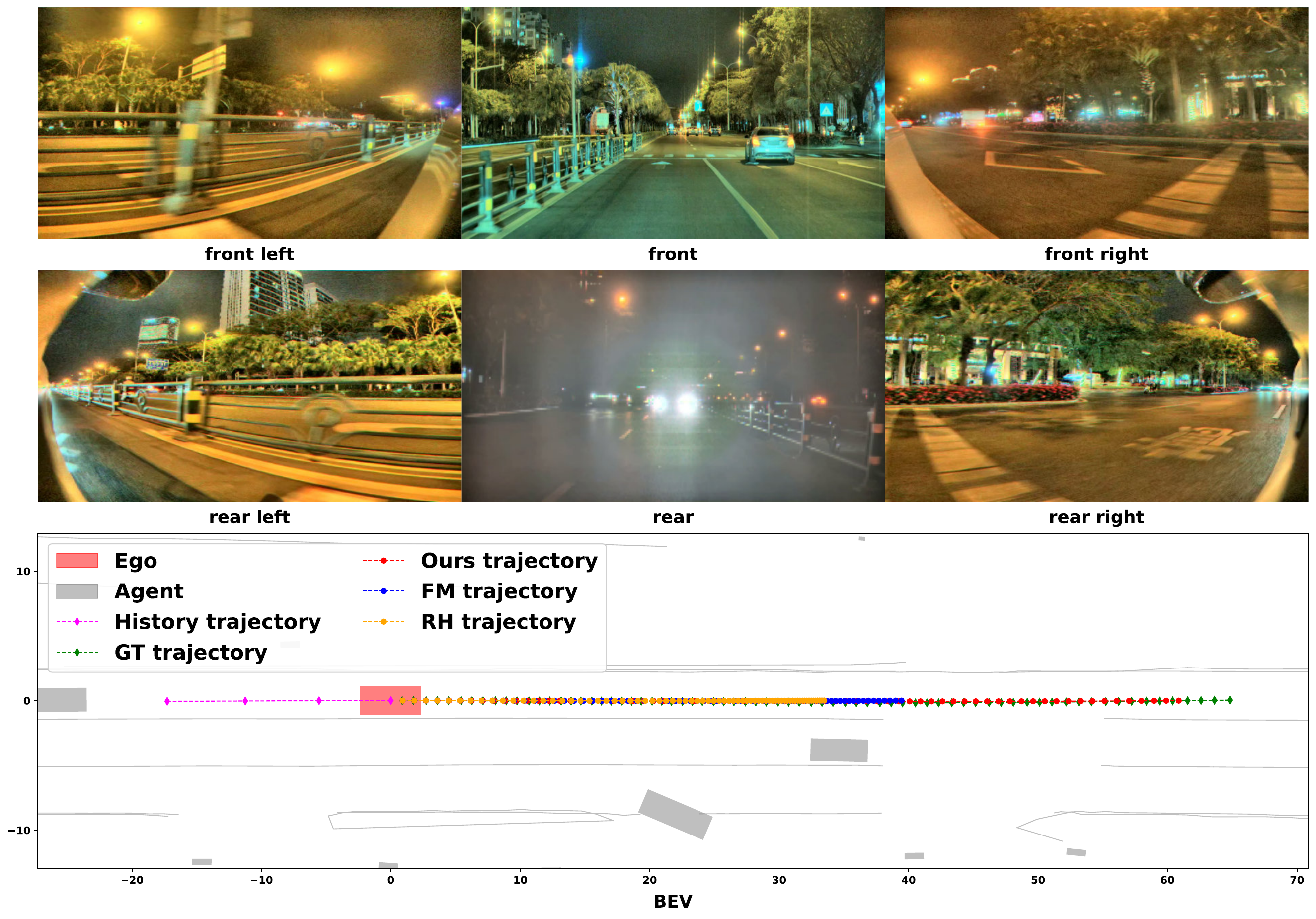}
}
\end{subfigure}

\vspace{1mm}

\begin{subfigure}{0.72\linewidth}
\centering
\scalebox{0.98}[0.90]{\includegraphics[width=\linewidth]{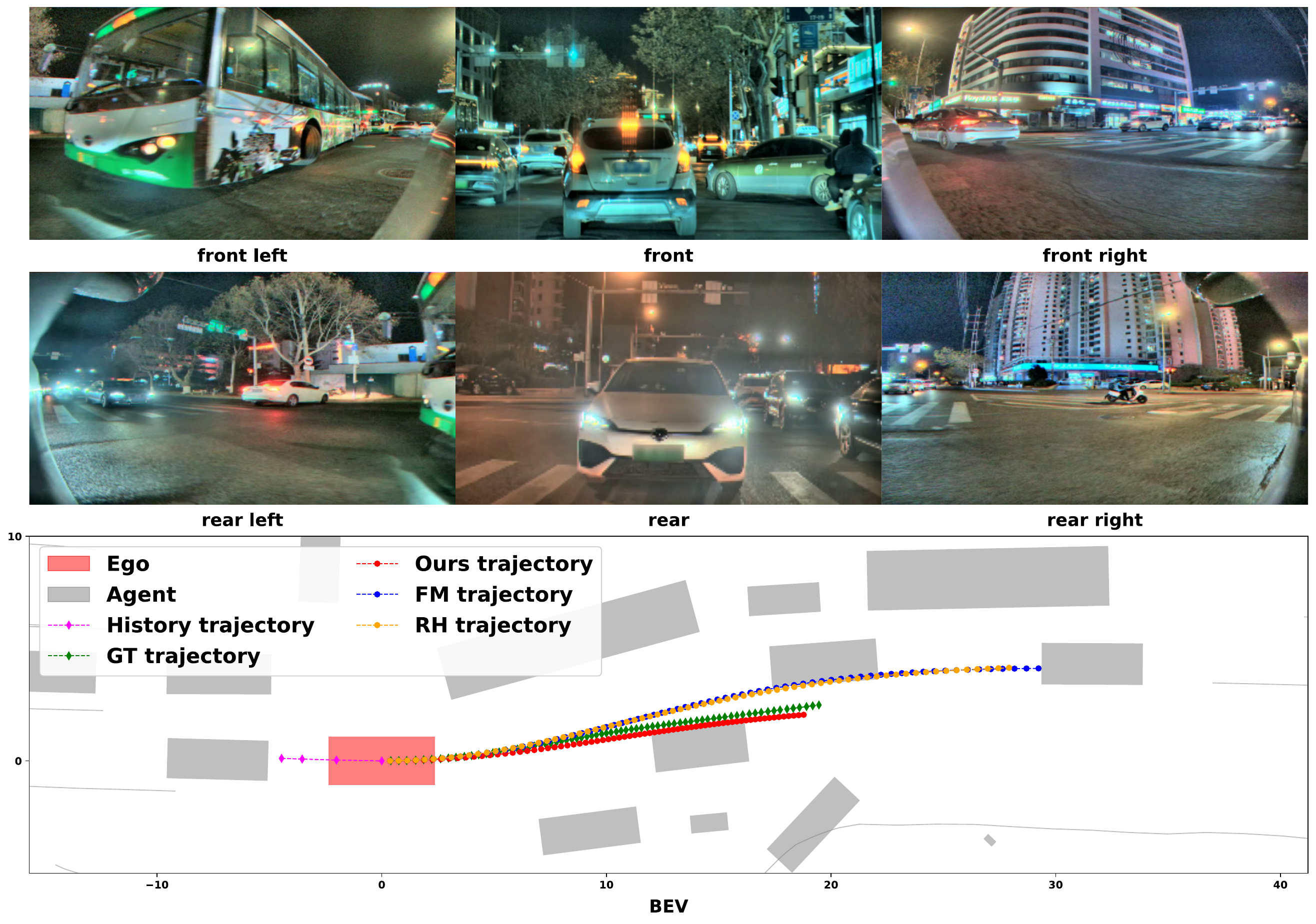}
}
\end{subfigure}

\vspace{1mm}

\begin{subfigure}{0.72\linewidth}
\centering
\scalebox{0.98}[0.90]{\includegraphics[width=\linewidth]{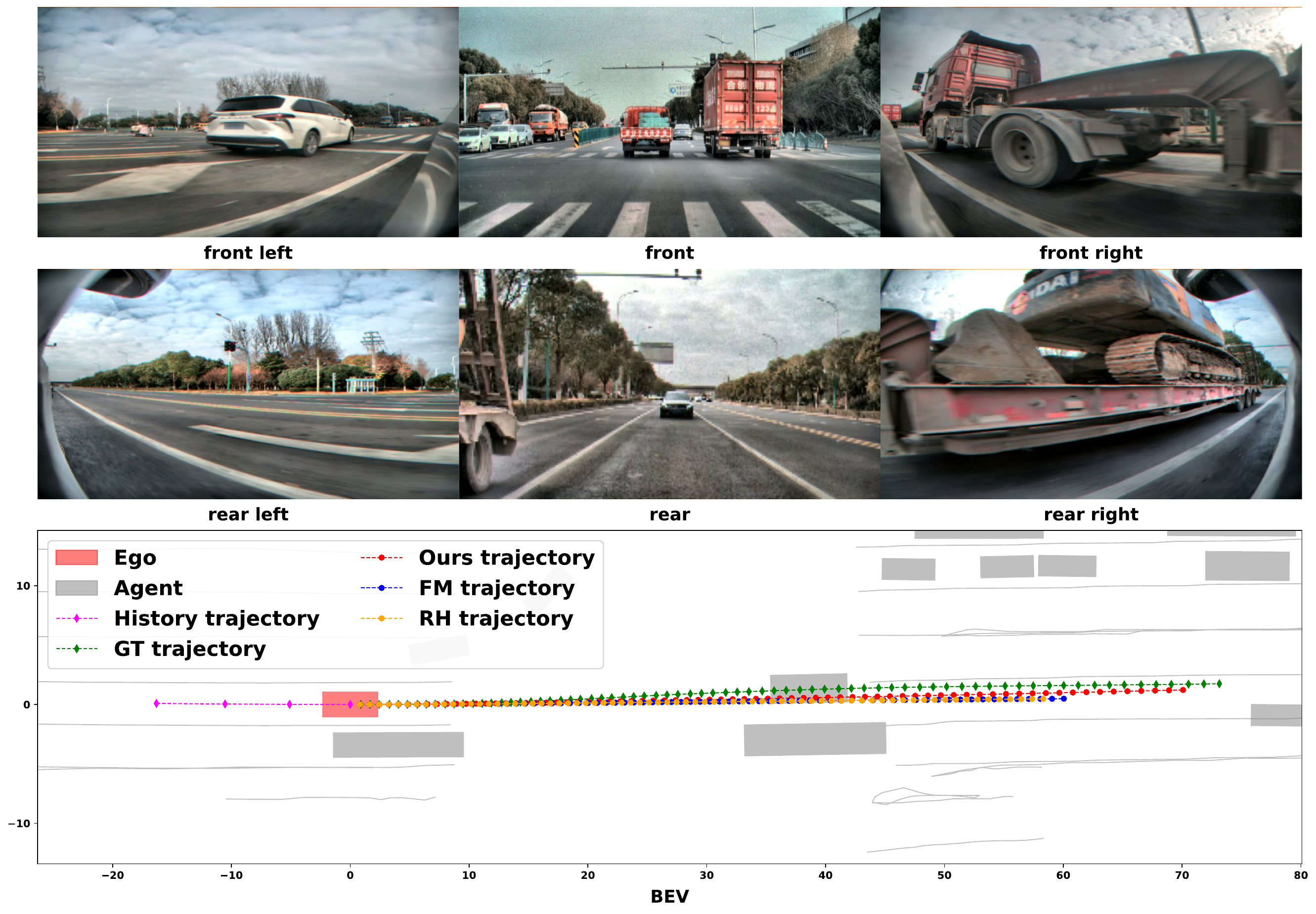}
}
\end{subfigure}

\captionsetup{skip=4pt}
\caption{\textbf{Internal qualitative comparison of VLM+regression, VLM+flow matching, and AUGUstus (part 1/2).}}
\label{fig:app_diff_a}

\end{figure}

\begin{figure}[p]
\centering
\begin{subfigure}{0.72\linewidth}
\includegraphics[width=\linewidth]{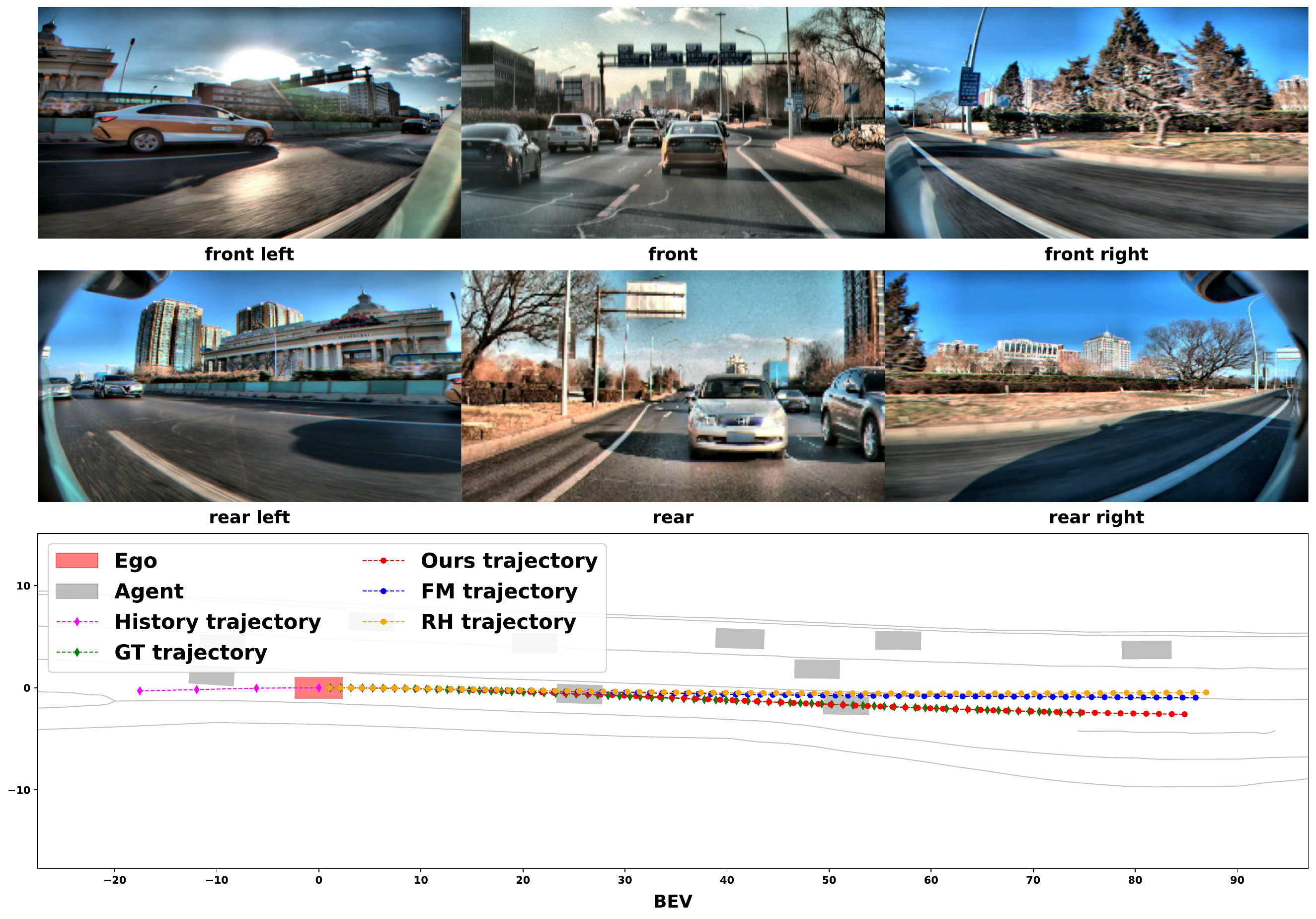}
\end{subfigure}

\vspace{2.5mm}
\begin{subfigure}{0.72\linewidth}
\includegraphics[width=\linewidth]{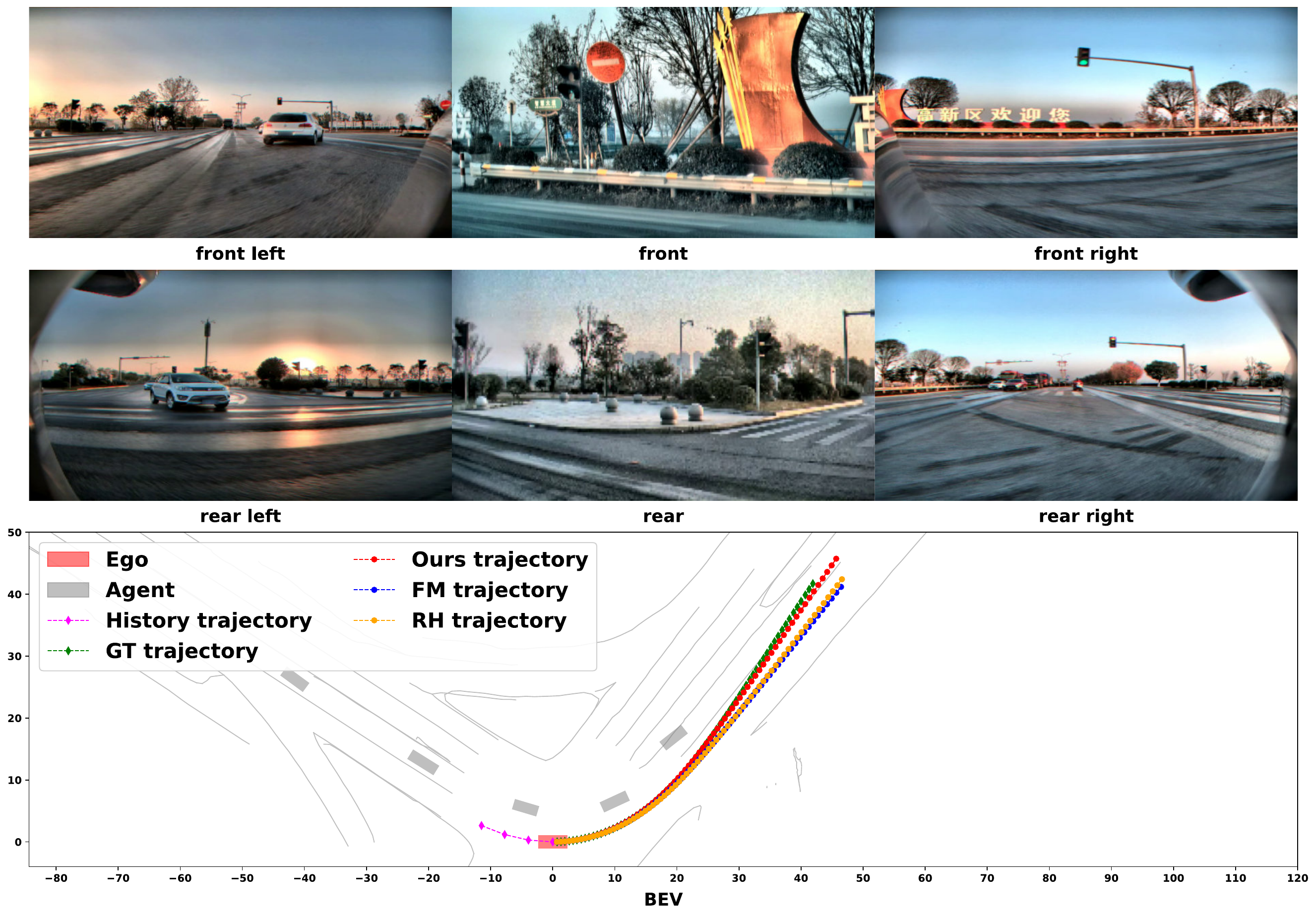}
\end{subfigure}

\vspace{2.5mm}
\begin{subfigure}{0.72\linewidth}
\includegraphics[width=\linewidth]{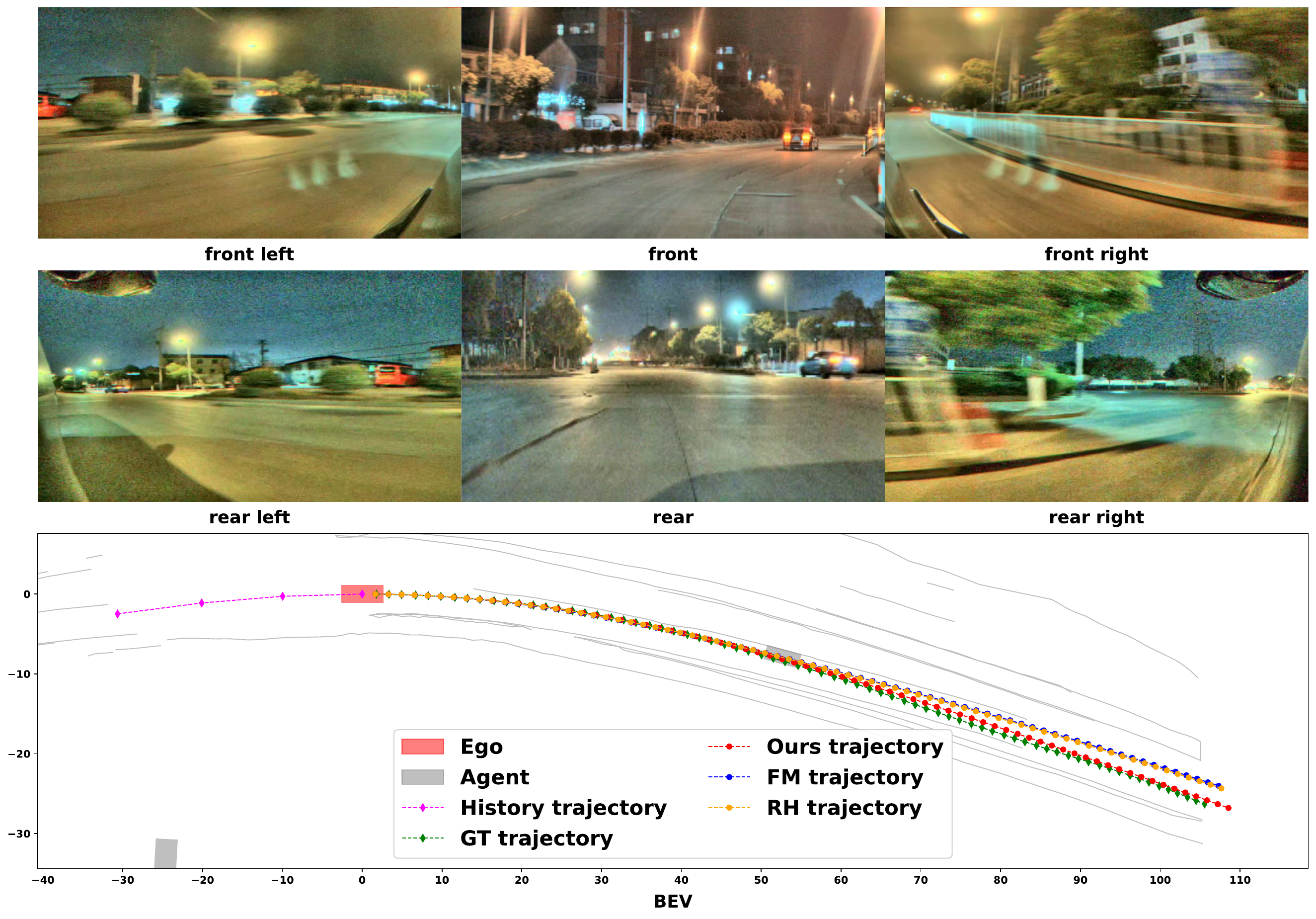}
\end{subfigure}
\caption{\textbf{Internal qualitative comparison of VLM+regression, VLM+flow matching, and AUGUstus (part 2/2).}}
\label{fig:app_diff_b}
\end{figure}

Figure~\ref{fig:app_diff_a} and Figure~\ref{fig:app_diff_b} keeps the direct VLM+regression and VLM+flow-matching comparison in one image. Both baselines use the same scene input; the comparison isolates the action parameterization and the shared CoT training rather than a sensor change.

\FloatBarrier
\FloatBarrier

\FloatBarrier
\FloatBarrier

\clearpage

\FloatBarrier
\FloatBarrier
\FloatBarrier

\FloatBarrier
\subsection{Real-vehicle visualization}
The real-vehicle examples use the same action-only deployment path as the latency measurement. The camera mosaic and bird's-eye trajectory in Fig.~\ref{fig:app_real} show that the one-step output remains smooth in nighttime and multi-camera conditions.

\begin{center}

\scalebox{0.98}[0.90]{\includegraphics[width=0.95\textwidth]{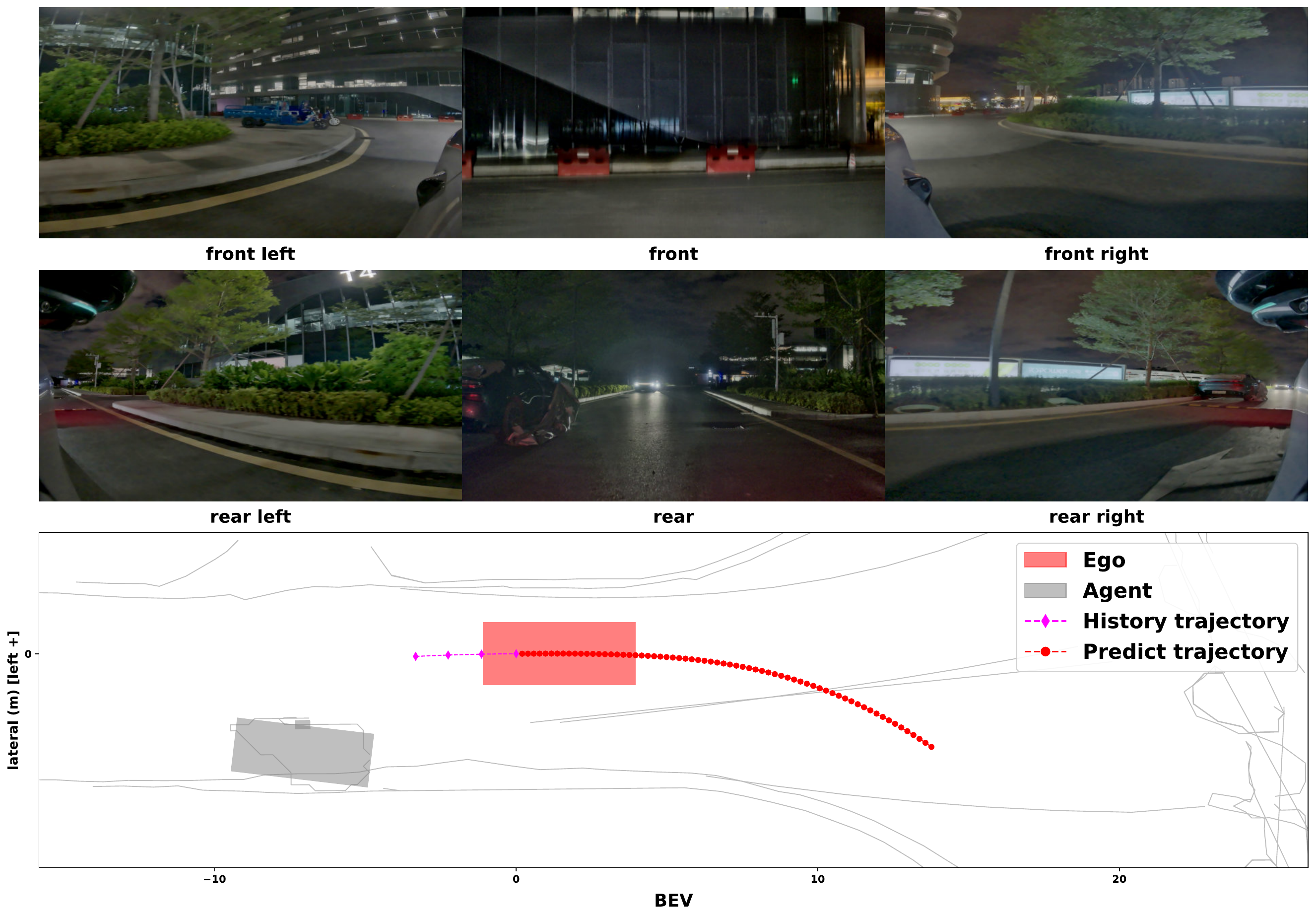}
}

\vspace{2pt}

\scalebox{0.98}[0.90]{\includegraphics[width=0.95\textwidth]{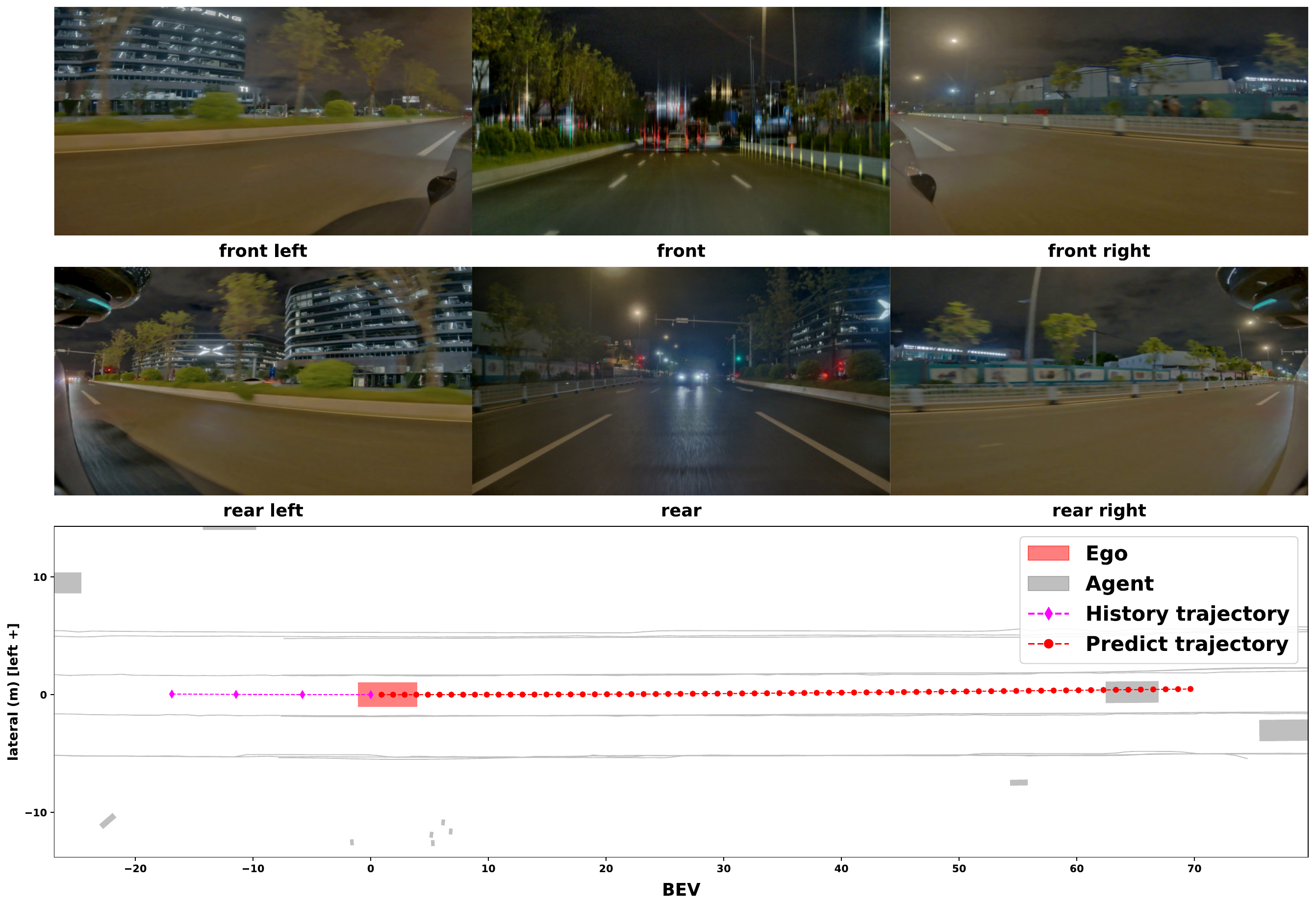}
}

\captionsetup{skip=4pt}
\captionof{figure}{\textbf{Real-vehicle qualitative examples.} The deployed action-only path runs at 12\,Hz on the vehicle. The BEV trajectory is generated directly by the shared model; CoT decoding is omitted from this loop.}
\label{fig:app_real}
\end{center}

\end{document}